\documentclass[sigconf]{acmart}

\usepackage{amsthm}
\usepackage{amsmath}
\usepackage{algorithm}
\usepackage{enumitem}
\usepackage{picinpar}
\usepackage{lineno}
\usepackage{graphicx}

\usepackage{algorithmicx}
\usepackage[noend]{algpseudocode}
\usepackage{subfigure}
\usepackage{multirow}
\usepackage{color}
\usepackage{balance}
\usepackage{enumitem}
\usepackage{hhline}
\usepackage[normalem]{ulem}
\usepackage{booktabs}
\usepackage{wrapfig}
\usepackage{cancel}
\usepackage{hyperref}
\usepackage{makecell}

\usepackage{tcolorbox}

\newtcolorbox{rqbox}{
  colback=gray!8,
  colframe=black!40,
  boxrule=0.6pt,
  arc=2pt,
  left=6pt,
  right=6pt,
  top=4pt,
  bottom=4pt
}

\newcommand{\stitle}[1]{\vspace{1mm} \noindent {\bf #1}}

\newcommand{\method}[1]{\textsc{#1}}

\newcommand{\eat}[1]{}

\newcommand{\stkout}[1]{\ifmmode\text{\sout{\ensuremath{#1}}}\else\sout{#1}\fi}

\AtBeginDocument{%
  }

\setcopyright{acmlicensed}
\copyrightyear{2018}
\acmYear{2018}
\acmDOI{XXXXXXX.XXXXXXX}
\acmConference[Conference acronym 'XX]{Make sure to enter the correct
  conference title from your rights confirmation email}{June 03--05,
  2018}{Woodstock, NY}
\acmISBN{978-1-4503-XXXX-X/2018/06}

\begin{document}

%%
%% The "title" command has an optional parameter,
%% allowing the author to define a "short title" to be used in page headers.
\title[Evaluating Progress in Graph Foundation Models: A Comprehensive Benchmark and New Insights]{Evaluating Progress in Graph Foundation Models:\\A Comprehensive Benchmark and New Insights}

%%
%% The "author" command and its associated commands are used to define
%% the authors and their affiliations.
%% Of note is the shared affiliation of the first two authors, and the
%% "authornote" and "authornotemark" commands
%% used to denote shared contribution to the research.
\author{Xingtong Yu}
\affiliation{%
 \institution{The Chinese University of Hong Kong*}
  \country{China}}
\email{xtyu@se.cuhk.edu.hk}

\author{Shenghua Ye}
\affiliation{%
 \institution{Singapore Management University}
  \country{Singapore}}
\email{shenghua.ye.2025@phdcs.smu.edu.sg}

\author{Ruijuan Liang}
\affiliation{%
 \institution{University of Science and Technology of China}
  \country{China}}
\email{lrjuan@mail.ustc.edu.cn}

\author{Chang Zhou}
\affiliation{%
 \institution{University of Science and Technology of China}
  \country{China}}
\email{zhouchang21sy@mail.ustc.edu.cn}

\author{Hong Cheng}
\affiliation{%
 \institution{The Chinese University of Hong Kong}
  \country{China}}
\email{hcheng@se.cuhk.edu.hk}

\author{Xinming Zhang}
\affiliation{%
 \institution{University of Science and Technology of China}
  \country{China}}
\email{xinming@ustc.edu.cn}

\author{Yuan Fang}
\affiliation{%
  \institution{Singapore Management University}
  \country{Singapore}}
\email{yfang@smu.edu.sg}

\thanks{
    $^*$Most of this work was done while at Singapore Management University.
}

%%
%% By default, the full list of authors will be used in the page
%% headers. Often, this list is too long, and will overlap
%% other information printed in the page headers. This command allows
%% the author to define a more concise list
%% of authors' names for this purpose.

%%
%% The abstract is a short summary of the work to be presented in the
%% article.
\begin{abstract}
Graph foundation models (GFM) aim to acquire transferable knowledge by pre-training on diverse graphs, which can be adapted to various downstream tasks. However, domain shift in graphs is inherently two-dimensional: graphs differ not only in what they describe (\emph{topic} domains) but also in how they are represented (\emph{format} domains). Most existing GFM benchmarks vary only topic domains, %shifts under a fixed format---
thereby obscuring how knowledge transfers across both dimensions. %In addition, they cover only a narrow slice of the rapidly evolving GFM landscape, making it difficult to draw conclusions that hold for modern methods.
We present a new benchmark that jointly evaluates topic and format gaps across the full GFM pipeline, including multi-domain self-supervised pre-training and few-shot downstream adaptation, and provides a timely evaluation of recent GFMs in the rapidly evolving landscape. Our protocol enables controlled assessment in four settings: (i) pre-training on diverse topics and formats, while adapting to unseen downstream datasets; (ii) same pre-training as in (i), while adapting to seen datasets; (iii) pre-training on a single topic domain, while adapting to other topics; (iv) pre-training on a base format, while adapting to other formats. This two-axis evaluation disentangles semantic generalization from robustness to representational shifts. We conduct extensive evaluations of eight state-of-the-art GFMs on 33 datasets spanning seven topic domains and six format domains, surfacing new empirical observations and practical insights for future research.
Codes/data are available at \textcolor{blue}{\url{https://github.com/smufang/GFMBenchmark}}.
\end{abstract}

%%
%% The code below is generated by the tool at http://dl.acm.org/ccs.cfm.
%% Please copy and paste the code instead of the example below.
%%


%%
%% Keywords. The author(s) should pick words that accurately describe
%% the work being presented. Separate the keywords with commas.
%% A "teaser" image appears between the author and affiliation
%% information and the body of the document, and typically spans the
%% page.

\received{20 February 2007}
\received[revised]{12 March 2009}
\received[accepted]{5 June 2009}

%%
%% This command processes the author and affiliation and title
%% information and builds the first part of the formatted document.
\maketitle
\section{Introduction}
Foundation models have emerged as a unifying paradigm in modern artificial intelligence, where models are pre-trained on broad and heterogeneous data and subsequently adapted to a wide range of downstream tasks \cite{achiam2023gpt,touvron2023llama}. Their success in natural language processing \cite{achiam2023gpt,touvron2023llama,liu2024deepseek} and computer vision \cite{yuan2021florence,wang2023internimage} demonstrates that large-scale pre-training can produce general-purpose representations that transfer across tasks and data distributions. Motivated by the progress, recent studies have extended this paradigm to graph-structured data, giving rise to \emph{graph foundation models} (GFM) \cite{liu2025graph,mao2024position}. By pre-training on graph datasets from multiple sources or domains, GFMs aim to learn transferable structural and semantic knowledge that generalizes to unseen tasks, graphs, or domains with limited task-specific labeled data.

However, benchmarking GFMs %is substantially more challenging than in unstructured modalities, 
can be complex and challenging, since domain shift in graphs is inherently multi-faceted. Graph heterogeneity manifests not only in what graphs represent but also in how they are structured and represented. Inspired by previous work \cite{wettigorganize}, we categorize graph domains into \emph{topic} domains and \emph{format} domains. From the semantic perspective, graphs arise from diverse \textbf{topic domains}, including citation networks \cite{radicchi2011citation,mclaren2022citation}, social networks and the Web \cite{freeman2004development,marin2011social}, e-commerce graphs \cite{hu2020gpt,zhao2020persistence}, molecular graphs \cite{kearnes2016molecular,you2018graph}, protein interaction networks \cite{giuliani2008proteins,borgwardt2005protein}, and financial networks \cite{bardoscia2021physics,haldane2013rethinking}. These domains differ in node/edge semantics and feature spaces, creating pronounced semantic gaps that challenge multi-domain pre-training and cross-domain transfer. Beyond semantics, graphs also vary widely in \textbf{format domains}, including homogeneous vs.~heterogeneous graphs \cite{zhang2019heterogeneous,wang2022survey}, homophilic vs.~heterophilic structures \cite{rossi2024edge,bi2024make}, static vs.~dynamic graphs \cite{harary1997dynamic,manessi2020dynamic}, knowledge graphs \cite{chen2020review}, and text-attributed graphs \cite{zhaolearning,wen2023augmenting}. These format differences impose distinct modeling assumptions across topological patterns, temporal dependencies, and attribute or relational schemas, which can fundamentally affect both pre-training and adaptation.

Despite the variety across topics and formats, existing GFM benchmarks typically adopt a one-dimensional notion of domain diversity \cite{yang2025benchmarking,xu2024graphfm,liu2025roft,chen2024text}. Specifically, they vary topic domains while keeping the graph format fixed \cite{yang2025benchmarking,xu2024graphfm,chen2024text}, alternatively restrict evaluation to a narrow application area \cite{liu2025roft}. Under such a setup, it remains unclear whether observed transfer performance reflects universal semantic and structural generalization across topics and formats.
This mismatch between the heterogeneity of graph data and the design of existing benchmarks leads to an incomplete and even misleading understanding of GFMs. 
Without explicitly disentangling topic and format domains, benchmark results risk conflating distinct sources of generalization and obscuring key failure modes.
Beyond dataset design, existing benchmarks are also limited in method coverage: they often evaluate a specific type of methods \cite{liu2025roft} or tend to lag behind the rapidly evolving literature~\cite{yang2025benchmarking,xu2024graphfm,chen2024text}. As a result, conclusions from prior benchmark studies may not extrapolate to more general or newer GFMs.

To address these gaps, we present a new benchmark for GFMs that explicitly and jointly accounts for topic domain gaps and format domain gaps throughout both pre-training and downstream adaptation, while covering representative GFMs across major modeling regimes. Our benchmark is built from diverse datasets that covers commonly encountered topic domains and format domains. We construct unified evaluation settings by (i) performing multi-domain pre-training over graphs spanning diverse topics and formats, and (ii) adapting the resulting models to downstream datasets drawn from \emph{seen} and \emph{unseen} domains. Crucially, by controlling the topic and format facets, our benchmark enables fine-grained analyses of (a) cross-topic transfer, (b) cross-format transfer, and (c) transfer under \emph{joint} topic-and-format shifts. 
This two-axis evaluation, together with broader method coverage, makes it possible to isolate different forms of generalization and to characterize their interactions.

In summary, we make the following contributions.
(1) We formalize a two-dimensional view of graph domains, distinguishing their \emph{topics} from \emph{formats}.
%, and identify a key limitation of existing GFM benchmarks that evaluate only one dimension.
(2) We construct a comprehensive benchmark that spans diverse topic and format domains and includes a broad range of representative GFMs, enabling systematic multi-domain pre-training and cross-domain evaluation.
(3) We introduce unified evaluation settings that separately and jointly assess %cross-topic transfer, cross-format transfer, and joint domain shifts, 
the effect of topic and format shifts in both seen and unseen downstream targets.
(4) We provide extensive empirical analyses that reveal previously uncharacterized generalization behaviors and limitations of existing GFMs,
%under combined topic and format shifts, 
offering actionable insights for future GFM design.

\section{Related Work}
% \cite{yang2025benchmarking} only conduct NC and LP on two hetergeneous graph datasets DBLP, IMDB, and pre-train on a single dataset OGA-CS.
% \cite{xu2024graphfm} only leverage 4 citation dataset Cora, Citeseer, Pubmed, OGB and two social network Flickr and Reddit, and the methods it evluate are graph self-supervised learning methods, which conduct experiments on sigle domian. Far from GFM, pre-training on multi-domain and cross domain adaptation.
% \cite{liu2025roft} only focus on molecular graphs, doesn't evaluate GFM across multiple domains.
% \cite{chen2024text} focus on text space datasets and methods using LLM to  generate embeddings to obtain a unified feature space.

% they only focus on serval topic domain, while ignore the format domain 

\stitle{Graph foundation models.}
Existing GFMs can be broadly grouped by how they mitigate domain gaps in graphs. 
Several GFMs \cite{yu2024text,zhao2024all,wang2025multi} target feature mismatch  by introducing domain-conditioned tokens and prompts, shared coordinators, or lightweight adapters that allow a single backbone to operate across heterogeneous input spaces.
Another line of GFMs \cite{yu2025samgpt,wang2025multi,wang2024gft,zhao2024all} focus on structural mismatch by aligning topology statistics or modulating message passing with structure-aware mechanisms, aiming to reduce domain-dependent inductive biases and improve transfer across graphs with different structural regimes.
Additionally, some GFMs \cite{zhu2025graphclip,wen2023augmenting,he2025unigraph2,wang2024gft} leverage textual and multimodal signals as an external semantic anchor, aligning graph representations with text or other modalities via contrastive or prompt-based objectives. While effective when such signals are available, their generalization to universal graph datasets is limited.
%these methods typically require additional metadata, such as text attributes or label names, which reduces coverage and complicates comparisons across graph formats . 
Another orthogonal consideration is supervision in pre-training. Some GFM-style models \cite{liu2023one,huang2023prodigy,tang2023graphgpt,tang2024higpt,xia2024opengraph,yuangraver,chen2024llaga,li2024zerog,zhao2024graphany} require labeled source-domain data, which diverges from the traditional pre-training paradigm, and are therefore outside the scope of this benchmark. We summarize representative GFMs in Table~\ref{table.gfms}, with further details in Appendix~\ref{app.baselines}.

\begin{table}[tbp] 
    \centering
    \footnotesize
    \caption{Comparison of representative GFMs.}
    \vspace{-2mm}
    \label{table.gfms}%
    \addtolength{\tabcolsep}{0.4mm}
    %\resizebox{1\linewidth}{!}{%
    \begin{tabular}{@{}c|c|cccccc|c|c@{}}
    \toprule
    \multirow{2}{*}{\makecell[c]{\\\\Methods}} & \multirow{2}{*}{\rotatebox{90}{Labeled source\ }} &\multicolumn{6}{c|}{Alignment} &\multirow{2}{*}{\makecell[c]{\\\\PLM/LLM}} &\multirow{2}{*}{\rotatebox{90}{Adaptation\ \ \ }}\\
    & & \rotatebox{90}{Structure} & \rotatebox{90}{Feature} & \rotatebox{90}{Codebook} & \rotatebox{90}{Graph mod.} & \rotatebox{90}{Feature mod.} & \rotatebox{90}{Textual} & & \\
    \midrule
    \multicolumn{10}{c}{Benchmarked graph foundation models}\\\midrule
    \method{GCOPE} \cite{zhao2024all} & $\times$ & $\times$ & $\checkmark$ & $\times$ & $\checkmark$ & $\times$ &$\times$ &$\times$ &F, P\\
    \method{MDGPT} \cite{yu2024text} & $\times$ & $\times$ & $\checkmark$ & $\times$ & $\times$ & $\checkmark$ &$\times$ &$\times$ &P \\
    \method{MDGFM} \cite{wang2025multi} & $\times$ & $\times$ & $\checkmark$ & $\times$ & $\times$ & $\times$ &$\times$ &$\times$ &P \\
    \method{SAMGPT} \cite{yu2025samgpt} & $\times$ & $\checkmark$ & $\checkmark$ & $\times$ & $\times$ & $\checkmark$ &$\times$ &$\times$ &P \\
    \method{G2P2}~\cite{wen2023augmenting} & $\times$ & $\times$ & $\checkmark$ & $\times$ & $\times$ & $\checkmark$ &$\checkmark$ &E, A &P\\
    \method{GraphCLIP}~\cite{zhu2025graphclip} & $\times$ & $\times$ & $\checkmark$ & $\times$ & $\times$ & $\checkmark$ &$\checkmark$ &A &P \\
    \method{GFT}~\cite{wang2024gft} & $\times$ & $\checkmark$ & $\checkmark$ & $\checkmark$ & $\times$ & $\times$ &$\times$ &E &F, P \\
    \method{UniGraph2}~\cite{he2025unigraph2} & $\times$ & $\checkmark$ & $\checkmark$ & $\times$ & $\times$ & $\checkmark$ &$\checkmark$ &E &F\\\midrule
    \multicolumn{10}{c}{Other GFM-style methods requiring labeled source data (out of scope)}\\\midrule
    \method{OFA} \cite{liu2023one} & $\checkmark$ & $\times$ & $\checkmark$ & $\times$ & $\checkmark$ & $\times$ &$\checkmark$ &E &P\\
    \method{Prodigy} \cite{huang2023prodigy} & $\checkmark$ & $\times$ & $\checkmark$ & $\times$ & $\checkmark$ & $\times$ &$\times$ &$\times$ &P\\
    \method{GraphGPT} \cite{tang2023graphgpt} & $\checkmark$ & $\times$ & $\checkmark$ & $\times$ & $\times$ & $\times$ &$\checkmark$ &A, P &P\\
    \method{HiGPT} \cite{tang2024higpt} & $\checkmark$ & $\times$ & $\checkmark$ & $\times$ & $\times$ & $\times$ &$\checkmark$ &A, P &P\\
    \method{OpenGraph} \cite{xia2024opengraph} & $\checkmark$ & $\checkmark$ & $\checkmark$ & $\times$ & $\checkmark$ & $\times$ &$\times$ &G & $\times$\\
    \method{GRAVER} \cite{yuangraver} & $\checkmark$ & $\checkmark$ & $\times$ & $\times$ & $\checkmark$ & $\checkmark$ &$\times$ &E &F, P \\
    \method{Llaga} \cite{chen2024llaga} & $\checkmark$ & $\times$ & $\checkmark$ & $\times$ & $\times$ & $\times$ &$\checkmark$ &P &P\\
    \method{ZeroG} \cite{li2024zerog} & $\checkmark$ & $\times$ & $\checkmark$ & $\times$ & $\times$ & $\times$ &$\checkmark$ &A &P\\
    \method{GraphAny} \cite{zhao2024graphany} & $\checkmark$ & $\checkmark$ & $\checkmark$ & $\times$ & $\times$ & $\checkmark$ &$\times$ &$\times$ &F\\
    \bottomrule
    \end{tabular}%}
    \vspace{1mm}
    \parbox{0.99\linewidth}{\footnotesize Under \emph{PLM/LLM}, `E': extractor, `A': alignment, `P': predictor, and `G': generator. Under \emph{Adaptation}, `F': finetuning, and `P': prompting. }
    %\vspace{-3mm}
\end{table}

\stitle{Benchmarks for graph foundation models.}
Existing GFM benchmarks characterize domain diversity along a single axis. They typically vary topic domains while implicitly fixing (or tightly constraining) the graph format \cite{yang2025benchmarking,xu2024graphfm,liu2025roft,chen2024text}. This one-dimensional design makes results hard to attribute: observed gains may stem from semantic or structural generalization, robustness to format changes, or simply a favorable match between the pre-training and downstream formats, but these factors are not separated.
Beyond this shared limitation, existing benchmarks also differ in scope and protocol. \citet{yang2025benchmarking} primarily focuses on heterogeneous graphs and performs pre-training on a single dataset, which differs from the standard GFM setting of multi-domain pre-training.
\citet{xu2024graphfm} primarily benchmarks graph self-supervised learning on a small set of citation and social datasets, rather than the standard GFM pipeline of multi-domain pre-training followed by cross-domain adaptation. \citet{liu2025roft} provides a careful study of robust fine-tuning, but its focus on molecular graphs limits the conclusions to a single topic domain. \citet{chen2024text} targets text-attributed graphs where LLM-derived embeddings help unify feature spaces, and is not designed to evaluate broader graph formats.
Our benchmark complements these efforts by treating domain shift as inherently two-dimensional, across both semantic topics and graph formats, by evaluating GFMs in the full pipeline of multi-domain self-supervised pre-training followed by cross-domain adaptation with limited task-specific labeled data.

\section{Target Models and Domain Composition}

In this section, we introduce the evaluated GFMs, and define a two-dimensional notion of graph domains (topic vs.\ format).

\subsection{Graph Foundation Models}
\label{sec:prelim_gfm}

A GFM is a general purpose graph representation model that follows the foundation model paradigm \cite{achiam2023gpt,touvron2023llama} on graph-structured data. It typically operates in two stages: (i) self-supervised pre-training on a broad multi-domain corpus of graphs, and (ii) cross-domain downstream adaptation to various tasks on graphs. An overview of the GFM pipeline is provided in Fig.~\ref{fig.gfm-pipeline},  Appendix~\ref{app.pipeline}.

Formally, let $\mathcal{G}$ denote a collection of graphs from various domains. We represent a graph $G\in \mathcal{G}$ as $G=(V,E,\mathbf{X}_V,\mathbf{X}_E)$, where $V$ and $E$ are the node and edge sets, and
$\mathbf{X}_V$ and $\mathbf{X}_E$ denote node and edge attributes\footnote{
Minimal notation is used for brevity; full dataset descriptions are in Appendix~\ref{app.datasets}.}, respectively.
A GFM specifies an encoder $\mathtt{GE}(\cdot;\theta):\mathcal{G}\rightarrow\mathbb{R}^{|V|\times d}$ that maps $G$ to $d$-dimensional node representations.
In the pre-training stage, given an \emph{unlabeled} multi-domain graph corpus 
$\mathcal{D}_{\text{pre}}\subset \mathcal{G}$, 
the encoder's parameters are learned by minimizing a self-supervised objective $\mathcal{L}_{\text{pre}}$:
\begin{align}
\theta_\text{pre}
=\arg\min_{\theta}\;
\mathbb{E}_{G\in\mathcal{D}_{\text{pre}}}
\mathcal{L}_{\text{pre}}\big(\mathtt{GE}(G;\theta)\big).
\end{align}
In the downstream stage,  assume task-specific data
$(\mathcal{D}_{\text{down}}, \mathcal{Y}_\text{down})$, where $\mathcal{D}_\text{down}\subset\mathcal{G}$ denotes the task graphs, and   $\mathcal{Y}_\text{down}$ denotes labels associated with the task.  When $\mathcal{Y}_{\text{down}} = \emptyset$, the setting corresponds to zero-shot inference; when $\mathcal{Y}_{\text{down}}$ contains only $K$ labeled instances per class, it corresponds to $K$-shot learning.
% $\mathcal{D}_{\text{down}}=\{\mathcal{G}_\text{down}, \mathcal{Y}_\text{down}\}$, where $\mathcal{G}_\text{down}\subset\mathcal{G}$ denotes the task graphs, and   $\mathcal{Y}_\text{down}$ denotes labels associated with $\mathcal{G}_\text{down}$.  When $\mathcal{Y}_{\text{down}} = \emptyset$, the setting corresponds to zero-shot inference; when $\mathcal{Y}_{\text{down}}$ contains only $K$ labeled instances per class, it corresponds to $K$-shot learning.
In few-shot settings with a small $K$, a lightweight task adapter $f_\phi$ are optimized using a supervised loss $\mathcal{L}_{\text{down}}$, while the pre-trained weights $\theta_\text{pre}$ are frozen, as follows:
\begin{align}
%\phi_\text{down}
%=\arg
\min_{\phi}\;
\mathbb{E}_{(G,Y)\in (\mathcal{D}_{\text{down}},\mathcal{Y}_{\text{down}})}
\mathcal{L}_{\text{down}}\Big(f_{\phi}\big(\mathtt{GE}(G;\theta_\text{pre})\big), Y\Big).
\end{align}

In this benchmark, we only evaluate GFMs that follow the above self-supervised multi-domain pre-training and few-shot downstream adaptation pipeline~\cite{yu2024text,yu2025samgpt,wang2025multi,zhao2024all,zhu2025graphclip,he2025unigraph2,wen2023augmenting,wang2024gft}. Other GFM-style approaches, especially those requiring labeled source-domain data during pre-training, are outside the scope of this benchmark.
%We exclude approaches that require task-level supervision during the pre-training stage \cite{liu2023one,huang2023prodigy,tang2023graphgpt,tang2024higpt,xia2024opengraph,chen2024llaga,li2024zerog,zhao2024graphany, yuangraver}.
%We provide details of the evaluated GFMs, and other evaluated methods, including single domain supervised learning methods, single domain pre-training and adaptation methods, and other representative GFM-style methods in Appendix~\ref{app.baselines}.

\subsection{Domain Composition}
A central design goal of our benchmark is to make graph domain explicit along two largely orthogonal axes: topics and formats. Topic domain describes what a graph represents (semantics), whereas format domain describes how the graph is represented (schema). Table~\ref{table.datasets} summarizes the datasets used in our benchmark, with additional descriptions provided in Appendix~\ref{app.datasets}.

\begin{table}[tbp] 
    \centering
    \footnotesize
    \caption{Summary of the datasets by topics and formats.}
    \vspace{-2mm}
    \label{table.datasets}%
    \resizebox{1\linewidth}{!}{%
    \addtolength{\tabcolsep}{-1mm}
    \begin{tabular}{@{}c|c|rrr|ccccc|c|ccc@{}}
    \toprule
    \multirow{2}{*}{\rotatebox{90}{\makecell{Topic \ \ \ \ \ \ \ \ \ \ \ }}} &\multirow{2}{*}{\makecell{\\\\\\Datasets}} & \multirow{2}{*}{\rotatebox{90}{\makecell{\# Graphs \ \ \ \ \ \ \ \ }}} &\multirow{2}{*}{\rotatebox{90}{\makecell{Avg.~\# nodes\ \ \ \ \ \ }}} &\multirow{2}{*}{\rotatebox{90}{\makecell{Avg.~\# edges\ \ \ \ \ \ }}} &\multicolumn{5}{c|}{Format} &\multirow{2}{*}{\rotatebox{90}{Used for pre-train}} &\multicolumn{3}{c}{\multirow{2}{*}{\makecell{\\\\Task}}}\\
    & &  & &  & \rotatebox{90}{Heterogeneous} & \rotatebox{90}{\makecell{Homophily\\[-0.5mm]ratio}} & \rotatebox{90}{Relational} & \rotatebox{90}{Dynamic} & \rotatebox{90}{Textual} & &\rotatebox{90}{NC} & \rotatebox{90}{EC} &\rotatebox{90}{GC} \\
    \midrule
    \multirow{6}{*}{\rotatebox{90}{Citation}}
    &Cora \cite{liu2023one}  & 1 & 2,708 & 10,556 & $\times$ & 0.81 & $\times$ &$\times$ &$\checkmark$ &$\checkmark$ &$\checkmark$ &$\times$ &$\times$\\
    &Pubmed  \cite{liu2023one}  & 1 & 19,717 & 44,338 & $\times$ & 0.80 & $\times$ &$\times$ &$\checkmark$ &$\times$ &$\checkmark$ &$\times$ &$\times$\\
    &ogbn-arxiv \cite{hu2020open} & 1 & 169,343 & 1,166,243 & $\times$ & 0.66 & $\times$ &$\times$ &$\checkmark$ &$\checkmark$ &$\times$ &$\times$ &$\times$\\
    &ogbn-mag \cite{hu2020open} & 1 & 1,939,743 & 21,111,007 & $\checkmark$ & 0.30 & $\times$ &$\times$ &$\times$ &$\times$ &$\checkmark$ &$\times$ &$\times$\\
    &ACM \cite{tang2024higpt} & 1 & 10,942 & 547,872 & $\checkmark$ & 0.88 & $\times$ &$\times$ &$\times$ &$\checkmark$ &$\checkmark$ &$\times$ &$\times$\\
    &DBLP \cite{wang2019heterogeneous} & 1 & 26,128 & 239,566 & $\checkmark$ & - & $\times$ &$\times$ &$\times$ &$\checkmark$ &$\times$ &$\times$ &$\times$\\\midrule
    
    \multirow{8}{*}{\rotatebox{90}{\makecell{Social \& Web}}}
    &Reddit \cite{yu2024dygprompt} & 1 & 10,984 & 672,447 & $\times$ &- & $\times$ &$\checkmark$ &$\times$ &$\checkmark$ &$\checkmark$ &$\times$ &$\times$\\
    &Wikipedia \cite{yu2024dygprompt} & 1 & 9,227 & 157,474 & $\times$ &- & $\times$ &$\checkmark$ &$\times$ &$\times$ &$\checkmark$ &$\times$ &$\times$\\
    &Actor \cite{li2025hetgb} & 1 & 4,416 & 12,172 & $\times$ & 0.56 & $\times$ &$\times$ &$\checkmark$ &$\times$ &$\checkmark$ &$\times$ &$\times$\\
    &Texas \cite{li2025hetgb} & 1 & 187 & 328 & $\times$ & 0.12 & $\times$ &$\times$ &$\checkmark$ &$\checkmark$ &$\times$ &$\times$ &$\times$\\
    &Wisconsin \cite{li2025hetgb} & 1 & 265 & 530 & $\times$ & 0.20 & $\times$ &$\times$ &$\checkmark$ &$\checkmark$ &$\checkmark$ &$\times$ &$\times$\\
    &Chameleon \cite{rozemberczki2021multi} & 1 & 2,277 & 36,101 & $\times$ & 0.24 & $\times$ &$\times$ &$\times$ &$\times$ &$\checkmark$ &$\times$ &$\times$\\
    &Cornell \cite{li2025hetgb} & 1 & 195 & 304 & $\times$ & 0.13 & $\times$ &$\times$ &$\checkmark$ &$\checkmark$ &$\checkmark$ &$\times$ &$\times$\\
    &IMDB \cite{wang2019heterogeneous} & 1 & 21,420 & 86,642 & $\checkmark$ &- & $\times$ &$\times$ &$\times$ &$\checkmark$ &$\times$ &$\times$ &$\times$\\
    %&Yelp \cite{} & 1 & 1,174,776 & 34,557,089 & $\times$ &not available & $\times$ &$\times$ &$\times$ &$\times$ &$\times$ &$\times$ &$\times$\\
    \midrule
    
    \multirow{5}{*}{\rotatebox{90}{E-commerce}}
    &Photo \cite{yu2024text,yu2025samgpt} & 1 & 7,650 & 238,162 & $\times$ & 0.83 & $\times$ &$\times$ &$\times$ &$\checkmark$ &$\checkmark$ &$\times$ &$\times$\\
    &Computers \cite{yu2024text,yu2025samgpt} & 1 & 13,752 & 491,722 & $\times$ & 0.78 & $\times$ &$\times$ &$\times$ &$\checkmark$ &$\times$ &$\times$ &$\times$\\
    &Amazon \cite{cen2019representation} & 1 & 10,099 & 148,659 & $\checkmark$ &- & $\times$ &$\times$ &$\times$ &$\checkmark$ &$\times$ &$\times$ &$\times$\\
    &Amazon-HeTGB \cite{li2025hetgb} & 1 & 24,492 & 93,050 & $\times$ & 0.38 & $\times$ &$\times$ &$\checkmark$ &$\checkmark$ &$\times$ &$\times$ &$\times$\\
    &Products \cite{chen2024llaga} & 1 & 2,449,029 & 61,859,140 & $\times$ & 0.81 & $\times$ &$\times$ &$\checkmark$ &$\times$ &$\checkmark$ &$\times$ &$\times$\\\midrule
    
    \multirow{4}{*}{\rotatebox{90}{\makecell{Common\\[-1mm] sense}}}
    &Wiki \cite{huang2023prodigy} & 1 & 4,815,483 & 20,624,575 & $\times$ &- & $\checkmark$ &$\times$ &$\times$ &$\times$ &$\times$ &$\checkmark$ &$\times$\\
    &FB15K-237 \cite{liu2023one,wang2024gft,huang2023prodigy} & 1 & 14,505 & 310,079 & $\times$ &- & $\checkmark$ &$\times$ &$\times$ &$\checkmark$ &$\times$ &$\checkmark$ &$\times$\\
    &NELL \cite{huang2023prodigy} & 1 & 65,755 & 234,355 & $\times$ & 0.83 & $\checkmark$ &$\times$ &$\times$ &$\checkmark$ &$\times$ &$\times$ &$\times$\\
    &WN18RR \cite{liu2023one,wang2024gft} & 1 & 40,559 & 92,583 & $\times$ &- & $\checkmark$ &$\times$ &$\times$ &$\times$ &$\times$ &$\checkmark$ &$\times$\\\midrule
    
    \multirow{3}{*}{\rotatebox{90}{Finance}}
    &T-Finanace \cite{tang2022rethinking} & 1 & 39,357 & 21,222,543 & $\times$ & 0.97 & $\times$ &$\times$ &$\times$ &$\times$ &$\checkmark$ &$\times$ &$\times$\\
    &Elliptic \cite{weber2019anti} & 1 & 203,769 & 234,355 & $\times$ & 0.71 & $\times$ &$\times$ &$\times$ &$\checkmark$ &$\checkmark$ &$\times$ &$\times$\\
    &DGraph \cite{huang2022dgraph} & 1 & 3,700,550 & 4,300,999 & $\times$ & 0.37 & $\times$ &$\times$ &$\times$ &$\times$ &$\checkmark$ &$\checkmark$ &$\times$\\\midrule
    \multirow{4}{*}{\rotatebox{90}{Molecule}}
    &HIV \cite{liu2023one} & 41,120 & 25.5 & 27.5 & $\times$ &- & $\times$ &$\times$ &$\checkmark$ &$\checkmark$ &$\times$ &$\times$ &$\checkmark$\\
    &PCBA \cite{liu2023one} & 437,927 & 26.0 & 28.1 & $\times$ &- & $\times$ &$\times$ &$\checkmark$ &$\times$ &$\times$ &$\times$ &$\checkmark$\\
    &COX2 \cite{morris2020tudataset} & 467 & 41.2 & 43.4 & $\times$ &- & $\times$ &$\times$ &$\times$ &$\checkmark$ &$\times$ &$\times$ &$\checkmark$\\\
    &BZR \cite{morris2020tudataset} & 405 & 35.8 & 38.4 & $\times$ &- & $\times$ &$\times$ &$\times$ &$\times$ &$\times$ &$\times$ &$\checkmark$\\\midrule
   
    \multirow{3}{*}{\rotatebox{90}{\makecell{Protein}}}
    &PROTEINS \cite{morris2020tudataset} & 1,113 & 39.06 & 72.82 & $\times$ &- & $\times$ &$\times$ &$\times$ &$\checkmark$ &$\times$ &$\times$ &$\checkmark$\\
    &ENZYMES \cite{morris2020tudataset} & 600 & 32.63 & 62.14 & $\times$ &- & $\times$ &$\times$ &$\times$ &$\checkmark$ &$\times$ &$\times$ &$\times$\\
    &ogbn-proteins \cite{hu2020open} & 1 & 132,534 & 79,122,504 & $\times$ &- & $\times$ &$\times$ &$\times$ &$\times$ &$\checkmark$ &$\times$ &$\times$\\
    %&ogba-ppa \cite{hu2020open}\\
    \bottomrule
    \end{tabular}}\\[1mm]
    \parbox{0.97\linewidth}{\footnotesize 
    The \emph{Used for pre-train} column indicates whether the dataset is used for pre-training. The \emph{Task} column specifies the supported downstream tasks (NC: node classification, EC: edge  classification, and GC: graph classification).}
    \vspace{-3mm}
\end{table}

\stitle{Topic domains.}
We define a topic domain by the semantic meaning of nodes and edges and the typical downstream applications. Thus, shifts in topic domains manifest as changes in node and edge semantics (e.g., papers and citations versus users and interactions), feature distributions, and downstream label semantics and distributions. Our benchmark covers a broad range of common topics in graph datasets, including citation networks, social networks and the Web, e-commerce graphs, financial networks, common-sense graphs, molecular graphs, and protein interaction networks.

\stitle{Format domains.}
%The format domain captures the structural and representational form of a graph and therefore induces different modeling assumptions. 
The format domain captures the schema or representational form of a graph and therefore induces different modeling assumptions over topological patterns, temporal dependencies and attribute or relational schemas.  
Concretely, we consider variations such as homogeneous vs.~heterogeneous graphs, homophilic vs.~ heterophilic graphs, static versus dynamic graphs, relational graphs, and text-attributed graphs. These format differences influence core components of graph learning, such as message passing behavior, temporal dynamics modeling, and relational reasoning.

\stitle{Datasets and composition.}
Topic and format are conceptually orthogonal and can vary independently. Graphs may share topic semantics but adopt different formats. For instance, \textit{Cora}~\cite{liu2023one} and \textit{DBLP}~\cite{wang2022survey} both fall under citation networks, yet \textit{Cora} is typically represented as a homogeneous graph while \textit{DBLP} is heterogeneous with typed nodes and edges. Conversely, graphs may share the same format while originating from different topics: \textit{Actor}~\cite{yu2024non} and \textit{HIV}~\cite{liu2023one} are both homogeneous graphs, but belong to social and molecular domains, respectively.

\section{Benchmarking Protocols}\label{sec.setting}

This section  outlines the benchmarking protocols, including data pre-processing, task construction and data splits, and four evaluation settings designed to probe GFM generalization across topic and format domains.

\stitle{Data pre-processing.}
To enable fair and reproducible comparisons, we standardize all datasets into a unified interface while retaining the semantics required by each task and graph format.
1) \textit{Graph canonicalization.} We remove duplicate edges. For node- and graph-level tasks, we treat graphs as undirected by symmetrizing edges for message passing. For edge-level classification, where $u\!\rightarrow\!v$ and $v\!\rightarrow\!u$ have different labels, we preserve directionality. % and use directed graphs.
2) \textit{Heterogeneous graphs.} We keep node/edge types. % and relation identifiers. 
Methods that natively support heterogeneity take the original typed graphs as input. For methods that assume homogeneous graphs, we convert the graph to a homogeneous form and encode type information as auxiliary textual attributes (with node/edge type tokens).
3)  \textit{Dynamic graphs.} Dynamic graph models operate on the original temporal graphs. For static graph models, we use the graph at the final timestamp as input.
%construct a snapshot by taking the graph at the final timestamp and use it as input.
4) \textit{Relational graphs.} When node features are provided, we use them directly. Otherwise, we assign each node a unique identifier represented as a one-hot vector. For non-relation-aware methods, relation information is not leveraged, as they do not utilize edge attributes.
%, avoiding the introduction of external features.
5) \textit{Text-attributed graphs.} We keep raw text fields unchanged. For methods with textual encoders, tokenization and text embeddings follow the authors' implementations. For text-free methods, we do not use textual content and rely on the numeric node/edge features.
Additional dataset-specific pre-processing details are provided in Appendix~\ref{app.pre-process}.

\stitle{Downstream tasks and data splits.}
We evaluate GFMs on three standard graph learning tasks: node classification, edge classification, and graph classification. The datasets used for pre-training are \emph{unlabeled} and restricted to those indicated in Table~\ref{table.datasets}. For downstream tasks, we adopt few-shot evaluation for each task type. Following the standard $K$-shot protocol~\cite{liu2023graphprompt,yu2023generalized}, we randomly split each downstream dataset into a \emph{training pool} (20\%) and a \emph{test set} (80\%). A $K$-shot task is then constructed by sampling $K$ labeled instances per class from the training pool. We repeat this procedure 50 times to form 50 few-shot episodes per dataset. For each episode, we repeat with five random seeds, resulting in $50\times 5 = 250$ runs per dataset and per shot setting. We report the mean and standard deviation of Accuracy and Macro-F1,
%Accuracy~\cite{yu2025gcot} and Macro-F1~\cite{yu2023hgprompt}. 
benchmarking one- and five-shot settings (i.e., $K=1$ or $5$). 
More details are provided in Appendix~\ref{app.data-split}.

\begin{table*}[tbp] % [!t]
    \centering
    \footnotesize
    \caption{Accuracy of one-shot node classification on unseen datasets. Marco-F1 reported in Appendix~\ref{app.exp1}.}
    \vspace{-2mm}
    \addtolength{\tabcolsep}{0.5mm}
    \label{table.exp1-1shotnc-acc}
    %\resizebox{1\linewidth}{!}{
    \begin{tabular}{l|ccccccccc}
    \toprule
    {Methods} & {Pubmed} & {ogbn-mag} & {Wikipedia} & {Actor}  & {Chameleon} & {Products} & {T-Finance} & {DGraph} & {ogbn-proteins}\\
    \midrule\midrule
    \method{GCN} &48.62{\tiny$\pm$8.40} &\underline{5.83}{\tiny$\pm$0.70} &48.02{\tiny$\pm$39.11}  &26.50{\tiny$\pm$11.82}  &26.71{\tiny$\pm$4.33} &21.12{\tiny$\pm$5.67} &42.16{\tiny$\pm$39.45} &32.88{\tiny$\pm$15.35} &\underline{51.54}{\tiny$\pm$13.11}\\
    \method{GAT} &45.50{\tiny$\pm$9.29} &4.86{\tiny$\pm$0.60} &45.32{\tiny$\pm$24.72} &23.90{\tiny$\pm$11.13} &26.24{\tiny$\pm$4.27} &22.11{\tiny$\pm$6.28} &53.86{\tiny$\pm$38.15}&33.46{\tiny$\pm$16.12}&47.93{\tiny$\pm$14.10}\\\midrule
    \method{GCOPE}&41.85{\tiny$\pm$8.34} &0.27{\tiny$\pm$0.09} &43.23{\tiny$\pm$25.98} &24.16{\tiny$\pm$6.83} &21.66{\tiny$\pm$3.41} &OOT &45.42{\tiny$\pm$22.22} &OOT &45.01{\tiny$\pm$12.73} \\
    \method{MDGPT}&40.50{\tiny$\pm$7.33} &\textbf{5.94}{\tiny$\pm$0.63} &51.02{\tiny$\pm$25.40} &\textbf{31.72}{\tiny$\pm$8.93} &\underline{28.36}{\tiny$\pm$5.04} &15.99{\tiny$\pm$3.48} &\underline{58.95}{\tiny$\pm$23.03} &37.52{\tiny$\pm$12.02} &50.72{\tiny$\pm$10.78}\\
    \method{MDGFM} &39.50{\tiny$\pm$7.20} &2.93{\tiny$\pm$1.05} &44.25{\tiny$\pm$28.21} &19.80{\tiny$\pm$7.00} &25.95{\tiny$\pm$3.75} &8.11{\tiny$\pm$2.17} &57.07{\tiny$\pm$19.12} &36.08{\tiny$\pm$24.19} &50.28{\tiny$\pm$8.80}\\
    \method{SAMGPT} &\underline{49.23}{\tiny$\pm$8.07} &5.32{\tiny$\pm$0.68} &\textbf{56.50}{\tiny$\pm$27.26} &27.09{\tiny$\pm$7.73} &\textbf{28.65}{\tiny$\pm$5.12} &\textbf{22.68}{\tiny$\pm$4.03} &58.28{\tiny$\pm$15.86} &\underline{39.42}{\tiny$\pm$13.36} &\textbf{52.25}{\tiny$\pm$11.97}\\
    \method{G2P2} &48.56{\tiny$\pm$6.77} &- &38.62{\tiny$\pm$10.45} &\underline{29.91}{\tiny$\pm$6.81} &- &- &\textbf{63.04}{\tiny$\pm$18.64} &38.30{\tiny$\pm$10.71} &-\\
    \method{GraphCLIP} &36.67{\tiny$\pm$4.26} &- &34.67{\tiny$\pm$21.87} &25.18{\tiny$\pm$5.87} &- &- &44.77{\tiny$\pm$21.85} &\textbf{39.63}{\tiny$\pm$26.42} &-\\
    \method{GFT} &\textbf{53.56}{\tiny$\pm$10.27} &1.50{\tiny$\pm$0.29} &\underline{55.00}{\tiny$\pm$17.19} &25.69{\tiny$\pm$5.93} &27.32{\tiny$\pm$4.08} &11.36{\tiny$\pm$4.49} &48.43{\tiny$\pm$23.94} &35.24{\tiny$\pm$12.68} &51.03{\tiny$\pm$7.93}\\
    \method{UniGraph2} &45.27{\tiny$\pm$9.41} &2.21{\tiny$\pm$1.19} &52.23{\tiny$\pm$30.37} &27.35{\tiny$\pm$9.23} &26.02{\tiny$\pm$5.71} &\underline{22.61}{\tiny$\pm$4.70} &58.59{\tiny$\pm$22.94} &36.37{\tiny$\pm$13.57} &50.06{\tiny$\pm$6.42}\\\bottomrule
    \end{tabular}%}
    \\[1mm]
    \parbox{0.86\linewidth}{\footnotesize OOT: out-of-time, downstream execution exceeding three days. OOM: GPU out-of-memory. \method{G2P2} and \method{GraphCLIP} require textual labels for downstream tasks and are marked as `-' when unavailable. Best/runner-up results are in bold/underlined. \textbf{The same notations apply to subsequent tables.}}
\end{table*}

\begin{table*}[tbp] % [!t]
    \centering
    \footnotesize
    \caption{Evaluation of one-shot edge classification and graph classification on unseen datasets.}
    \vspace{-2mm}
    \addtolength{\tabcolsep}{0.2mm}
    \label{table.exp1-1shotecgc}
 %   \resizebox{1\linewidth}{!}{
    \begin{tabular}{l|cccccc|cccc}
    \toprule
    \multirow{3}{*}{Methods} & \multicolumn{6}{c|}{Edge classification} & \multicolumn{4}{c}{Graph classification}\\
     &\multicolumn{2}{c}{DGraph} &\multicolumn{2}{c}{Wiki} &\multicolumn{2}{c|}{WN18RR} &\multicolumn{2}{c}{PCBA} &\multicolumn{2}{c}{BZR} \\
    &Acc &MacroF &Acc &MacroF &Acc &MacroF &Acc &MacroF &Acc &MacroF\\
    \midrule\midrule
    \method{GCN} &\textbf{10.13}{\tiny$\pm$3.16} &6.50{\tiny$\pm$0.99} &9.35{\tiny$\pm$3.01} &3.21{\tiny$\pm$0.19} &\underline{15.45}{\tiny$\pm$11.60} &10.85{\tiny$\pm$4.39} &55.09{\tiny$\pm$31.92} &32.54{\tiny$\pm$15.72} &52.73{\tiny$\pm$16.49} &44.72{\tiny$\pm$10.30}\\
    \method{GAT} &9.55{\tiny$\pm$3.56} &6.16{\tiny$\pm$1.05} &2.65{\tiny$\pm$0.62} &1.71{\tiny$\pm$0.15} &10.60{\tiny$\pm$14.10} &7.91{\tiny$\pm$3.93} &\underline{56.65}{\tiny$\pm$28.70} &33.97{\tiny$\pm$13.34} &52.36{\tiny$\pm$17.69} &43.63{\tiny$\pm$10.93}\\\midrule
    \method{GCOPE} &OOT &OOT &OOT &OOT &\textbf{16.59}{\tiny$\pm$7.25} &\textbf{11.42}{\tiny$\pm$3.29} &48.99{\tiny$\pm$38.52} &28.19{\tiny$\pm$19.08} &54.31{\tiny$\pm$19.35} &45.89{\tiny$\pm$12.33}\\
    \method{MDGPT} &9.92{\tiny$\pm$2.38} &\underline{6.85}{\tiny$\pm$0.95} &10.72{\tiny$\pm$3.67} &2.79{\tiny$\pm$0.21} &13.16{\tiny$\pm$4.71} &10.08{\tiny$\pm$2.01} &55.54{\tiny$\pm$21.39} &\underline{34.57}{\tiny$\pm$9.90} &53.88{\tiny$\pm$13.74} &\underline{47.26}{\tiny$\pm$9.46}\\
    \method{MDGFM} &8.37{\tiny$\pm$2.17} &5.86{\tiny$\pm$0.78} &\underline{11.04}{\tiny$\pm$2.95} &\underline{3.49}{\tiny$\pm$0.33} &13.72{\tiny$\pm$3.11} &10.32{\tiny$\pm$1.54} &\textbf{59.16}{\tiny$\pm$40.50} &32.23{\tiny$\pm$20.18} &44.55{\tiny$\pm$25.60} &31.17{\tiny$\pm$12.87}\\
    \method{SAMGPT} &\underline{10.07}{\tiny$\pm$2.52} &\textbf{6.87}{\tiny$\pm$0.94} &\textbf{11.45}{\tiny$\pm$3.31} &\textbf{3.58}{\tiny$\pm$0.20} &13.07{\tiny$\pm$3.67} &10.16{\tiny$\pm$1.75} &56.38{\tiny$\pm$23.72} &\textbf{34.63}{\tiny$\pm$10.98} &\textbf{56.06}{\tiny$\pm$17.10} &\textbf{48.07}{\tiny$\pm$10.92}\\
    \method{G2P2} &- &- &- &- &12.62{\tiny$\pm$3.51} &9.16{\tiny$\pm$1.59} &- &- &53.30{\tiny$\pm$13.91} &46.78{\tiny$\pm$8.22}\\
    \method{GraphCLIP} &- &- &- &- &15.44{\tiny$\pm$4.60} &\underline{11.20}{\tiny$\pm$2.04} &- &- &\underline{54.47}{\tiny$\pm$15.52} &44.67{\tiny$\pm$8.77}\\
    \method{GFT} &9.28{\tiny$\pm$3.25} &6.50{\tiny$\pm$1.12} &OOT &OOT &12.66{\tiny$\pm$6.67} &9.96{\tiny$\pm$1.64} &52.55{\tiny$\pm$31.20} &34.25{\tiny$\pm$16.59} &43.67{\tiny$\pm$24.29} &33.56{\tiny$\pm$15.81}\\
    \method{UniGraph2} &9.63{\tiny$\pm$3.75} &5.55{\tiny$\pm$1.01} &10.10{\tiny$\pm$0.21} &3.00{\tiny$\pm$0.16} &10.33{\tiny$\pm$3.76} &6.92{\tiny$\pm$1.69} &50.34{\tiny$\pm$9.42} &33.04{\tiny$\pm$4.36} &51.57{\tiny$\pm$16.08} &43.79{\tiny$\pm$9.16}\\\bottomrule
    \end{tabular}%}
\end{table*}

\stitle{Evaluation settings.}
We design four evaluation settings to probe GFM generalization under seen and unseen data, as well as topic and format shifts (Fig.~\ref{fig.ben-pipeline}, Appendix~\ref{app.ben-pipeline}).

\stitle{\textit{Setting I: Pre-train on diverse topics and formats; adapt to unseen datasets.}}
We construct $\mathcal{D}_{\text{pre}}$ by selecting unlabeled datasets spanning a broad range of topic and format domains, and hold out a disjoint set $\mathcal{D}_{\text{unseen}}$ such that no graph in $\mathcal{D}_{\text{down}}$ is used during pre-training, i.e., $\mathcal{D}_{\text{down}} = \mathcal{D}_{\text{unseen}}$ and $\mathcal{D}_{\text{pre}} \cap \mathcal{D}_{\text{unseen}} = \emptyset$.
% \begin{equation}
% \mathcal{D}_{\text{down}} \;=\; \mathcal{D}_{\text{unseen}}, \qquad
% \mathcal{D}_{\text{pre}} \cap \mathcal{D}_{\text{unseen}} = \emptyset.
% \end{equation}
This setting evaluates extrapolative transfer to datasets not encountered during pre-training. Concretely, we use all those indicated under \emph{Used for pre-train} column in Table~\ref{table.datasets} as $\mathcal{D}_{\text{pre}}$, and treat the remaining datasets as $\mathcal{D}_{\text{unseen}}$.

\stitle{\textit{Setting II: Pre-train on diverse topics and formats; adapt to seen datasets.}}
In this setting, downstream evaluation is performed on the same datasets used for pre-training, i.e., $\mathcal{D}_{\text{down}} = \mathcal{D}_{\text{pre}}$, while task labels $\mathcal{Y}_{\text{down}}$ are available only at the downstream stage.
%\begin{equation}
%\mathcal{D}_{\text{down}} \;=\; \mathcal{D}_{\text{pre}}.
%\end{equation}
This setting serves as an interpolation reference, assessing the extent to which self-supervised pre-training is beneficial when the label-free downstream distribution is already covered during pre-training. Here, we adopt the same $\mathcal{D}_{\text{pre}}$ as in Setting~I.

\stitle{\textit{Setting III: Pre-train on a single topic domain; adapt to other topics.}}
To isolate semantic generalization, we restrict pre-training to citation networks and evaluate transfer to datasets in other topic domains.
This setting enables a fine-grained analysis of downstream adaptation as a function of semantic proximity between topics. For instance, citation and social graphs both model human activities and may share more transferable patterns than molecular graphs, which arise from distinct generative processes.
Specifically, we use \textit{Cora}, \textit{ACM}, and \textit{DBLP} for pre-training. Downstream evaluation includes both same-topic datasets (\textit{Pubmed}, \textit{ogbn-mag}) and out-of-topic datasets (\textit{Wikipedia}, \textit{Chameleon}, \textit{Actor}, \textit{Products}, \textit{ogbn-proteins}, \textit{T-Finance}, \textit{DGraph}, \textit{BZR}, \textit{Wiki}, \textit{WN18RR}).

\stitle{\textit{Setting IV: Pre-train on a base format; adapt to other formats.}}
To isolate format generalization, we pre-train on datasets from a canonical ``base'' graph format, corresponding to homogeneous, homophilic, static, single-relation, text-free graphs as specified in Table~\ref{table.datasets}. We then evaluate on datasets in other formats.
This setting directly probes whether representations learned under a canonical format remain effective when transferred to heterogeneous, dynamic, relational, or text-attributed graphs.
Particularly, we use \textit{Photo}, \textit{Computer}, \textit{COX2}, \textit{Protein}, \textit{ENZYMES}, \textit{Elliptic} in the base format for pre-training, and evaluate downstream performance on datasets in diverse formats, including \textit{Pubmed}, \textit{Products}, \textit{Actor}, \textit{Wikipedia}, \textit{Chameleon}, \textit{ogbn-mag}, \textit{PCBA}, \textit{Wiki}, \textit{WN18RR}.

\stitle{Implementation details.}
We use the authors' official open-source implementations whenever available. Hyperparameters are tuned within the search spaces recommended in the original papers.
%based on training curves. 
Full implementation details and hyperparameter configurations are deferred to Appendix~\ref{app.hyperparams}.
\section{Empirical Results and Analysis}
We present and analyze the evaluation results across the four settings under the proposed benchmarking protocols. Each setting addresses a distinct research question (RQ).

\subsection{Setting I: Adapting to Unseen Datasets}\label{sec.exp1}

%\begin{rqbox}
\noindent
\textbf{RQ1:} \textit{After multi-domain pre-training over diverse topic and format domains, can GFMs reliably adapt to unseen downstream datasets?}
%\end{rqbox}

We report one-shot classification performance in Tables~\ref{table.exp1-1shotnc-acc} and \ref{table.exp1-1shotecgc}. 
Besides GFMs, we include conventional supervised GNNs (GCN and GAT) as baselines, which are trained directly on downstream data without pre-training. Note that conventional graph pre-training methods that pre-train and evaluate on the same dataset are not applicable to the unseen downstream setting considered here.
Additional results, such as Macro-F1 for node classification, five-shot performance, and efficiency, are provided in Appendix~\ref{app.exp1}.

%We evaluate -shot adaptation on unseen datasets for node classification and report Accuracy in Table~\ref{table.exp1-1shotnc-acc}. We further report 1-shot results for edge and graph classification in Table~\ref{table.exp1-1shotecgc}. Additional results, including 1-shot Macro-F1 for node classification, 5-shot performance for node/edge/graph tasks, and downstream adaptation tuning time, are provided in Appendix~\ref{app.exp1}. 
%Methods that exceed one day of runtime are marked as \textit{OOT}, and those that run out of GPU memory as \textit{OOM}. For \method{G2P2} and \method{GraphCLIP}, downstream adaptation requires label text; therefore, these methods are not applicable to datasets without label text and are marked as ``-''.

\stitle{Results.}\label{sec:unseen}
% \textit{(i) No single GFM dominates across unseen datasets.}
We first observe that
no single GFM dominates across unseen datasets.
Depending on the dataset and task, \method{SAMGPT}, \method{MDGPT}, \method{GFT} and \method{MDGFM} are among the most competitive methods. When textual labels are available, \method{G2P2} and \method{GraphCLIP} tend to perform well due to their explicit use of textual labels for classification. For instance, \method{G2P2} excels on \textit{T-Finance} and \method{GraphCLIP} leads on \textit{DGraph} in node classification.

%The top method changes with the downstream dataset: \method{SAMGPT} leads on several targets (e.g., \textit{Wikipedia}, \textit{Chameleon}, \textit{Products}, and \textit{ogbn-proteins}), \method{MDGPT} is strongest on \textit{Actor} and \textit{ogbn-mag}, and \method{GFT} performs best on \textit{Pubmed}. When label text is available, text-assisted approaches can be particularly competitive, with \method{G2P2} leading on \textit{T-Finance} and \method{GraphCLIP} on \textit{DGraph}.
% \textit{(ii) Improvements over supervised GNN baselines are inconsistent.}
Second, while GFMs largely outperform conventional supervised GNNs, the improvements are inconsistent.
On certain datasets (e.g., \textit{DGraph} and \textit{T-Finance}), GFM yields clear improvements over GCN and GAT. However, on some datasets, improvements are limited, with supervised GNNs trained from scratch remaining competitive (e.g., \textit{ogbn-mag}). Therefore, broad multi-domain pre-training does not automatically translate into reliable dataset-level transfer.

% \stitle{Results on unseen edge/graph classification.}
% Table~\ref{table.exp1-1shotecgc} suggests that these issues are not specific to node-level tasks. Generalization to unseen datasets remains both method- and task-dependent: relative rankings can change when moving from node- to edge- or graph-level tasks, and different models become preferable on different datasets. Similarly, performance improvements over GCN or GAT vary. 
%This cautions against extrapolating robustness from node classification alone.

\stitle{Findings and insights.}
\emph{Current GFMs demonstrate promising but uneven performance on unseen datasets, highlighting the need for improved multi-domain knowledge integration and transfer.}
Compared with conventional supervised GNNs, the best-performing GFMs yield clear improvements on most unseen targets, though the benefits are not extended across all GFMs or datasets. This suggests that further progress may require reexamining how multi-domain information is integrated during pre-training, as well as developing more effective adaptation strategies for transferring multi-domain knowledge to unseen targets.
%\textit{Current GFMs do not provide consistent improvements on unseen datasets.}
%Relative to conventional supervised GNNs, GFMs can yield clear gains on some unseen targets, but the advantage is not always even: on several datasets it narrows substantially or vanishes. This suggests that current GFMs is not sufficient to ensure robust dataset transfer across domains to unseen datasets.

% \textit{(ii) Different designs help in different failure modes.}
% Structure-aware mechanisms appear more consistently competitive on structurally atypical graphs, whereas text-assisted methods can excel when text is available. 

\begin{table*}[tbp] % [!t]
    \centering
    \footnotesize
    \addtolength{\tabcolsep}{-0.3mm}
    \caption{Evaluation of one-shot node classification on seen datasets.}
    \vspace{-2mm}
    \label{table.exp2-1shotnc}
    %\resizebox{1\linewidth}{!}{
    \begin{tabular}{l|cccccccccccc}
    \toprule
    \multirow{2}{*}{Methods} & \multicolumn{2}{c}{Cora} & \multicolumn{2}{c}{ACM} & \multicolumn{2}{c}{Reddit} & \multicolumn{2}{c}{Wisconsin}  & \multicolumn{2}{c}{Photo} & \multicolumn{2}{c}{Elliptic} \\
    &Acc &MacroF &Acc &MacroF &Acc &MacroF &Acc &MacroF &Acc &MacroF &Acc &MacroF\\
    \midrule\midrule
    \method{GCN} &41.76{\tiny$\pm$10.38} &37.85{\tiny$\pm$9.98} &38.31{\tiny$\pm$8.08} &24.12{\tiny$\pm$10.36} &50.90{\tiny$\pm$33.56} &32.75{\tiny$\pm$16.73} &34.20{\tiny$\pm$7.94} &22.49{\tiny$\pm$4.55} &53.87{\tiny$\pm$12.35} &52.05{\tiny$\pm$11.90} &41.74{\tiny$\pm$18.55} &32.74{\tiny$\pm$11.50}\\
    \method{GAT} &43.48{\tiny$\pm$9.64} &39.92{\tiny$\pm$8.96} &35.82{\tiny$\pm$5.75} &22.37{\tiny$\pm$8.94} &48.68{\tiny$\pm$30.94} &32.30{\tiny$\pm$15.74} &34.18{\tiny$\pm$8.57} &23.06{\tiny$\pm$5.94} &46.38{\tiny$\pm$11.11} &45.49{\tiny$\pm$9.65} &38.73{\tiny$\pm$20.26} &29.95{\tiny$\pm$11.87}\\
 %   \method{Simple-HGN} &- &- &39.52{\tiny$\pm$9.01} &30.74{\tiny$\pm$12.82} &- &- &- &- &- &- &- &-\\
 %   \method{FAGCN} &- &- &- &- &- &- &\underline{36.97}{\tiny$\pm$11.25} &27.33{\tiny$\pm$7.95} &- &- &- &-\\
 \midrule
    \method{DGI} &43.82{\tiny$\pm$9.33} &40.78{\tiny$\pm$8.06} &36.01{\tiny$\pm$7.24} &28.11{\tiny$\pm$9.48} &46.92{\tiny$\pm$20.14} &33.27{\tiny$\pm$9.98} &27.29{\tiny$\pm$8.32} &21.34{\tiny$\pm$5.51} &47.26{\tiny$\pm$7.47} &45.23{\tiny$\pm$6.57} &44.56{\tiny$\pm$15.54} &34.25{\tiny$\pm$10.64}\\
    \method{GraphPrompt} &\textbf{55.51}{\tiny$\pm$10.51} &\textbf{52.53}{\tiny$\pm$9.76} &42.18{\tiny$\pm$9.78} &35.12{\tiny$\pm$12.51} &48.32{\tiny$\pm$14.02} &35.11{\tiny$\pm$6.97} &32.73{\tiny$\pm$4.80} &26.41{\tiny$\pm$3.88} &\underline{57.06}{\tiny$\pm$9.09} &\underline{55.81}{\tiny$\pm$7.99} &\textbf{50.55}{\tiny$\pm$17.03} &\textbf{39.50}{\tiny$\pm$13.23}\\
    \method{HeCo} &/ &/ &\textbf{80.85}{\tiny$\pm$12.79} &\textbf{80.25}{\tiny$\pm$13.63} &/ &/ &/ &/ &/ &/ &/ &/\\
    \method{TGN} &/ &/ &/ &/ &47.98{\tiny$\pm$11.60} &34.93{\tiny$\pm$5.45} &/ &/ &/ &/ &/ &/\\
    \method{DDGCL} &/ &/ &/ &/ &54.12{\tiny$\pm$18.92} &\underline{36.93}{\tiny$\pm$8.88} &/ &/ &/ &/ &/ &/\\
    \method{DSSL} &/ &/ &/ &/ &/ &/ &31.34{\tiny$\pm$6.95} &24.25{\tiny$\pm$4.12} &/ &/ &/ &/\\\midrule
    \method{GCOPE} &42.56{\tiny$\pm$7.22} &42.02{\tiny$\pm$7.33} &33.21{\tiny$\pm$1.27} &28.50{\tiny$\pm$3.89} &25.22{\tiny$\pm$25.68} &18.86{\tiny$\pm$14.39} &18.41{\tiny$\pm$6.01} &15.92{\tiny$\pm$4.16} &35.66{\tiny$\pm$6.70} &30.32{\tiny$\pm$5.55} &41.33{\tiny$\pm$16.67} &30.51{\tiny$\pm$7.98}\\
    \method{MDGPT} &29.41{\tiny$\pm$6.30} &28.09{\tiny$\pm$6.01} &52.81{\tiny$\pm$9.25} &51.01{\tiny$\pm$9.68} &43.38{\tiny$\pm$24.31} &30.76{\tiny$\pm$12.58} &27.53{\tiny$\pm$6.27} &22.86{\tiny$\pm$3.96} &44.56{\tiny$\pm$8.15} &43.01{\tiny$\pm$7.41} &44.48{\tiny$\pm$14.06} &34.13{\tiny$\pm$7.75}\\
    \method{MDGFM} &28.15{\tiny$\pm$7.26} &26.76{\tiny$\pm$6.98} &43.86{\tiny$\pm$11.66} &42.93{\tiny$\pm$11.86} &\textbf{55.73}{\tiny$\pm$30.46} &35.46{\tiny$\pm$15.50} &23.38{\tiny$\pm$8.39} &18.93{\tiny$\pm$6.01} &24.76{\tiny$\pm$8.67} &23.78{\tiny$\pm$9.23} &47.31{\tiny$\pm$21.01} &34.24{\tiny$\pm$11.53}\\
    \method{SAMGPT} &44.16{\tiny$\pm$6.11} &42.35{\tiny$\pm$6.21} &52.88{\tiny$\pm$9.70} &49.94{\tiny$\pm$11.84} &\underline{54.43}{\tiny$\pm$22.11} &36.80{\tiny$\pm$10.62} &32.81{\tiny$\pm$9.09} &26.16{\tiny$\pm$5.34} &55.48{\tiny$\pm$8.82} &54.41{\tiny$\pm$7.65} &\underline{47.35}{\tiny$\pm$12.75} &\underline{37.08}{\tiny$\pm$8.36}\\
    \method{G2P2} &45.12{\tiny$\pm$4.84} &\underline{44.63}{\tiny$\pm$4.40} &53.75{\tiny$\pm$12.77} &49.32{\tiny$\pm$15.61} &44.55{\tiny$\pm$14.27} &22.31{\tiny$\pm$5.03} &\underline{34.44}{\tiny$\pm$9.05} &\underline{27.48}{\tiny$\pm$5.38} &- &- &36.14{\tiny$\pm$6.16} &25.53{\tiny$\pm$3.24}\\
    \method{GraphCLIP} &34.46{\tiny$\pm$5.51} &32.40{\tiny$\pm$4.87} &27.66{\tiny$\pm$5.17} &20.17{\tiny$\pm$2.74} &31.46{\tiny$\pm$17.08} &16.78{\tiny$\pm$7.25} &29.70{\tiny$\pm$7.39} &25.40{\tiny$\pm$4.97} &- &- &39.72{\tiny$\pm$20.55} &23.45{\tiny$\pm$8.34}\\
    \method{GFT} &44.39{\tiny$\pm$6.52} &42.65{\tiny$\pm$6.03} &45.17{\tiny$\pm$5.90} &44.18{\tiny$\pm$5.92} &53.12{\tiny$\pm$11.90} &\textbf{37.33}{\tiny$\pm$5.56} &\textbf{38.96}{\tiny$\pm$7.47} &\textbf{32.09}{\tiny$\pm$5.68} &40.35{\tiny$\pm$6.01} &40.06{\tiny$\pm$5.54} &40.51{\tiny$\pm$16.07} &32.17{\tiny$\pm$9.12}\\
    \method{UniGraph2} &\underline{46.53}{\tiny$\pm$8.79} &43.73{\tiny$\pm$7.16} &\underline{60.82}{\tiny$\pm$8.36} &\underline{57.60}{\tiny$\pm$9.45} &45.91{\tiny$\pm$23.93} &29.62{\tiny$\pm$11.63} &25.92{\tiny$\pm$8.21} &19.24{\tiny$\pm$4.63} &\textbf{63.00}{\tiny$\pm$10.09} &\textbf{62.20}{\tiny$\pm$8.60} &44.36{\tiny$\pm$11.05} &34.27{\tiny$\pm$5.57}\\\bottomrule
    \end{tabular}%}
    \\[1mm]
    \parbox{0.98\linewidth}{\footnotesize Format-specific methods, including \method{HeCo}, \method{TGN}, \method{DDGCL}, and \method{DSSL}, are evaluated only on graphs of their intended formats and are maked with `/' for all other formats.}
\end{table*}

\begin{table}[tbp] % [!t]
    \centering
    \footnotesize
    \caption{Evaluation of one-shot edge classification and graph classification on seen datasets.}
    \vspace{-2mm}
    \addtolength{\tabcolsep}{-1.3mm}
    \label{table.exp2-1shotecgc}
    \resizebox{1\linewidth}{!}{
    \begin{tabular}{@{}l|cc|cccccc@{}}
    \toprule
    \multirow{3}{*}{Methods} & \multicolumn{2}{c|}{Edge classification} & \multicolumn{6}{c}{Graph classification}\\
     &\multicolumn{2}{c|}{FB15K-237} &\multicolumn{2}{c}{HIV} &\multicolumn{2}{c}{COX2} &\multicolumn{2}{c}{PROTEINS} \\
    &Acc &MacroF &Acc &MacroF &Acc &MacroF &Acc &MacroF \\
    \midrule\midrule
    \method{GCN} &14.63{\tiny$\pm$1.4} &\underline{13.35}{\tiny$\pm$0.8} &45.79{\tiny$\pm$27.1} &31.22{\tiny$\pm$13.8} &48.09{\tiny$\pm$14.2} &42.15{\tiny$\pm$9.5} &\textbf{58.67}{\tiny$\pm$9.2} &52.33{\tiny$\pm$8.6}\\
    \method{GAT} &11.05{\tiny$\pm$1.2} &10.90{\tiny$\pm$0.8} &46.00{\tiny$\pm$27.4} &31.50{\tiny$\pm$14.4} &50.71{\tiny$\pm$14.9} &43.67{\tiny$\pm$9.1} &57.79{\tiny$\pm$9.4} &51.44{\tiny$\pm$8.7}\\
    %\method{Simple-HGN} \\
    %\method{TGN} \\
    %\method{FAGFN} \\
    \midrule
    \method{DGI} &4.84{\tiny$\pm$1.4} &1.61{\tiny$\pm$0.3} &48.47{\tiny$\pm$13.2} &34.93{\tiny$\pm$6.4} &46.99{\tiny$\pm$12.8} &41.99{\tiny$\pm$8.3} &54.28{\tiny$\pm$8.7} &51.56{\tiny$\pm$8.1}\\
    %\method{HeCo} \\
    %\method{DDGCL} \\
    %\method{DSSL} \\
    \method{\scriptsize GraphPrompt} &5.52{\tiny$\pm$1.6} &1.42{\tiny$\pm$0.3} &\underline{50.50}{\tiny$\pm$9.3} &\textbf{36.15}{\tiny$\pm$4.2} &49.30{\tiny$\pm$10.9} &44.18{\tiny$\pm$7.1} &56.83{\tiny$\pm$7.0} &\underline{54.32}{\tiny$\pm$6.3}\\\midrule
    \method{GCOPE} &1.51{\tiny$\pm$0.4} &1.01{\tiny$\pm$0.1} &\textbf{54.20}{\tiny$\pm$34.7} &33.51{\tiny$\pm$16.2} &39.04{\tiny$\pm$13.6} &35.05{\tiny$\pm$9.3} &53.27{\tiny$\pm$9.5} &50.48{\tiny$\pm$9.9}\\
    \method{MDGPT} &\underline{15.56}{\tiny$\pm$1.2} &12.52{\tiny$\pm$0.7} &47.51{\tiny$\pm$15.2} &34.26{\tiny$\pm$8.3} &53.06{\tiny$\pm$12.2} &45.75{\tiny$\pm$6.7} &55.51{\tiny$\pm$9.9} &48.87{\tiny$\pm$8.8}\\
    \method{MDGFM} &12.26{\tiny$\pm$1.2} &10.28{\tiny$\pm$1.0} &42.45{\tiny$\pm$29.9} &28.78{\tiny$\pm$15.4} &\underline{54.47}{\tiny$\pm$22.8} &38.67{\tiny$\pm$11.7} &56.00{\tiny$\pm$9.0} &49.44{\tiny$\pm$9.9}\\
    \method{SAMGPT} &14.91{\tiny$\pm$1.5} &12.05{\tiny$\pm$1.1} &46.21{\tiny$\pm$17.8} &33.43{\tiny$\pm$9.6} &\textbf{55.53}{\tiny$\pm$12.0} &\textbf{47.63}{\tiny$\pm$7.1} &\underline{58.02}{\tiny$\pm$8.6} &52.25{\tiny$\pm$7.7}\\
    \method{G2P2} &\textbf{17.26}{\tiny$\pm$1.4} &\textbf{14.23}{\tiny$\pm$0.8} &49.36{\tiny$\pm$15.3} &\underline{35.30}{\tiny$\pm$7.7} &52.35{\tiny$\pm$13.1} &\underline{45.84}{\tiny$\pm$8.8} &53.48{\tiny$\pm$9.3} &49.26{\tiny$\pm$9.6}\\
    \method{GraphCLIP} &2.68{\tiny$\pm$0.5} &1.31{\tiny$\pm$0.2} &47.77{\tiny$\pm$20.4} &33.69{\tiny$\pm$9.9} &49.16{\tiny$\pm$13.3} &42.66{\tiny$\pm$7.6} &57.00{\tiny$\pm$7.6} &\textbf{55.26}{\tiny$\pm$7.6}\\
    \method{GFT} &2.00{\tiny$\pm$0.4} &1.62{\tiny$\pm$0.2} &46.61{\tiny$\pm$34.7} &29.30{\tiny$\pm$18.6} &49.01{\tiny$\pm$21.4} &37.45{\tiny$\pm$13.1} &51.52{\tiny$\pm$10.1} &44.56{\tiny$\pm$9.5}\\
    \method{UniGraph2} &1.41{\tiny$\pm$0.7} &1.02{\tiny$\pm$0.0} &48.56{\tiny$\pm$13.7} &32.11{\tiny$\pm$6.5} &49.14{\tiny$\pm$15.8} &41.65{\tiny$\pm$9.1} &52.30{\tiny$\pm$10.7} &48.56{\tiny$\pm$9.3}\\\bottomrule
    \end{tabular}}
\end{table}

\subsection{Setting II: Adapting to Seen Datasets}\label{sec.exp2}
\label{sec:seen}

%\begin{rqbox}
\textbf{RQ2:} \textit{After multi-domain pre-training over diverse topic and format domains, can GFMs reliably adapt to seen downstream datasets?}
%\end{rqbox}

We report one-shot performance on seen datasets in Tables~\ref{table.exp2-1shotnc} and \ref{table.exp2-1shotecgc}, 
with five-shot results and efficiency provided in Appendix~\ref{app.exp2}.
Similar to RQ1, we also compare with \emph{supervised}  GCN and GAT. 
%and format-aware methods (\method{Simple-HGN} for heterogeneous graphs and \method{FAGCN} for heterophilic graph); (2) 
Additionally, we compare with \emph{conventional pre-training methods} that are pre-trained on the downstream graph without any labels and then adapted to the same graph with task-specific labels. Such methods include general-purpose approaches (\method{DGI} and \method{GraphPrompt}), and where applicable, format-specific approaches (\method{HeCo} for heterogeneous graphs, \method{TGN}, \method{DDGCL} for dynamic graphs, and \method{DSSL} for heterophilic graphs).

\stitle{Results.}
Similar to the performance on unseen datasets (RQ1), leading GFMs generally outperform supervised approaches, suggesting that broad pre-training can be beneficial to both seen and unseen downstream data.
%
%The seen dataset setting largely echoes the qualitative patterns observed on unseen datasets: GFMs performance remains dataset dependent. 
We further observe the following patterns.

%First, multi-domain pre-training can induce interference rather than cooperation.
First, multi-domain knowledge may not be effectively pre-trained or transferred to some target datasets. 
We observe that \method{GraphPrompt}, which is pre-trained and evaluated on the same graph, remains competitive across many datasets and even outperforms all GFMs on \textit{Cora} and \textit{Elliptic} in node classification, as well as on \textit{HIV} in graph classification. In cases where GFMs underperform, either the multi-domain pre-training is dominated by certain domain-specific knowledge, or the downstream target is unable to effectively leverage the pre-trained multi-domain representations. Consequently, graph pre-training methods that focus exclusively on the target graph can be more effective. 
%This suggests that mixing multiple domains during pre-training can introduce negative transfer, and current GFMs still lack effective mechanisms to mitigate this gap.

Second, GFMs are generally robust across formats. 
On the dynamic graph \textit{Reddit}, 
several GFMs, including \method{MDGFM}, \method{SAMGPT}, and \method{GFT}, attain better or comparable node classification performance than pre-training methods specifically designed for dynamic graphs, namely \method{DDGCL} and \method{TGN}. Similarly, on the heterophilic graph \textit{Wisconsin}, the heterophily-specific pre-training method \method{DSSL} improves over supervised GNNs in Macro-F, but still trails multiple GFMs, such as \method{GFT}, \method{G2P2}, \method{SAMGPT}, and \method{GraphCLIP}. A notable exception arises for heterogeneous graphs, such as \textit{ACM}. In particular, \method{HeCo}, a pre-training method specifically designed for heterogeneous graphs, substantially outperform all general-purpose GFMs and supervised baselines. One potential reason is that heterogeneous graphs exhibit rich and distinctive type-based semantics  that are not fully captured by general pre-training objectives.

Nevertheless, the best GFMs still achieve strong overall performance. More importantly, GFMs offer broader applicability than conventional graph pre-training methods, as pre-trained GFMs can generalize to both seen and unseen downstream graphs across formats, whereas the latter are limited to the seen setting and, in the case of format-aware approaches, to specific formats.

%\stitle{Results on seen edge/graph classification.}
% Table~\ref{table.exp2-1shotecgc} highlights behaviors that are less apparent from node classification alone.
%On \textit{FB15K-237}, several GFMs perform near the lowest range (e.g., \method{GCOPE}, \method{UniGraph2}), whereas \method{MDGPT} and \method{G2P2} are notably stronger and GCN is also competitive. This gap indicates that relational prediction is not automatically improved by broad pre-training, and that explicit relation-aware modeling remains important.

\stitle{Findings and insights.}
\textit{(i) Current GFMs are generally competitive on seen datasets, but effective integration and utilization of multi-domain knowledge remain a key bottleneck.}
While broad multi-domain pre-training can provide useful priors over the supervised methods for unseen and seen data alike, models pre-trained directly on the downstream graph can occasionally offer greater advantages. Hence, echoing the findings in RQ1, future GFM studies could focus on improving the alignment of multi-domain knowledge in both pre-training and downstream adaptation.

\textit{(ii) Current GFMs are generally robust across graph formats, with heterogeneous graphs being a notable exception, indicating that improving reliable format awareness remains an important next step.}
On dynamic and heterophilic graphs, the strongest GFMs can match or exceed format-specific baselines, indicating that multi-domain pre-training can implicitly recover useful format-specific inductive biases beyond the backbone. However, heterogeneous graphs still favor objectives that explicitly model typed structure, suggesting that robust cross-format transfer likely requires format-aware objectives or modular components that preserve and activate format-specific cues when present, rather than relying on a single uniform objective that may obscure such information.

\begin{table*}[tbp] % [!t]
    \centering
    \footnotesize
    \caption{Accuracy of one-shot node classification on topic domain adaptation. Marco-F1 reported in Appendix~\ref{app.exp3}.}
    \vspace{-2mm}
    \addtolength{\tabcolsep}{0.5mm}
    \label{table.exp3-1shotnc}
    %\resizebox{1\linewidth}{!}{
    \begin{tabular}{l|cc|ccc|c|cc|c}
    \toprule
    \multirow{2}{*}{Methods} &\multicolumn{2}{c|}{Citation} &\multicolumn{3}{c|}{Social \& Web} &{E-commerce} &\multicolumn{2}{c|}{Finance} &{Proteins} \\
    &Pubmed &ogbn-mag & {Wikipedia} & {Actor}  & {Chameleon} & {Products} & {T-Finance} & {DGraph} & {ogbn-proteins}\\
    \midrule\midrule
    \method{GCOPE} &37.93{\tiny$\pm$5.54} &0.26{\tiny$\pm$0.03} &52.79{\tiny$\pm$27.79}  &23.16{\tiny$\pm$7.74}  &21.82{\tiny$\pm$3.43}  &OOT  &33.94{\tiny$\pm$20.92} &OOT &47.70{\tiny$\pm$13.36} \\
    \method{MDGPT} &47.93{\tiny$\pm$10.33} &\underline{4.64}{\tiny$\pm$0.58} &51.80{\tiny$\pm$29.13}  &27.84{\tiny$\pm$10.71}  &\underline{27.26}{\tiny$\pm$4.97}  &\underline{18.98}{\tiny$\pm$5.42}  &\underline{59.88}{\tiny$\pm$23.85}  &36.79{\tiny$\pm$12.12}  &51.07{\tiny$\pm$10.98} \\
    \method{MDGFM} &35.62{\tiny$\pm$6.59}  &1.21{\tiny$\pm$1.24}  &50.67{\tiny$\pm$28.53}  &18.19{\tiny$\pm$5.80}  &23.24{\tiny$\pm$3.59}  &3.07{\tiny$\pm$0.90}  &49.22{\tiny$\pm$21.44}  &\textbf{41.96}{\tiny$\pm$28.01}  &49.78{\tiny$\pm$8.57} \\
    \method{SAMGPT} &\underline{48.11}{\tiny$\pm$6.83} &\textbf{6.57}{\tiny$\pm$0.68}  &\textbf{59.19}{\tiny$\pm$30.30}  &25.77{\tiny$\pm$7.33}  &\textbf{28.07}{\tiny$\pm$4.76}  &\textbf{24.32}{\tiny$\pm$4.42}  &48.87{\tiny$\pm$17.22}  &\underline{38.37}{\tiny$\pm$14.55}  &\underline{51.66}{\tiny$\pm$10.73} \\
    \method{G2P2} &43.30{\tiny$\pm$8.57} & -  &44.27{\tiny$\pm$7.80}  &\textbf{28.85}{\tiny$\pm$9.66}  &- &-  &59.72{\tiny$\pm$16.22}  &38.25{\tiny$\pm$10.39}  &- \\
    \method{GraphCLIP} &35.57{\tiny$\pm$3.98} & - &42.96{\tiny$\pm$24.16}  &23.58{\tiny$\pm$5.55}  &- &-  &44.73{\tiny$\pm$21.75}  &36.69{\tiny$\pm$25.38}  &- \\
    \method{GFT} &44.76{\tiny$\pm$7.97} &1.60{\tiny$\pm$0.99} &\underline{57.95}{\tiny$\pm$26.04}  &27.81{\tiny$\pm$7.64}  &25.82{\tiny$\pm$4.77}  &13.17{\tiny$\pm$3.39}  &55.81{\tiny$\pm$22.10}  &34.91{\tiny$\pm$14.26}  &\textbf{55.99}{\tiny$\pm$11.61} \\
    \method{UniGraph2} &\textbf{50.12}{\tiny$\pm$9.40} & 1.88{\tiny$\pm$0.24} &48.78{\tiny$\pm$21.22}  &\underline{28.00}{\tiny$\pm$8.25}  &25.61{\tiny$\pm$4.46}  &13.96{\tiny$\pm$3.99}  &\textbf{60.78}{\tiny$\pm$21.00}  &35.44{\tiny$\pm$12.68}  &49.89{\tiny$\pm$7.83} \\\bottomrule
    \end{tabular}%}
\end{table*}

\begin{table*}[tbp] % [!t]
    \centering
    \footnotesize
    \caption{Evaluation of one-shot edge classification and graph classification on topic domain adaptation.}
    \vspace{-2mm}
    %\addtolength{\tabcolsep}{0.5mm}
    \label{table.exp3-1shotecgc}
    %\resizebox{0.9\linewidth}{!}{
    \begin{tabular}{l|cc|cccc|cccc}
    \toprule
    \multirow{3}{*}{Methods} &\multicolumn{2}{c|}{Finance (Edge Classification)} &\multicolumn{4}{c|}{Common sense (Edge Classification)} &\multicolumn{4}{c}{Molecule (Graph Classification)}\\
    &\multicolumn{2}{c|}{DGraph} &\multicolumn{2}{c}{Wiki} &\multicolumn{2}{c|}{WN18RR} &\multicolumn{2}{c}{PCBA} &\multicolumn{2}{c}{BZR} \\
    &Acc &MacroF &Acc &MacroF &Acc &MacroF &Acc &MacroF &Acc &MacroF\\
    \midrule\midrule
    \method{GCOPE} &OOT &OOT &OOT &OOT &\underline{16.05}{\tiny$\pm$6.72} &\textbf{11.76}{\tiny$\pm$3.33} &\underline{61.25}{\tiny$\pm$24.71} &\underline{36.57}{\tiny$\pm$10.62} &\underline{57.37}{\tiny$\pm$19.40} &47.43{\tiny$\pm$11.66}\\
    \method{MDGPT} &9.10{\tiny$\pm$2.59} &\underline{6.42}{\tiny$\pm$0.94} &\underline{7.90}{\tiny$\pm$3.14} &\underline{1.57}{\tiny$\pm$0.23} &14.48{\tiny$\pm$4.91} &10.84{\tiny$\pm$2.25} &\textbf{62.51}{\tiny$\pm$24.39} &\textbf{37.11}{\tiny$\pm$11.11} &57.01{\tiny$\pm$14.30} &\underline{49.79}{\tiny$\pm$9.77}\\
    \method{MDGFM} &9.35{\tiny$\pm$3.31} &5.79{\tiny$\pm$1.16} &4.34{\tiny$\pm$2.91} &0.50{\tiny$\pm$0.09} &14.14{\tiny$\pm$5.30} &9.60{\tiny$\pm$2.52} &58.16{\tiny$\pm$42.55} &31.39{\tiny$\pm$21.02} &52.12{\tiny$\pm$24.02} &35.20{\tiny$\pm$9.12}\\
    \method{SAMGPT} &\textbf{10.82}{\tiny$\pm$3.14} &\textbf{6.97}{\tiny$\pm$1.04} &\textbf{11.59}{\tiny$\pm$3.03} &\textbf{3.52}{\tiny$\pm$0.16} &13.64{\tiny$\pm$5.11} &10.03{\tiny$\pm$1.91} &59.51{\tiny$\pm$24.76} &35.91{\tiny$\pm$10.89} &\textbf{59.29}{\tiny$\pm$15.52} &\textbf{50.81}{\tiny$\pm$10.13}\\
    \method{G2P2} &- &- &- &- &11.56{\tiny$\pm$4.44} &8.70{\tiny$\pm$1.78} &- &- &53.43{\tiny$\pm$15.93} &45.31{\tiny$\pm$8.49}\\
    \method{GraphCLIP} &- &- &- &- &\textbf{16.06}{\tiny$\pm$5.60} &\underline{11.54}{\tiny$\pm$1.98} &- &- &47.93{\tiny$\pm$16.58} &40.53{\tiny$\pm$10.30}\\
    \method{GFT} &9.45{\tiny$\pm$3.05} &5.57{\tiny$\pm$1.12} &OOT &OOT &12.77{\tiny$\pm$5.85} &9.87{\tiny$\pm$1.39} &45.03{\tiny$\pm$20.06} &28.44{\tiny$\pm$7.10} &55.66{\tiny$\pm$25.31} &37.46{\tiny$\pm$13.15}\\
    \method{UniGraph2} &\underline{9.78}{\tiny$\pm$3.13} &5.40{\tiny$\pm$1.03} &5.16{\tiny$\pm$2.69} &1.03{\tiny$\pm$0.10} &9.22{\tiny$\pm$6.90} &6.24{\tiny$\pm$1.44} &48.76{\tiny$\pm$14.24} &32.15{\tiny$\pm$6.77} &51.38{\tiny$\pm$7.15} &45.84{\tiny$\pm$4.66}\\\bottomrule
    \end{tabular}%}
\end{table*}

\subsection{Setting III: Adapting across Topic Domains}\label{sec.exp3}
\label{sec:topic}

%\begin{rqbox}
\noindent
\textbf{RQ3:} \textit{How do GFMs pre-trained on a single topic domain differ from those pre-trained on broad multi-topic domains when adapting to downstream topic domains?}
%\end{rqbox}

\begin{figure}[t]
\centering
\includegraphics[width=1\linewidth]{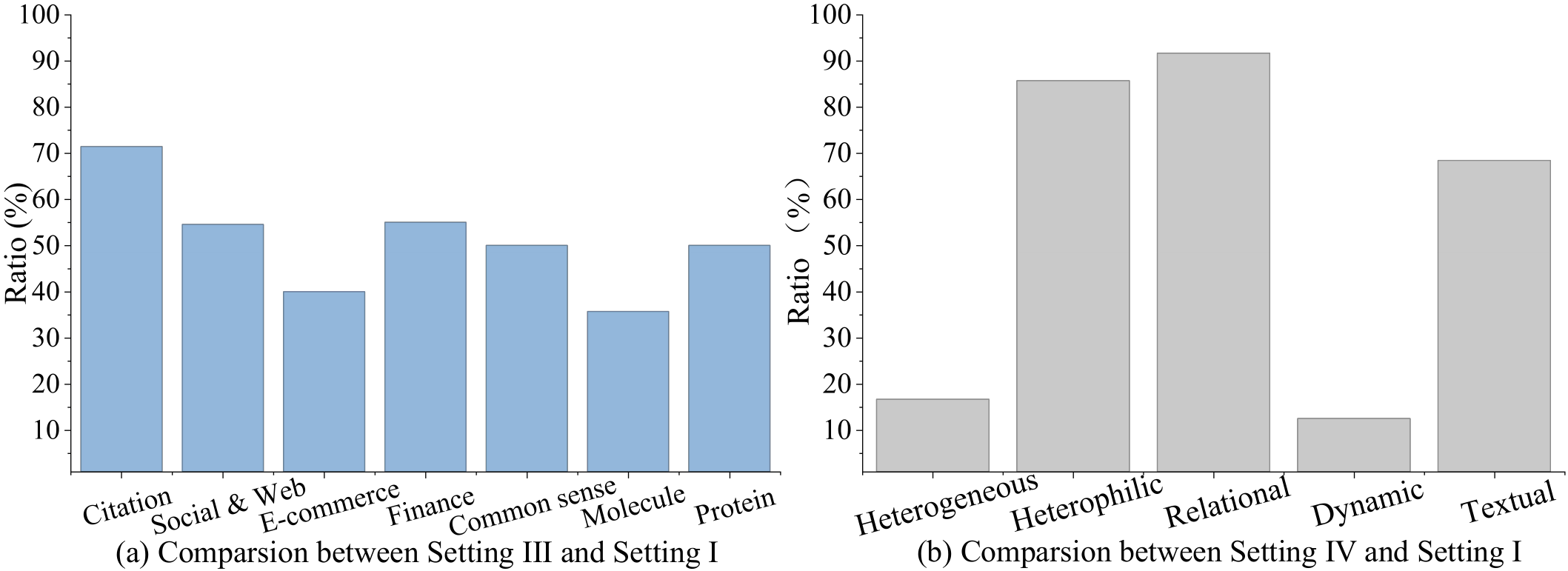}
\vspace{-5mm}
\caption{Comparisons with Setting I, with higher values favoring Setting I more. (a) Single-topic  (Setting III) vs.~multi-topic pre-training (Setting I). (b) Base format (Setting IV) vs.~multi-format pre-training (Setting I). }
\label{fig.setting3-4}
\end{figure}

As discussed in Sect.~\ref{sec.setting}, 
we pre-train on the citation topic domain only, and adapt to downstream datasets from various topic domains. 
We report the one-shot performance in Tables~\ref{table.exp3-1shotnc} and \ref{table.exp3-1shotecgc}, with additional Macro-F1 and five-shot results in Appendix~\ref{app.exp3}. 

To answer RQ3, we further compare the performance of Setting~III with that of Setting~I in Fig.~\ref{fig.setting3-4}(a). For each target domain, the $y$-axis reports the fraction of evaluated GFMs whose performance in Setting~III is lower than in Setting~I on the same dataset within that domain. This fraction is computed as the number of GFM-dataset pairs for which Setting~III underperforms Setting~I, divided by the total number of evaluated pairs in that domain.

\stitle{Results.}
First, expanding pre-training from a single citation topic to multi-topic domains is generally beneficial. As shown in Fig.~\ref{fig.setting3-4}(a), five out of seven downstream topic domains exhibit improved performance for at least half of the evaluated GFM-dataset pairs under Setting I, relative to Setting III.
Thus, in the majority of cases, increased topic diversity in pre-training provides complementary knowledge that enhances downstream adaptation to various topics.

%For adaptation to social, web, commerce and common sense datasets, results in the multi-domain pre-training setting (Setting~I) are typically stronger than citation-only pre-training for the same GFM, suggesting that additional topic diversity provide complementary knowledge that improves transfer.

Second, the benefit of multi-topic pre-training does not appear to correlate with topic proximity. One might expect broad multi-topic pre-training to be more advantageous for natural science domains (e.g., molecules) than citation-only pre-training, given the low topical proximity between citation and molecule graphs and the inclusion of natural science domains in multi-topic pre-training. %Conversely, smaller gains are expected for human activity-related domains (e.g., Social \& Web, Finance), given their higher topical proximity to citation graphs. 
However, we observe the opposite trend: when adapting to the molecule domain, citation-only pre-training is often competitive and can even slightly outperform multi-topic mixtures.

%citation-only pre-training to transfer more readily to other human-activity datasets than to natural-science graphs. Yet we observe the opposite trend in several cases: on natural-science domains such as molecules, citation-only pre-training is often competitive and can even slightly outperform multi-domain mixtures.

%Third, citation-only pre-training does not separate in-topic citation networks from other human-activity domains as much as expected. When adapting from citation graphs to other human-activity datasets, performance typically drops by a similar magnitude to adapting back to citation datasets themselves, relative to Setting~I.

\stitle{Findings and insights.}
\emph{(i) Broader topic coverage often strengthens GFMs, underscoring the importance for more topic diversity in pre-training.}
Adding more topic domains in pre-training generally improves downstream adaptation across various topics. This points to a concrete direction for building more robust GFMs: incorporating broader and more balanced topic domains, to encourage the learning of more transferable representations.

\textit{(ii) Topic proximity is not a good predictor of topic adaptation; dataset-level invariances matter more.}
Even within the same topic domain, datasets can differ markedly in feature semantics, label definitions, graph scale, and structural statistics. In practice, these dataset-level shifts often dominate transfer outcomes, making topic only a coarse descriptor. Accordingly, when curating pre-training corpora for GFMs, it is important to account for dataset-level diversity and mismatches, and to learn invariances at this level, rather than relying solely on topic proximity.

\begin{table*}[tbp] % [!t]
    \centering
    \footnotesize
    \caption{Evaluation of one-shot node classification on format domain adaptation.}
    \vspace{-2mm}
    %\addtolength{\tabcolsep}{0.5mm}
    \label{table.exp4-1shotnc}
    %\resizebox{1\linewidth}{!}{
    \begin{tabular}{l|cc|cccc|cc|cccc}
    \toprule
    \multirow{3}{*}{Methods} &\multicolumn{2}{c|}{Heterogeneous} &\multicolumn{4}{c|}{Heterophilic} &\multicolumn{2}{c|}{Dynamic} &\multicolumn{4}{c}{Textual} \\
     & \multicolumn{2}{c|}{ogbn-mag} & \multicolumn{2}{c}{Chameleon} & \multicolumn{2}{c|}{Actor} & \multicolumn{2}{c|}{Wikipedia} &\multicolumn{2}{c}{Pubmed} & \multicolumn{2}{c}{Products}  \\
     &Acc &MacroF &Acc &MacroF &Acc &MacroF &Acc &MacroF &Acc &MacroF &Acc &MacroF\\
    \midrule\midrule
    \method{GCOPE} &0.32{\tiny$\pm$0.04} &0.11{\tiny$\pm$0.01} &22.46{\tiny$\pm$2.93} &19.99{\tiny$\pm$3.38} &20.66{\tiny$\pm$8.01} &16.05{\tiny$\pm$4.58} &\textbf{58.69}{\tiny$\pm$29.29} &\textbf{36.89}{\tiny$\pm$14.91} &39.06{\tiny$\pm$7.55} &35.09{\tiny$\pm$8.07} &OOT &OOT\\
    \method{MDGPT} &\underline{6.01}{\tiny$\pm$0.81} &\underline{4.42}{\tiny$\pm$0.37} &\underline{27.90}{\tiny$\pm$5.13} &\textbf{25.06}{\tiny$\pm$5.03} &\textbf{27.23}{\tiny$\pm$8.64} &\textbf{22.22}{\tiny$\pm$5.09} &54.17{\tiny$\pm$22.13} &35.72{\tiny$\pm$10.71} &45.80{\tiny$\pm$10.14} &\underline{45.10}{\tiny$\pm$10.47} &\underline{21.23}{\tiny$\pm$4.96} &\underline{11.50}{\tiny$\pm$1.88}\\
    \method{MDGFM} &0.92{\tiny$\pm$0.54} &0.56{\tiny$\pm$0.41} &21.50{\tiny$\pm$3.12} &17.54{\tiny$\pm$4.00} &20.20{\tiny$\pm$7.73} &13.83{\tiny$\pm$3.12} &50.66{\tiny$\pm$29.90} &32.51{\tiny$\pm$14.61} &34.51{\tiny$\pm$5.84} &28.24{\tiny$\pm$4.55} &3.92{\tiny$\pm$2.17} &1.78{\tiny$\pm$0.47}\\
    \method{SAMGPT} &\textbf{7.35}{\tiny$\pm$0.73} &\textbf{5.49}{\tiny$\pm$0.27} &\textbf{28.14}{\tiny$\pm$4.77} &\underline{24.48}{\tiny$\pm$4.66} &24.58{\tiny$\pm$6.49} &19.21{\tiny$\pm$4.17} &55.95{\tiny$\pm$31.62} &34.51{\tiny$\pm$15.82} &\textbf{50.00}{\tiny$\pm$7.93} &\textbf{48.68}{\tiny$\pm$8.50} &\textbf{26.10}{\tiny$\pm$5.55} &\textbf{13.38}{\tiny$\pm$1.52}\\
    \method{G2P2} &- &- &- &- &\underline{27.04}{\tiny$\pm$10.41} &\underline{19.83}{\tiny$\pm$6.23} &54.78{\tiny$\pm$16.60} &28.69{\tiny$\pm$8.64} &44.78{\tiny$\pm$7.92} &41.63{\tiny$\pm$7.81} &- &-\\
    \method{GraphCLIP} &- &- &- &- &22.82{\tiny$\pm$5.34} &16.50{\tiny$\pm$2.44} &43.23{\tiny$\pm$20.45} &20.90{\tiny$\pm$7.66} &33.68{\tiny$\pm$4.22} &30.95{\tiny$\pm$3.54} &- &-\\
    \method{GFT} &1.71{\tiny$\pm$1.48} &0.54{\tiny$\pm$0.33} &26.31{\tiny$\pm$4.17} &22.15{\tiny$\pm$4.35} &24.29{\tiny$\pm$7.76} &19.22{\tiny$\pm$4.37} &\underline{56.94}{\tiny$\pm$31.46} &\underline{36.33}{\tiny$\pm$16.66} &45.32{\tiny$\pm$7.89} &43.17{\tiny$\pm$8.21} &6.30{\tiny$\pm$4.34} &3.26{\tiny$\pm$1.18}\\
    \method{UniGraph2} &2.36{\tiny$\pm$1.32} &0.66{\tiny$\pm$0.78} &25.96{\tiny$\pm$4.36} &23.77{\tiny$\pm$4.31} &22.52{\tiny$\pm$8.12} &17.12{\tiny$\pm$4.23} &54.87{\tiny$\pm$24.50} &35.63{\tiny$\pm$11.55} &\underline{45.97}{\tiny$\pm$9.20} &44.60{\tiny$\pm$9.27} &12.91{\tiny$\pm$8.53} &7.38{\tiny$\pm$1.05}\\\bottomrule
    \end{tabular}%}
\end{table*}

\begin{table}[tbp] % [!t]
    \centering
    \footnotesize
    \caption{Evaluation of one-shot edge classification and graph classification on format domain adaptation.}
    \vspace{-2mm}
    \addtolength{\tabcolsep}{-0.5mm}
    \label{table.exp4-1shotecgc}
    \resizebox{1\linewidth}{!}{
    \begin{tabular}{@{}l|cccc|cc@{}}
    \toprule
    \multirow{3}{*}{Methods} &\multicolumn{4}{c|}{Relational (Edge Classification)} &\multicolumn{2}{c}{Textual (Graph Classif.)} \\
    &\multicolumn{2}{c}{Wiki} &\multicolumn{2}{c|}{WN18RR} &\multicolumn{2}{c}{PCBA} \\
    &Acc &MacroF &Acc &MacroF &Acc &MacroF \\
    \midrule\midrule
    \method{GCOPE} &OOT &OOT &\textbf{17.76}{\tiny$\pm$6.44} &\textbf{12.87}{\tiny$\pm$3.03} &50.93{\tiny$\pm$34.41} &30.23{\tiny$\pm$16.26}\\
    \method{MDGPT} &\underline{10.07}{\tiny$\pm$3.05} &\underline{2.75}{\tiny$\pm$0.18} &14.71{\tiny$\pm$6.07} &10.55{\tiny$\pm$2.46} &\underline{55.25}{\tiny$\pm$24.41} &34.04{\tiny$\pm$11.43}\\
    \method{MDGFM} &3.85{\tiny$\pm$2.53} &0.37{\tiny$\pm$0.09} &13.77{\tiny$\pm$5.17} &9.93{\tiny$\pm$2.07} &51.68{\tiny$\pm$48.73} &26.37{\tiny$\pm$24.37}\\
    \method{SAMGPT} &\textbf{12.33}{\tiny$\pm$3.59} &\textbf{3.69}{\tiny$\pm$0.18} &13.53{\tiny$\pm$4.88} &9.97{\tiny$\pm$2.11} &\textbf{56.06}{\tiny$\pm$22.56} &\underline{34.63}{\tiny$\pm$10.44}\\
    \method{G2P2} &- &- &16.35{\tiny$\pm$8.67} &9.50{\tiny$\pm$2.37} &- &-\\
    \method{GraphCLIP} &- &- &\underline{16.85}{\tiny$\pm$4.74} &\underline{12.19}{\tiny$\pm$1.96} &- &-\\
    \method{GFT} &OOT &OOT &13.24{\tiny$\pm$6.67} &8.84{\tiny$\pm$1.60} &51.30{\tiny$\pm$28.10} &\textbf{35.62}{\tiny$\pm$15.45}\\
    \method{UniGraph2} &5.02{\tiny$\pm$3.20} &2.29{\tiny$\pm$0.19} &10.87{\tiny$\pm$8.36} &8.08{\tiny$\pm$1.58} &49.37{\tiny$\pm$22.55} &32.39{\tiny$\pm$11.87}\\\bottomrule
    \end{tabular}}
\end{table}

\subsection{Setting IV: Format Domain Adaptation}\label{sec.exp4}

%\begin{rqbox}
\textbf{RQ4:} \textit{How do GFMs pre-trained on a base format domain differ from those pre-trained on a multi-format domains when adapting to downstream format domains?}
%\end{rqbox}

As discussed in Sect.~\ref{sec.setting}, we pre-train on a base format (homogeneous, homophilic, static, single-relation, text-free) and adapt to datasets from other format domains. We report the one-shot performance in Tables~\ref{table.exp4-1shotnc} and \ref{table.exp4-1shotecgc}, with five-shot results in Appendix~\ref{app.exp4}. We further compare Setting IV to Setting~I in Fig.~\ref{fig.setting3-4}(b) to answer RQ4. Similar to Fig.~\ref{fig.setting3-4}(a), the $y$-axis reports the per format domain ratio of evaluated GFM-dataset pairs whose performance in Setting~IV is lower than in Setting~I.

\stitle{Results.}
We first observe that multi-format pre-training is not uniformly helpful across formats. As shown in Fig.~\ref{fig.setting3-4}(b), compare to Setting~I, pre-training restricted to base-format graphs (Setting~IV)  typically performs worse on heterophilic, relational, and text-attributed graphs. This indicates that most GFMs can benefit from moderate format diversification even without explicit format-specific architectures or objectives. In contrast, for heterogeneous and dynamic graphs, the pattern flips: Setting~I often underperforms Setting~IV, suggesting that mixing in various formats may introduce interference and degrade adaptation to these two formats whose representational gap from the base format is more pronounced.

Second, when adapting to text-attributed graphs, GFMs with text encoders (e.g., \method{G2P2}, \method{GraphCLIP}, \method{GFT}, \method{UniGraph2}) pre-trained on the base format (Setting IV) generally tend to underperform  pre-training with multiple formats (Setting I). The reason is that, under Setting IV, textual information is absent during pre-training, leaving their text encoders with little meaningful supervision. When text becomes available downstream, the additional modality behaves more like noise than a useful signal, leading to weaker transfer.

% \stitle{Edge and graph classification under format adaptation.}
% Compared with node-level results, edge- and graph-level transfer is noticeably more sensitive to \emph{how} the pre-training corpus is assembled. Expanding pre-training within a format-consistent base regime often improves over citation-only pre-training, suggesting that scale helps when the added data is representationally compatible. Yet this advantage does not extend to naive multi-domain mixtures: even with the broadest corpus, multi-domain pre-training can underperform citation-only on the same relational targets.

\stitle{Findings and insights.}
\textit{(i) GFMs show encouraging potential for cross-format adaptation, but reliability depends on the gap between the target and base format.}
Format diversification in pre-training can help downstream adaptation to heterophilic, relational and text-attributed graphs, plausibly because these format shifts remain reasonably aligned with base-format inductive biases. In contrast, heterogeneous and dynamic graphs behave differently, where na\"ive format mixing often degrades adaptation quality. This suggests a larger representational gap that requires more explicit modeling. Overall, for such formats, devising format-specific pre-training modules and downstream alignment appears necessary.

\textit{(ii) Text-assisted GFMs can be effective, but their performance degrade severely in the absence of text during pre-training.}
When pre-training does not include textual signals, GFMs  with text encoders become underfit and struggle to adapt to downstream text-attributed graphs. Thus, a practical direction is to treat text as an optional enhancement, while maintaining a robust text-free pathway so that adaptation remains competitive in text-sparse regimes.

\section{Conclusions and Future Directions}
We introduced a new benchmark for graph foundation models that treats domain shift as inherently two-dimensional, disentangling topic semantics from graph formats. We evaluated eight up-to-date GFMs on 33 datasets spanning seven topic domains
and six format domains, under four complementary settings: (i) 
pre-training on diverse topics and formats and adapting to unseen datasets; (ii) same pre-training as in (i), but adapting to seen datasets; (iii) pre-training on a single topic domain, and adapting to other topics; and (iv) pre-training on a base format, and adapting to other formats.

Across all four settings, we find that current GFMs are promising but still fall short of robustness. On unseen downstream datasets, GFMs can readily outperform supervised GNNs, yet the gains are uneven across methods and targets. On seen datasets, improvements over conventional graph pre-training methods are not monotonic, indicating that effective integration and utilization of multi-domain knowledge remain a key bottleneck.
We further observe that broader topic coverage often helps, but transfer is governed more by dataset-level invariances than by coarse topic proximity. 
Finally, expanding format diversity during pre-training can benefit target formats that exhibit smaller shifts from the base format, whereas others may require more explicit format-aware designs.

These findings motivate several important directions for future GFMs. First, future studies should focus on integrating complementary rather than conflicting multi-domain knowledge during pre-training, and on better aligning and reusing cross-domain signals during downstream adaptation. Second, while expanding topic domains is a useful strategy for broader pre-training, it is equally important to account for dataset-level differences and to exploit shared invariances across datasets, rather than relying solely on coarse topic classification. Finally, improving robustness under large format gaps likely requires format-specific architectures or objectives, together with explicit cross-format alignment mechanisms.

\clearpage
\newpage

\bibliographystyle{ACM-Reference-Format}
\bibliography{references}

%%% -*-BibTeX-*-
%%% Do NOT edit. File created by BibTeX with style
%%% ACM-Reference-Format-Journals [18-Jan-2012].

\begin{thebibliography}{85}

%%% ====================================================================
%%% NOTE TO THE USER: you can override these defaults by providing
%%% customized versions of any of these macros before the \bibliography
%%% command.  Each of them MUST provide its own final punctuation,
%%% except for \shownote{} and \showURL{}.  The latter two
%%% do not use final punctuation, in order to avoid confusing it with
%%% the Web address.
%%%
%%% To suppress output of a particular field, define its macro to expand
%%% to an empty string, or better, \unskip, like this:
%%%
%%% \newcommand{\showURL}[1]{\unskip}   % LaTeX syntax
%%%
%%% \def \showURL #1{\unskip}           % plain TeX syntax
%%%
%%% ====================================================================

\ifx \showCODEN    \undefined \def \showCODEN     #1{\unskip}     \fi
\ifx \showISBNx    \undefined \def \showISBNx     #1{\unskip}     \fi
\ifx \showISBNxiii \undefined \def \showISBNxiii  #1{\unskip}     \fi
\ifx \showISSN     \undefined \def \showISSN      #1{\unskip}     \fi
\ifx \showLCCN     \undefined \def \showLCCN      #1{\unskip}     \fi
\ifx \shownote     \undefined \def \shownote      #1{#1}          \fi
\ifx \showarticletitle \undefined \def \showarticletitle #1{#1}   \fi
\ifx \showURL      \undefined \def \showURL       {\relax}        \fi
% The following commands are used for tagged output and should be
% invisible to TeX
\providecommand\bibfield[2]{#2}
\providecommand\bibinfo[2]{#2}
\providecommand\natexlab[1]{#1}
\providecommand\showeprint[2][]{arXiv:#2}

\bibitem[Achiam et~al\mbox{.}(2023)]%
        {achiam2023gpt}
\bibfield{author}{\bibinfo{person}{Josh Achiam}, \bibinfo{person}{Steven Adler}, \bibinfo{person}{Sandhini Agarwal}, \bibinfo{person}{Lama Ahmad}, \bibinfo{person}{Ilge Akkaya}, \bibinfo{person}{Florencia~Leoni Aleman}, \bibinfo{person}{Diogo Almeida}, \bibinfo{person}{Janko Altenschmidt}, \bibinfo{person}{Sam Altman}, \bibinfo{person}{Shyamal Anadkat}, {et~al\mbox{.}}} \bibinfo{year}{2023}\natexlab{}.
\newblock \showarticletitle{Gpt-4 technical report}.
\newblock \bibinfo{journal}{\emph{arXiv preprint arXiv:2303.08774}} (\bibinfo{year}{2023}).
\newblock


\bibitem[Bardoscia et~al\mbox{.}(2021)]%
        {bardoscia2021physics}
\bibfield{author}{\bibinfo{person}{Marco Bardoscia}, \bibinfo{person}{Paolo Barucca}, \bibinfo{person}{Stefano Battiston}, \bibinfo{person}{Fabio Caccioli}, \bibinfo{person}{Giulio Cimini}, \bibinfo{person}{Diego Garlaschelli}, \bibinfo{person}{Fabio Saracco}, \bibinfo{person}{Tiziano Squartini}, {and} \bibinfo{person}{Guido Caldarelli}.} \bibinfo{year}{2021}\natexlab{}.
\newblock \showarticletitle{The physics of financial networks}.
\newblock \bibinfo{journal}{\emph{Nature Reviews Physics}} \bibinfo{volume}{3}, \bibinfo{number}{7} (\bibinfo{year}{2021}), \bibinfo{pages}{490--507}.
\newblock


\bibitem[Bi et~al\mbox{.}(2024)]%
        {bi2024make}
\bibfield{author}{\bibinfo{person}{Wendong Bi}, \bibinfo{person}{Lun Du}, \bibinfo{person}{Qiang Fu}, \bibinfo{person}{Yanlin Wang}, \bibinfo{person}{Shi Han}, {and} \bibinfo{person}{Dongmei Zhang}.} \bibinfo{year}{2024}\natexlab{}.
\newblock \showarticletitle{Make heterophilic graphs better fit gnn: A graph rewiring approach}.
\newblock \bibinfo{journal}{\emph{TKDE}} (\bibinfo{year}{2024}).
\newblock


\bibitem[Bo et~al\mbox{.}(2021)]%
        {bo2021beyond}
\bibfield{author}{\bibinfo{person}{Deyu Bo}, \bibinfo{person}{Xiao Wang}, \bibinfo{person}{Chuan Shi}, {and} \bibinfo{person}{Huawei Shen}.} \bibinfo{year}{2021}\natexlab{}.
\newblock \showarticletitle{Beyond low-frequency information in graph convolutional networks}. In \bibinfo{booktitle}{\emph{Proceedings of the AAAI conference on artificial intelligence}}, Vol.~\bibinfo{volume}{35}. \bibinfo{pages}{3950--3957}.
\newblock


\bibitem[Borgwardt et~al\mbox{.}(2005)]%
        {borgwardt2005protein}
\bibfield{author}{\bibinfo{person}{Karsten~M Borgwardt}, \bibinfo{person}{Cheng~Soon Ong}, \bibinfo{person}{Stefan Sch{\"o}nauer}, \bibinfo{person}{SVN Vishwanathan}, \bibinfo{person}{Alex~J Smola}, {and} \bibinfo{person}{Hans-Peter Kriegel}.} \bibinfo{year}{2005}\natexlab{}.
\newblock \showarticletitle{Protein function prediction via graph kernels}.
\newblock \bibinfo{journal}{\emph{Bioinformatics}} \bibinfo{volume}{21}, \bibinfo{number}{suppl\_1} (\bibinfo{year}{2005}), \bibinfo{pages}{i47--i56}.
\newblock


\bibitem[Carlson et~al\mbox{.}(2010)]%
        {carlson2010toward}
\bibfield{author}{\bibinfo{person}{Andrew Carlson}, \bibinfo{person}{Justin Betteridge}, \bibinfo{person}{Bryan Kisiel}, \bibinfo{person}{Burr Settles}, \bibinfo{person}{Estevam Hruschka}, {and} \bibinfo{person}{Tom Mitchell}.} \bibinfo{year}{2010}\natexlab{}.
\newblock \showarticletitle{Toward an architecture for never-ending language learning}. In \bibinfo{booktitle}{\emph{Proceedings of the AAAI conference on artificial intelligence}}, Vol.~\bibinfo{volume}{24}. \bibinfo{pages}{1306--1313}.
\newblock


\bibitem[Cen et~al\mbox{.}(2019)]%
        {cen2019representation}
\bibfield{author}{\bibinfo{person}{Yukuo Cen}, \bibinfo{person}{Xu Zou}, \bibinfo{person}{Jianwei Zhang}, \bibinfo{person}{Hongxia Yang}, \bibinfo{person}{Jingren Zhou}, {and} \bibinfo{person}{Jie Tang}.} \bibinfo{year}{2019}\natexlab{}.
\newblock \showarticletitle{Representation learning for attributed multiplex heterogeneous network}. In \bibinfo{booktitle}{\emph{Proceedings of the 25th ACM SIGKDD international conference on knowledge discovery \& data mining}}. \bibinfo{pages}{1358--1368}.
\newblock


\bibitem[Chen et~al\mbox{.}(2024b)]%
        {chen2024llaga}
\bibfield{author}{\bibinfo{person}{Runjin Chen}, \bibinfo{person}{Tong Zhao}, \bibinfo{person}{Ajay~Kumar Jaiswal}, \bibinfo{person}{Neil Shah}, {and} \bibinfo{person}{Zhangyang Wang}.} \bibinfo{year}{2024}\natexlab{b}.
\newblock \showarticletitle{LLaGA: Large Language and Graph Assistant}. In \bibinfo{booktitle}{\emph{ICML}}. \bibinfo{pages}{7809--7823}.
\newblock


\bibitem[Chen et~al\mbox{.}(2020)]%
        {chen2020review}
\bibfield{author}{\bibinfo{person}{Xiaojun Chen}, \bibinfo{person}{Shengbin Jia}, {and} \bibinfo{person}{Yang Xiang}.} \bibinfo{year}{2020}\natexlab{}.
\newblock \showarticletitle{A review: Knowledge reasoning over knowledge graph}.
\newblock \bibinfo{journal}{\emph{Expert Syst. Appl.}}  \bibinfo{volume}{141} (\bibinfo{year}{2020}).
\newblock


\bibitem[Chen et~al\mbox{.}(2023)]%
        {chen2023exploring}
\bibfield{author}{\bibinfo{person}{Zhikai Chen}, \bibinfo{person}{Haitao Mao}, \bibinfo{person}{Hang Li}, \bibinfo{person}{Wei Jin}, \bibinfo{person}{Hongzhi Wen}, \bibinfo{person}{Xiaochi Wei}, \bibinfo{person}{Shuaiqiang Wang}, \bibinfo{person}{Dawei Yin}, \bibinfo{person}{Wenqi Fan}, \bibinfo{person}{Hui Liu}, {et~al\mbox{.}}} \bibinfo{year}{2023}\natexlab{}.
\newblock \showarticletitle{Exploring the potential of large language models (llms) in learning on graphs}.
\newblock \bibinfo{journal}{\emph{{SIGKDD} Explor.}} \bibinfo{volume}{25}, \bibinfo{number}{2} (\bibinfo{year}{2023}), \bibinfo{pages}{42--61}.
\newblock


\bibitem[Chen et~al\mbox{.}(2024a)]%
        {chen2024text}
\bibfield{author}{\bibinfo{person}{Zhikai Chen}, \bibinfo{person}{Haitao Mao}, \bibinfo{person}{Jingzhe Liu}, \bibinfo{person}{Yu Song}, \bibinfo{person}{Bingheng Li}, \bibinfo{person}{Wei Jin}, \bibinfo{person}{Bahare Fatemi}, \bibinfo{person}{Anton Tsitsulin}, \bibinfo{person}{Bryan Perozzi}, \bibinfo{person}{Hui Liu}, {et~al\mbox{.}}} \bibinfo{year}{2024}\natexlab{a}.
\newblock \showarticletitle{Text-space graph foundation models: Comprehensive benchmarks and new insights}. In \bibinfo{booktitle}{\emph{NeurIPS}}, Vol.~\bibinfo{volume}{37}. \bibinfo{pages}{7464--7492}.
\newblock


\bibitem[Dettmers et~al\mbox{.}(2018)]%
        {dettmers2018convolutional}
\bibfield{author}{\bibinfo{person}{Tim Dettmers}, \bibinfo{person}{Pasquale Minervini}, \bibinfo{person}{Pontus Stenetorp}, {and} \bibinfo{person}{Sebastian Riedel}.} \bibinfo{year}{2018}\natexlab{}.
\newblock \showarticletitle{Convolutional 2d knowledge graph embeddings}. In \bibinfo{booktitle}{\emph{Proceedings of the AAAI conference on artificial intelligence}}, Vol.~\bibinfo{volume}{32}.
\newblock


\bibitem[Dong et~al\mbox{.}(2017)]%
        {dong2017metapath2vec}
\bibfield{author}{\bibinfo{person}{Yuxiao Dong}, \bibinfo{person}{Nitesh~V. Chawla}, {and} \bibinfo{person}{Ananthram Swami}.} \bibinfo{year}{2017}\natexlab{}.
\newblock \showarticletitle{metapath2vec: Scalable Representation Learning for Heterogeneous Networks}. In \bibinfo{booktitle}{\emph{KDD}}. \bibinfo{pages}{135--144}.
\newblock


\bibitem[Freeman et~al\mbox{.}(2004)]%
        {freeman2004development}
\bibfield{author}{\bibinfo{person}{Linton Freeman} {et~al\mbox{.}}} \bibinfo{year}{2004}\natexlab{}.
\newblock \showarticletitle{The development of social network analysis}.
\newblock \bibinfo{journal}{\emph{A Study in the Sociology of Science}} \bibinfo{volume}{1}, \bibinfo{number}{687} (\bibinfo{year}{2004}), \bibinfo{pages}{159--167}.
\newblock


\bibitem[Giuliani et~al\mbox{.}(2008)]%
        {giuliani2008proteins}
\bibfield{author}{\bibinfo{person}{Alessandro Giuliani}, \bibinfo{person}{Arun Krishnan}, \bibinfo{person}{Joseph~P Zbilut}, {and} \bibinfo{person}{Masaru Tomita}.} \bibinfo{year}{2008}\natexlab{}.
\newblock \showarticletitle{Proteins as networks: usefulness of graph theory in protein science}.
\newblock \bibinfo{journal}{\emph{Current Protein and Peptide Science}} \bibinfo{volume}{9}, \bibinfo{number}{1} (\bibinfo{year}{2008}), \bibinfo{pages}{28--38}.
\newblock


\bibitem[Haldane(2013)]%
        {haldane2013rethinking}
\bibfield{author}{\bibinfo{person}{Andrew~G Haldane}.} \bibinfo{year}{2013}\natexlab{}.
\newblock \showarticletitle{Rethinking the financial network}.
\newblock In \bibinfo{booktitle}{\emph{Fragile stabilit{\"a}t--stabile fragilit{\"a}t}}. \bibinfo{publisher}{Springer}, \bibinfo{pages}{243--278}.
\newblock


\bibitem[Hamilton et~al\mbox{.}(2017)]%
        {hamilton2017inductive}
\bibfield{author}{\bibinfo{person}{Will Hamilton}, \bibinfo{person}{Zhitao Ying}, {and} \bibinfo{person}{Jure Leskovec}.} \bibinfo{year}{2017}\natexlab{}.
\newblock \showarticletitle{Inductive representation learning on large graphs}. In \bibinfo{booktitle}{\emph{NeurIPS}}, Vol.~\bibinfo{volume}{30}.
\newblock


\bibitem[Harary and Gupta(1997)]%
        {harary1997dynamic}
\bibfield{author}{\bibinfo{person}{Frank Harary} {and} \bibinfo{person}{Gopal Gupta}.} \bibinfo{year}{1997}\natexlab{}.
\newblock \showarticletitle{Dynamic graph models}.
\newblock \bibinfo{journal}{\emph{Mathematical and Computer Modelling}} \bibinfo{volume}{25}, \bibinfo{number}{7} (\bibinfo{year}{1997}), \bibinfo{pages}{79--87}.
\newblock


\bibitem[He et~al\mbox{.}(2025)]%
        {he2025unigraph2}
\bibfield{author}{\bibinfo{person}{Yufei He}, \bibinfo{person}{Yuan Sui}, \bibinfo{person}{Xiaoxin He}, \bibinfo{person}{Yue Liu}, \bibinfo{person}{Yifei Sun}, {and} \bibinfo{person}{Bryan Hooi}.} \bibinfo{year}{2025}\natexlab{}.
\newblock \showarticletitle{Unigraph2: Learning a unified embedding space to bind multimodal graphs}. In \bibinfo{booktitle}{\emph{WWW}}. \bibinfo{pages}{1759--1770}.
\newblock


\bibitem[Hu et~al\mbox{.}(2020b)]%
        {hu2020open}
\bibfield{author}{\bibinfo{person}{Weihua Hu}, \bibinfo{person}{Matthias Fey}, \bibinfo{person}{Marinka Zitnik}, \bibinfo{person}{Yuxiao Dong}, \bibinfo{person}{Hongyu Ren}, \bibinfo{person}{Bowen Liu}, \bibinfo{person}{Michele Catasta}, {and} \bibinfo{person}{Jure Leskovec}.} \bibinfo{year}{2020}\natexlab{b}.
\newblock \showarticletitle{Open graph benchmark: Datasets for machine learning on graphs}. In \bibinfo{booktitle}{\emph{NeurIPS}}. \bibinfo{pages}{22118--22133}.
\newblock


\bibitem[Hu et~al\mbox{.}(2020a)]%
        {hu2020gpt}
\bibfield{author}{\bibinfo{person}{Ziniu Hu}, \bibinfo{person}{Yuxiao Dong}, \bibinfo{person}{Kuansan Wang}, \bibinfo{person}{Kai-Wei Chang}, {and} \bibinfo{person}{Yizhou Sun}.} \bibinfo{year}{2020}\natexlab{a}.
\newblock \showarticletitle{Gpt-gnn: Generative pre-training of graph neural networks}. In \bibinfo{booktitle}{\emph{SIGKDD}}. \bibinfo{pages}{1857--1867}.
\newblock


\bibitem[Huang et~al\mbox{.}(2023)]%
        {huang2023prodigy}
\bibfield{author}{\bibinfo{person}{Qian Huang}, \bibinfo{person}{Hongyu Ren}, \bibinfo{person}{Peng Chen}, \bibinfo{person}{Gregor Kr{\v{z}}manc}, \bibinfo{person}{Daniel Zeng}, \bibinfo{person}{Percy Liang}, {and} \bibinfo{person}{Jure Leskovec}.} \bibinfo{year}{2023}\natexlab{}.
\newblock \showarticletitle{PRODIGY: Enabling In-context Learning Over Graphs}. In \bibinfo{booktitle}{\emph{NeurIPS}}.
\newblock


\bibitem[Huang et~al\mbox{.}(2022)]%
        {huang2022dgraph}
\bibfield{author}{\bibinfo{person}{Xuanwen Huang}, \bibinfo{person}{Yang Yang}, \bibinfo{person}{Yang Wang}, \bibinfo{person}{Chunping Wang}, \bibinfo{person}{Zhisheng Zhang}, \bibinfo{person}{Jiarong Xu}, \bibinfo{person}{Lei Chen}, {and} \bibinfo{person}{Michalis Vazirgiannis}.} \bibinfo{year}{2022}\natexlab{}.
\newblock \showarticletitle{Dgraph: A large-scale financial dataset for graph anomaly detection}.
\newblock \bibinfo{journal}{\emph{Advances in Neural Information Processing Systems}}  \bibinfo{volume}{35} (\bibinfo{year}{2022}), \bibinfo{pages}{22765--22777}.
\newblock


\bibitem[Kearnes et~al\mbox{.}(2016)]%
        {kearnes2016molecular}
\bibfield{author}{\bibinfo{person}{Steven Kearnes}, \bibinfo{person}{Kevin McCloskey}, \bibinfo{person}{Marc Berndl}, \bibinfo{person}{Vijay Pande}, {and} \bibinfo{person}{Patrick Riley}.} \bibinfo{year}{2016}\natexlab{}.
\newblock \showarticletitle{Molecular graph convolutions: moving beyond fingerprints}.
\newblock \bibinfo{journal}{\emph{Journal of computer-aided molecular design}} \bibinfo{volume}{30}, \bibinfo{number}{8} (\bibinfo{year}{2016}), \bibinfo{pages}{595--608}.
\newblock


\bibitem[Kingma and Ba(2015)]%
        {kingma2015adam}
\bibfield{author}{\bibinfo{person}{Diederik~P. Kingma} {and} \bibinfo{person}{Jimmy Ba}.} \bibinfo{year}{2015}\natexlab{}.
\newblock \showarticletitle{Adam: A Method for Stochastic Optimization}. In \bibinfo{booktitle}{\emph{ICLR}}.
\newblock


\bibitem[Kipf and Welling(2017)]%
        {kipf2016semi}
\bibfield{author}{\bibinfo{person}{Thomas~N Kipf} {and} \bibinfo{person}{Max Welling}.} \bibinfo{year}{2017}\natexlab{}.
\newblock \showarticletitle{Semi-supervised classification with graph convolutional networks}. In \bibinfo{booktitle}{\emph{ICLR}}.
\newblock


\bibitem[Kumar et~al\mbox{.}(2019)]%
        {kumar2019predicting}
\bibfield{author}{\bibinfo{person}{Srijan Kumar}, \bibinfo{person}{Xikun Zhang}, {and} \bibinfo{person}{Jure Leskovec}.} \bibinfo{year}{2019}\natexlab{}.
\newblock \showarticletitle{Predicting dynamic embedding trajectory in temporal interaction networks}. In \bibinfo{booktitle}{\emph{Proceedings of the 25th ACM SIGKDD international conference on knowledge discovery \& data mining}}. \bibinfo{pages}{1269--1278}.
\newblock


\bibitem[Li et~al\mbox{.}(2025)]%
        {li2025hetgb}
\bibfield{author}{\bibinfo{person}{Shujie Li}, \bibinfo{person}{Yuxia Wu}, \bibinfo{person}{Chuan Shi}, {and} \bibinfo{person}{Yuan Fang}.} \bibinfo{year}{2025}\natexlab{}.
\newblock \showarticletitle{HeTGB: A Comprehensive Benchmark for Heterophilic Text-Attributed Graphs}.
\newblock \bibinfo{journal}{\emph{arXiv preprint arXiv:2503.04822}} (\bibinfo{year}{2025}).
\newblock


\bibitem[Li et~al\mbox{.}(2024)]%
        {li2024zerog}
\bibfield{author}{\bibinfo{person}{Yuhan Li}, \bibinfo{person}{Peisong Wang}, \bibinfo{person}{Zhixun Li}, \bibinfo{person}{Jeffrey~Xu Yu}, {and} \bibinfo{person}{Jia Li}.} \bibinfo{year}{2024}\natexlab{}.
\newblock \showarticletitle{Zerog: Investigating cross-dataset zero-shot transferability in graphs}. In \bibinfo{booktitle}{\emph{SIGKDD}}. \bibinfo{pages}{1725--1735}.
\newblock


\bibitem[Liu et~al\mbox{.}(2024b)]%
        {liu2024deepseek}
\bibfield{author}{\bibinfo{person}{Aixin Liu}, \bibinfo{person}{Bei Feng}, \bibinfo{person}{Bing Xue}, \bibinfo{person}{Bingxuan Wang}, \bibinfo{person}{Bochao Wu}, \bibinfo{person}{Chengda Lu}, \bibinfo{person}{Chenggang Zhao}, \bibinfo{person}{Chengqi Deng}, \bibinfo{person}{Chenyu Zhang}, \bibinfo{person}{Chong Ruan}, {et~al\mbox{.}}} \bibinfo{year}{2024}\natexlab{b}.
\newblock \showarticletitle{Deepseek-v3 technical report}.
\newblock \bibinfo{journal}{\emph{arXiv preprint arXiv:2412.19437}} (\bibinfo{year}{2024}).
\newblock


\bibitem[Liu et~al\mbox{.}(2024a)]%
        {liu2023one}
\bibfield{author}{\bibinfo{person}{Hao Liu}, \bibinfo{person}{Jiarui Feng}, \bibinfo{person}{Lecheng Kong}, \bibinfo{person}{Ningyue Liang}, \bibinfo{person}{Dacheng Tao}, \bibinfo{person}{Yixin Chen}, {and} \bibinfo{person}{Muhan Zhang}.} \bibinfo{year}{2024}\natexlab{a}.
\newblock \showarticletitle{One for All: Towards Training One Graph Model for All Classification Tasks}. In \bibinfo{booktitle}{\emph{ICLR}}.
\newblock


\bibitem[Liu et~al\mbox{.}(2025a)]%
        {liu2025graph}
\bibfield{author}{\bibinfo{person}{Jiawei Liu}, \bibinfo{person}{Cheng Yang}, \bibinfo{person}{Zhiyuan Lu}, \bibinfo{person}{Junze Chen}, \bibinfo{person}{Yibo Li}, \bibinfo{person}{Mengmei Zhang}, \bibinfo{person}{Ting Bai}, \bibinfo{person}{Yuan Fang}, \bibinfo{person}{Lichao Sun}, \bibinfo{person}{Philip~S Yu}, {et~al\mbox{.}}} \bibinfo{year}{2025}\natexlab{a}.
\newblock \showarticletitle{Graph foundation models: Concepts, opportunities and challenges}.
\newblock \bibinfo{journal}{\emph{IEEE TPAMI}} (\bibinfo{year}{2025}).
\newblock


\bibitem[Liu et~al\mbox{.}(2025b)]%
        {liu2025roft}
\bibfield{author}{\bibinfo{person}{Shikun Liu}, \bibinfo{person}{Deyu Zou}, \bibinfo{person}{Nima Shoghi}, \bibinfo{person}{Victor Fung}, \bibinfo{person}{Kai Liu}, {and} \bibinfo{person}{Pan Li}.} \bibinfo{year}{2025}\natexlab{b}.
\newblock \showarticletitle{RoFt-Mol: Benchmarking Robust Fine-Tuning with Molecular Graph Foundation Models}. In \bibinfo{booktitle}{\emph{NeurIPS}}.
\newblock


\bibitem[Liu et~al\mbox{.}(2023)]%
        {liu2023graphprompt}
\bibfield{author}{\bibinfo{person}{Zemin Liu}, \bibinfo{person}{Xingtong Yu}, \bibinfo{person}{Yuan Fang}, {and} \bibinfo{person}{Xinming Zhang}.} \bibinfo{year}{2023}\natexlab{}.
\newblock \showarticletitle{Graphprompt: Unifying pre-training and downstream tasks for graph neural networks}. In \bibinfo{booktitle}{\emph{WWW}}. \bibinfo{pages}{417--428}.
\newblock


\bibitem[Loshchilov and Hutter(2019)]%
        {loshchilovdecoupled}
\bibfield{author}{\bibinfo{person}{Ilya Loshchilov} {and} \bibinfo{person}{Frank Hutter}.} \bibinfo{year}{2019}\natexlab{}.
\newblock \showarticletitle{Decoupled Weight Decay Regularization}. In \bibinfo{booktitle}{\emph{ICLR}}.
\newblock


\bibitem[Lv et~al\mbox{.}(2021)]%
        {lv2021AreWeReally}
\bibfield{author}{\bibinfo{person}{Qingsong Lv}, \bibinfo{person}{Ming Ding}, \bibinfo{person}{Qiang Liu}, \bibinfo{person}{Yuxiang Chen}, \bibinfo{person}{Wenzheng Feng}, \bibinfo{person}{Siming He}, \bibinfo{person}{Chang Zhou}, \bibinfo{person}{Jianguo Jiang}, \bibinfo{person}{Yuxiao Dong}, {and} \bibinfo{person}{Jie Tang}.} \bibinfo{year}{2021}\natexlab{}.
\newblock \showarticletitle{Are we really making much progress?: Revisiting, benchmarking and refining heterogeneous graph neural networks}. In \bibinfo{booktitle}{\emph{KDD}}. \bibinfo{pages}{1150--1160}.
\newblock


\bibitem[Manessi et~al\mbox{.}(2020)]%
        {manessi2020dynamic}
\bibfield{author}{\bibinfo{person}{Franco Manessi}, \bibinfo{person}{Alessandro Rozza}, {and} \bibinfo{person}{Mario Manzo}.} \bibinfo{year}{2020}\natexlab{}.
\newblock \showarticletitle{Dynamic graph convolutional networks}.
\newblock \bibinfo{journal}{\emph{Pattern Recognition}}  \bibinfo{volume}{97} (\bibinfo{year}{2020}), \bibinfo{pages}{107000}.
\newblock


\bibitem[Mao et~al\mbox{.}(2024)]%
        {mao2024position}
\bibfield{author}{\bibinfo{person}{Haitao Mao}, \bibinfo{person}{Zhikai Chen}, \bibinfo{person}{Wenzhuo Tang}, \bibinfo{person}{Jianan Zhao}, \bibinfo{person}{Yao Ma}, \bibinfo{person}{Tong Zhao}, \bibinfo{person}{Neil Shah}, \bibinfo{person}{Mikhail Galkin}, {and} \bibinfo{person}{Jiliang Tang}.} \bibinfo{year}{2024}\natexlab{}.
\newblock \showarticletitle{Position: Graph foundation models are already here}. In \bibinfo{booktitle}{\emph{ICML}}.
\newblock


\bibitem[Marin and Wellman(2011)]%
        {marin2011social}
\bibfield{author}{\bibinfo{person}{Alexandra Marin} {and} \bibinfo{person}{Barry Wellman}.} \bibinfo{year}{2011}\natexlab{}.
\newblock \showarticletitle{Social network analysis: An introduction}.
\newblock \bibinfo{journal}{\emph{The SAGE handbook of social network analysis}}  \bibinfo{volume}{11} (\bibinfo{year}{2011}), \bibinfo{pages}{25}.
\newblock


\bibitem[McLaren and Bruner(2022)]%
        {mclaren2022citation}
\bibfield{author}{\bibinfo{person}{Colin~D McLaren} {and} \bibinfo{person}{Mark~W Bruner}.} \bibinfo{year}{2022}\natexlab{}.
\newblock \showarticletitle{Citation network analysis}.
\newblock \bibinfo{journal}{\emph{International Review of Sport and Exercise Psychology}} \bibinfo{volume}{15}, \bibinfo{number}{1} (\bibinfo{year}{2022}), \bibinfo{pages}{179--198}.
\newblock


\bibitem[Morris et~al\mbox{.}(2020)]%
        {morris2020tudataset}
\bibfield{author}{\bibinfo{person}{Christopher Morris}, \bibinfo{person}{Nils~M Kriege}, \bibinfo{person}{Franka Bause}, \bibinfo{person}{Kristian Kersting}, \bibinfo{person}{Petra Mutzel}, {and} \bibinfo{person}{Marion Neumann}.} \bibinfo{year}{2020}\natexlab{}.
\newblock \showarticletitle{Tudataset: A collection of benchmark datasets for learning with graphs}.
\newblock \bibinfo{journal}{\emph{arXiv preprint arXiv:2007.08663}} (\bibinfo{year}{2020}).
\newblock


\bibitem[Radicchi et~al\mbox{.}(2011)]%
        {radicchi2011citation}
\bibfield{author}{\bibinfo{person}{Filippo Radicchi}, \bibinfo{person}{Santo Fortunato}, {and} \bibinfo{person}{Alessandro Vespignani}.} \bibinfo{year}{2011}\natexlab{}.
\newblock \showarticletitle{Citation networks}.
\newblock \bibinfo{journal}{\emph{Models of science dynamics: Encounters between complexity theory and information sciences}} (\bibinfo{year}{2011}), \bibinfo{pages}{233--257}.
\newblock


\bibitem[Ramp{\'a}{\v{s}}ek et~al\mbox{.}(2022)]%
        {rampavsek2022recipe}
\bibfield{author}{\bibinfo{person}{Ladislav Ramp{\'a}{\v{s}}ek}, \bibinfo{person}{Michael Galkin}, \bibinfo{person}{Vijay~Prakash Dwivedi}, \bibinfo{person}{Anh~Tuan Luu}, \bibinfo{person}{Guy Wolf}, {and} \bibinfo{person}{Dominique Beaini}.} \bibinfo{year}{2022}\natexlab{}.
\newblock \showarticletitle{Recipe for a general, powerful, scalable graph transformer}.
\newblock \bibinfo{journal}{\emph{NeurIPS}} (\bibinfo{year}{2022}), \bibinfo{pages}{14501--14515}.
\newblock


\bibitem[Rossi et~al\mbox{.}(2020)]%
        {tgn_icml_grl2020}
\bibfield{author}{\bibinfo{person}{Emanuele Rossi}, \bibinfo{person}{Ben Chamberlain}, \bibinfo{person}{Fabrizio Frasca}, \bibinfo{person}{Davide Eynard}, \bibinfo{person}{Federico Monti}, {and} \bibinfo{person}{Michael Bronstein}.} \bibinfo{year}{2020}\natexlab{}.
\newblock \showarticletitle{Temporal Graph Networks for Deep Learning on Dynamic Graphs}. In \bibinfo{booktitle}{\emph{ICML 2020 Workshop on Graph Representation Learning}}.
\newblock


\bibitem[Rossi et~al\mbox{.}(2024)]%
        {rossi2024edge}
\bibfield{author}{\bibinfo{person}{Emanuele Rossi}, \bibinfo{person}{Bertrand Charpentier}, \bibinfo{person}{Francesco Di~Giovanni}, \bibinfo{person}{Fabrizio Frasca}, \bibinfo{person}{Stephan G{\"u}nnemann}, {and} \bibinfo{person}{Michael~M Bronstein}.} \bibinfo{year}{2024}\natexlab{}.
\newblock \showarticletitle{Edge directionality improves learning on heterophilic graphs}. In \bibinfo{booktitle}{\emph{Learning on graphs conference}}. \bibinfo{pages}{25--1}.
\newblock


\bibitem[Rozemberczki et~al\mbox{.}(2021)]%
        {rozemberczki2021multi}
\bibfield{author}{\bibinfo{person}{Benedek Rozemberczki}, \bibinfo{person}{Carl Allen}, {and} \bibinfo{person}{Rik Sarkar}.} \bibinfo{year}{2021}\natexlab{}.
\newblock \showarticletitle{Multi-scale attributed node embedding}.
\newblock \bibinfo{journal}{\emph{Journal of Complex Networks}} \bibinfo{volume}{9}, \bibinfo{number}{2} (\bibinfo{year}{2021}), \bibinfo{pages}{cnab014}.
\newblock


\bibitem[Sadeghi and Armanfard(2021)]%
        {sadeghi2021DSSL}
\bibfield{author}{\bibinfo{person}{M. Sadeghi} {and} \bibinfo{person}{N. Armanfard}.} \bibinfo{year}{2021}\natexlab{}.
\newblock \showarticletitle{Deep Successive Subspace Learning for Data Clustering}. In \bibinfo{booktitle}{\emph{IJCNN}}.
\newblock


\bibitem[Shchur et~al\mbox{.}(2018)]%
        {shchur2018pitfalls}
\bibfield{author}{\bibinfo{person}{Oleksandr Shchur}, \bibinfo{person}{Maximilian Mumme}, \bibinfo{person}{Aleksandar Bojchevski}, {and} \bibinfo{person}{Stephan G{\"u}nnemann}.} \bibinfo{year}{2018}\natexlab{}.
\newblock \showarticletitle{Pitfalls of graph neural network evaluation}.
\newblock \bibinfo{journal}{\emph{arXiv preprint arXiv:1811.05868}} (\bibinfo{year}{2018}).
\newblock


\bibitem[Tang et~al\mbox{.}(2022)]%
        {tang2022rethinking}
\bibfield{author}{\bibinfo{person}{Jianheng Tang}, \bibinfo{person}{Jiajin Li}, \bibinfo{person}{Ziqi Gao}, {and} \bibinfo{person}{Jia Li}.} \bibinfo{year}{2022}\natexlab{}.
\newblock \showarticletitle{Rethinking graph neural networks for anomaly detection}. In \bibinfo{booktitle}{\emph{International conference on machine learning}}. PMLR, \bibinfo{pages}{21076--21089}.
\newblock


\bibitem[Tang et~al\mbox{.}(2024a)]%
        {tang2023graphgpt}
\bibfield{author}{\bibinfo{person}{Jiabin Tang}, \bibinfo{person}{Yuhao Yang}, \bibinfo{person}{Wei Wei}, \bibinfo{person}{Lei Shi}, \bibinfo{person}{Lixin Su}, \bibinfo{person}{Suqi Cheng}, \bibinfo{person}{Dawei Yin}, {and} \bibinfo{person}{Chao Huang}.} \bibinfo{year}{2024}\natexlab{a}.
\newblock \showarticletitle{Graphgpt: Graph instruction tuning for large language models}. In \bibinfo{booktitle}{\emph{SIGIR}}. \bibinfo{pages}{491--500}.
\newblock


\bibitem[Tang et~al\mbox{.}(2024b)]%
        {tang2024higpt}
\bibfield{author}{\bibinfo{person}{Jiabin Tang}, \bibinfo{person}{Yuhao Yang}, \bibinfo{person}{Wei Wei}, \bibinfo{person}{Lei Shi}, \bibinfo{person}{Long Xia}, \bibinfo{person}{Dawei Yin}, {and} \bibinfo{person}{Chao Huang}.} \bibinfo{year}{2024}\natexlab{b}.
\newblock \showarticletitle{Hi{GPT}: Heterogeneous graph language model}. In \bibinfo{booktitle}{\emph{KDD}}. \bibinfo{pages}{2842--2853}.
\newblock


\bibitem[Tian et~al\mbox{.}(2021)]%
        {tian2021self}
\bibfield{author}{\bibinfo{person}{Sheng Tian}, \bibinfo{person}{Ruofan Wu}, \bibinfo{person}{Leilei Shi}, \bibinfo{person}{Liang Zhu}, {and} \bibinfo{person}{Tao Xiong}.} \bibinfo{year}{2021}\natexlab{}.
\newblock \showarticletitle{Self-supervised representation learning on dynamic graphs}. In \bibinfo{booktitle}{\emph{CIKM}}. \bibinfo{pages}{1814--1823}.
\newblock


\bibitem[Toutanova and Chen(2015)]%
        {toutanova2015observed}
\bibfield{author}{\bibinfo{person}{Kristina Toutanova} {and} \bibinfo{person}{Danqi Chen}.} \bibinfo{year}{2015}\natexlab{}.
\newblock \showarticletitle{Observed versus latent features for knowledge base and text inference}. In \bibinfo{booktitle}{\emph{Proceedings of the 3rd workshop on continuous vector space models and their compositionality}}. \bibinfo{pages}{57--66}.
\newblock


\bibitem[Touvron et~al\mbox{.}(2023)]%
        {touvron2023llama}
\bibfield{author}{\bibinfo{person}{Hugo Touvron}, \bibinfo{person}{Thibaut Lavril}, \bibinfo{person}{Gautier Izacard}, \bibinfo{person}{Xavier Martinet}, \bibinfo{person}{Marie-Anne Lachaux}, \bibinfo{person}{Timoth{\'e}e Lacroix}, \bibinfo{person}{Baptiste Rozi{\`e}re}, \bibinfo{person}{Naman Goyal}, \bibinfo{person}{Eric Hambro}, \bibinfo{person}{Faisal Azhar}, {et~al\mbox{.}}} \bibinfo{year}{2023}\natexlab{}.
\newblock \showarticletitle{Llama: Open and efficient foundation language models}.
\newblock \bibinfo{journal}{\emph{arXiv preprint arXiv:2302.13971}} (\bibinfo{year}{2023}).
\newblock


\bibitem[Veli{\v{c}}kovi{\'c} et~al\mbox{.}(2018a)]%
        {velivckovic2017graph}
\bibfield{author}{\bibinfo{person}{Petar Veli{\v{c}}kovi{\'c}}, \bibinfo{person}{Guillem Cucurull}, \bibinfo{person}{Arantxa Casanova}, \bibinfo{person}{Adriana Romero}, \bibinfo{person}{Pietro Lio}, {and} \bibinfo{person}{Yoshua Bengio}.} \bibinfo{year}{2018}\natexlab{a}.
\newblock \showarticletitle{Graph attention networks}. In \bibinfo{booktitle}{\emph{ICLR}}.
\newblock


\bibitem[Veli{\v{c}}kovi{\'c} et~al\mbox{.}(2018b)]%
        {velivckovic2018deep}
\bibfield{author}{\bibinfo{person}{Petar Veli{\v{c}}kovi{\'c}}, \bibinfo{person}{William Fedus}, \bibinfo{person}{William~L Hamilton}, \bibinfo{person}{Pietro Li{\`o}}, \bibinfo{person}{Yoshua Bengio}, {and} \bibinfo{person}{R~Devon Hjelm}.} \bibinfo{year}{2018}\natexlab{b}.
\newblock \showarticletitle{Deep Graph Infomax}. In \bibinfo{booktitle}{\emph{ICLR}}.
\newblock


\bibitem[Wang et~al\mbox{.}(2025)]%
        {wang2025multi}
\bibfield{author}{\bibinfo{person}{Shuo Wang}, \bibinfo{person}{Bokui Wang}, \bibinfo{person}{Zhixiang Shen}, \bibinfo{person}{Boyan Deng}, {et~al\mbox{.}}} \bibinfo{year}{2025}\natexlab{}.
\newblock \showarticletitle{Multi-Domain Graph Foundation Models: Robust Knowledge Transfer via Topology Alignment}. In \bibinfo{booktitle}{\emph{ICML}}.
\newblock


\bibitem[Wang et~al\mbox{.}(2023)]%
        {wang2023internimage}
\bibfield{author}{\bibinfo{person}{Wenhai Wang}, \bibinfo{person}{Jifeng Dai}, \bibinfo{person}{Zhe Chen}, \bibinfo{person}{Zhenhang Huang}, \bibinfo{person}{Zhiqi Li}, \bibinfo{person}{Xizhou Zhu}, \bibinfo{person}{Xiaowei Hu}, \bibinfo{person}{Tong Lu}, \bibinfo{person}{Lewei Lu}, \bibinfo{person}{Hongsheng Li}, {et~al\mbox{.}}} \bibinfo{year}{2023}\natexlab{}.
\newblock \showarticletitle{Internimage: Exploring large-scale vision foundation models with deformable convolutions}. In \bibinfo{booktitle}{\emph{CVPR}}. \bibinfo{pages}{14408--14419}.
\newblock


\bibitem[Wang et~al\mbox{.}(2022)]%
        {wang2022survey}
\bibfield{author}{\bibinfo{person}{Xiao Wang}, \bibinfo{person}{Deyu Bo}, \bibinfo{person}{Chuan Shi}, \bibinfo{person}{Shaohua Fan}, \bibinfo{person}{Yanfang Ye}, {and} \bibinfo{person}{Philip~S Yu}.} \bibinfo{year}{2022}\natexlab{}.
\newblock \showarticletitle{A survey on heterogeneous graph embedding: methods, techniques, applications and sources}.
\newblock \bibinfo{journal}{\emph{TBG}} \bibinfo{volume}{9}, \bibinfo{number}{2} (\bibinfo{year}{2022}), \bibinfo{pages}{415--436}.
\newblock


\bibitem[Wang et~al\mbox{.}(2021a)]%
        {wang2021kepler}
\bibfield{author}{\bibinfo{person}{Xiaozhi Wang}, \bibinfo{person}{Tianyu Gao}, \bibinfo{person}{Zhaocheng Zhu}, \bibinfo{person}{Zhengyan Zhang}, \bibinfo{person}{Zhiyuan Liu}, \bibinfo{person}{Juanzi Li}, {and} \bibinfo{person}{Jian Tang}.} \bibinfo{year}{2021}\natexlab{a}.
\newblock \showarticletitle{KEPLER: A unified model for knowledge embedding and pre-trained language representation}.
\newblock \bibinfo{journal}{\emph{Transactions of the Association for Computational Linguistics}}  \bibinfo{volume}{9} (\bibinfo{year}{2021}), \bibinfo{pages}{176--194}.
\newblock


\bibitem[Wang et~al\mbox{.}(2019)]%
        {wang2019heterogeneous}
\bibfield{author}{\bibinfo{person}{Xiao Wang}, \bibinfo{person}{Houye Ji}, \bibinfo{person}{Chuan Shi}, \bibinfo{person}{Bai Wang}, \bibinfo{person}{Yanfang Ye}, \bibinfo{person}{Peng Cui}, {and} \bibinfo{person}{Philip~S. Yu}.} \bibinfo{year}{2019}\natexlab{}.
\newblock \showarticletitle{Heterogeneous Graph Attention Network}. In \bibinfo{booktitle}{\emph{WWW}}. \bibinfo{pages}{2022--2032}.
\newblock


\bibitem[Wang et~al\mbox{.}(2021b)]%
        {wang2021self}
\bibfield{author}{\bibinfo{person}{Xiao Wang}, \bibinfo{person}{Nian Liu}, \bibinfo{person}{Hui Han}, {and} \bibinfo{person}{Chuan Shi}.} \bibinfo{year}{2021}\natexlab{b}.
\newblock \showarticletitle{Self-supervised heterogeneous graph neural network with co-contrastive learning}. In \bibinfo{booktitle}{\emph{KDD}}. \bibinfo{pages}{1726--1736}.
\newblock


\bibitem[Wang et~al\mbox{.}(2024)]%
        {wang2024gft}
\bibfield{author}{\bibinfo{person}{Zehong Wang}, \bibinfo{person}{Zheyuan Zhang}, \bibinfo{person}{Nitesh Chawla}, \bibinfo{person}{Chuxu Zhang}, {and} \bibinfo{person}{Yanfang Ye}.} \bibinfo{year}{2024}\natexlab{}.
\newblock \showarticletitle{Gft: Graph foundation model with transferable tree vocabulary}. In \bibinfo{booktitle}{\emph{NeurIPS}}, Vol.~\bibinfo{volume}{37}. \bibinfo{pages}{107403--107443}.
\newblock


\bibitem[Weber et~al\mbox{.}(2019)]%
        {weber2019anti}
\bibfield{author}{\bibinfo{person}{Mark Weber}, \bibinfo{person}{Giacomo Domeniconi}, \bibinfo{person}{Jie Chen}, \bibinfo{person}{Daniel Karl~I Weidele}, \bibinfo{person}{Claudio Bellei}, \bibinfo{person}{Tom Robinson}, {and} \bibinfo{person}{Charles~E Leiserson}.} \bibinfo{year}{2019}\natexlab{}.
\newblock \showarticletitle{Anti-money laundering in bitcoin: Experimenting with graph convolutional networks for financial forensics}.
\newblock \bibinfo{journal}{\emph{arXiv preprint arXiv:1908.02591}} (\bibinfo{year}{2019}).
\newblock


\bibitem[Wen and Fang(2023)]%
        {wen2023augmenting}
\bibfield{author}{\bibinfo{person}{Zhihao Wen} {and} \bibinfo{person}{Yuan Fang}.} \bibinfo{year}{2023}\natexlab{}.
\newblock \showarticletitle{Augmenting Low-Resource Text Classification with Graph-Grounded Pre-training and Prompting}. In \bibinfo{booktitle}{\emph{SIGIR}}. \bibinfo{pages}{506--516}.
\newblock


\bibitem[Wettig et~al\mbox{.}(2025)]%
        {wettigorganize}
\bibfield{author}{\bibinfo{person}{Alexander Wettig}, \bibinfo{person}{Kyle Lo}, \bibinfo{person}{Sewon Min}, \bibinfo{person}{Hannaneh Hajishirzi}, \bibinfo{person}{Danqi Chen}, {and} \bibinfo{person}{Luca Soldaini}.} \bibinfo{year}{2025}\natexlab{}.
\newblock \showarticletitle{Organize the Web: Constructing Domains Enhances Pre-Training Data Curation}. In \bibinfo{booktitle}{\emph{ICML}}.
\newblock


\bibitem[Wu et~al\mbox{.}(2018)]%
        {wu2018moleculenet}
\bibfield{author}{\bibinfo{person}{Zhenqin Wu}, \bibinfo{person}{Bharath Ramsundar}, \bibinfo{person}{Evan~N Feinberg}, \bibinfo{person}{Joseph Gomes}, \bibinfo{person}{Caleb Geniesse}, \bibinfo{person}{Aneesh~S Pappu}, \bibinfo{person}{Karl Leswing}, {and} \bibinfo{person}{Vijay Pande}.} \bibinfo{year}{2018}\natexlab{}.
\newblock \showarticletitle{MoleculeNet: a benchmark for molecular machine learning}.
\newblock \bibinfo{journal}{\emph{Chemical science}} \bibinfo{volume}{9}, \bibinfo{number}{2} (\bibinfo{year}{2018}), \bibinfo{pages}{513--530}.
\newblock


\bibitem[Xia et~al\mbox{.}(2024)]%
        {xia2024opengraph}
\bibfield{author}{\bibinfo{person}{Lianghao Xia}, \bibinfo{person}{Ben Kao}, {and} \bibinfo{person}{Chao Huang}.} \bibinfo{year}{2024}\natexlab{}.
\newblock \showarticletitle{OpenGraph: Towards Open Graph Foundation Models}. In \bibinfo{booktitle}{\emph{Findings of the Association for Computational Linguistics: EMNLP 2024}}. \bibinfo{pages}{2365--2379}.
\newblock


\bibitem[Xu et~al\mbox{.}(2024)]%
        {xu2024graphfm}
\bibfield{author}{\bibinfo{person}{Yuhao Xu}, \bibinfo{person}{Xinqi Liu}, \bibinfo{person}{Keyu Duan}, \bibinfo{person}{Yi Fang}, \bibinfo{person}{Yu-Neng Chuang}, \bibinfo{person}{Daochen Zha}, {and} \bibinfo{person}{Qiaoyu Tan}.} \bibinfo{year}{2024}\natexlab{}.
\newblock \showarticletitle{Graphfm: A comprehensive benchmark for graph foundation model}.
\newblock \bibinfo{journal}{\emph{arXiv preprint arXiv:2406.08310}} (\bibinfo{year}{2024}).
\newblock


\bibitem[Yang et~al\mbox{.}(2025)]%
        {yang2025benchmarking}
\bibfield{author}{\bibinfo{person}{Jinyu Yang}, \bibinfo{person}{Liangwei Yang}, \bibinfo{person}{Zeyuan Guo}, \bibinfo{person}{Jiayi Gao}, \bibinfo{person}{Jing Wu}, \bibinfo{person}{Tianhao Chai}, \bibinfo{person}{Hai Huang}, \bibinfo{person}{Cheng Yang}, {and} \bibinfo{person}{Chuan Shi}.} \bibinfo{year}{2025}\natexlab{}.
\newblock \showarticletitle{Benchmarking Graph Foundation Models}. In \bibinfo{booktitle}{\emph{SIGKDD}}. \bibinfo{pages}{5866--5875}.
\newblock


\bibitem[Yang et~al\mbox{.}(2016)]%
        {yang2016revisiting}
\bibfield{author}{\bibinfo{person}{Zhilin Yang}, \bibinfo{person}{William Cohen}, {and} \bibinfo{person}{Ruslan Salakhudinov}.} \bibinfo{year}{2016}\natexlab{}.
\newblock \showarticletitle{Revisiting semi-supervised learning with graph embeddings}. In \bibinfo{booktitle}{\emph{International conference on machine learning}}. PMLR, \bibinfo{pages}{40--48}.
\newblock


\bibitem[You et~al\mbox{.}(2018)]%
        {you2018graph}
\bibfield{author}{\bibinfo{person}{Jiaxuan You}, \bibinfo{person}{Bowen Liu}, \bibinfo{person}{Zhitao Ying}, \bibinfo{person}{Vijay Pande}, {and} \bibinfo{person}{Jure Leskovec}.} \bibinfo{year}{2018}\natexlab{}.
\newblock \showarticletitle{Graph convolutional policy network for goal-directed molecular graph generation}. In \bibinfo{booktitle}{\emph{NeurIPS}}, Vol.~\bibinfo{volume}{31}.
\newblock


\bibitem[Yu et~al\mbox{.}(2025a)]%
        {yu2025samgpt}
\bibfield{author}{\bibinfo{person}{Xingtong Yu}, \bibinfo{person}{Zechuan Gong}, \bibinfo{person}{Chang Zhou}, \bibinfo{person}{Yuan Fang}, {and} \bibinfo{person}{Hui Zhang}.} \bibinfo{year}{2025}\natexlab{a}.
\newblock \showarticletitle{Samgpt: Text-free graph foundation model for multi-domain pre-training and cross-domain adaptation}. In \bibinfo{booktitle}{\emph{WWW}}. \bibinfo{pages}{1142--1153}.
\newblock


\bibitem[Yu et~al\mbox{.}(2024a)]%
        {yu2023generalized}
\bibfield{author}{\bibinfo{person}{Xingtong Yu}, \bibinfo{person}{Zhenghao Liu}, \bibinfo{person}{Yuan Fang}, \bibinfo{person}{Zemin Liu}, \bibinfo{person}{Sihong Chen}, {and} \bibinfo{person}{Xinming Zhang}.} \bibinfo{year}{2024}\natexlab{a}.
\newblock \showarticletitle{Generalized Graph Prompt: Toward a Unification of Pre-Training and Downstream Tasks on Graphs}.
\newblock \bibinfo{journal}{\emph{IEEE TKDE}} (\bibinfo{year}{2024}).
\newblock


\bibitem[Yu et~al\mbox{.}(2025b)]%
        {yu2024dygprompt}
\bibfield{author}{\bibinfo{person}{Xingtong Yu}, \bibinfo{person}{Zhenghao Liu}, \bibinfo{person}{Yuan Fang}, {and} \bibinfo{person}{Xinming Zhang}.} \bibinfo{year}{2025}\natexlab{b}.
\newblock \showarticletitle{DyGPrompt: Learning Feature and Time Prompts on Dynamic Graphs}. In \bibinfo{booktitle}{\emph{ICLR}}.
\newblock


\bibitem[Yu et~al\mbox{.}(2025c)]%
        {yu2024non}
\bibfield{author}{\bibinfo{person}{Xingtong Yu}, \bibinfo{person}{Jie Zhang}, \bibinfo{person}{Yuan Fang}, {and} \bibinfo{person}{Renhe Jiang}.} \bibinfo{year}{2025}\natexlab{c}.
\newblock \showarticletitle{Non-Homophilic Graph Pre-Training and Prompt Learning}. In \bibinfo{booktitle}{\emph{SIGKDD}}. \bibinfo{pages}{1844--1854}.
\newblock


\bibitem[Yu et~al\mbox{.}(2024b)]%
        {yu2024text}
\bibfield{author}{\bibinfo{person}{Xingtong Yu}, \bibinfo{person}{Chang Zhou}, \bibinfo{person}{Yuan Fang}, {and} \bibinfo{person}{Xinming Zhang}.} \bibinfo{year}{2024}\natexlab{b}.
\newblock \showarticletitle{Text-Free Multi-domain Graph Pre-training: Toward Graph Foundation Models}.
\newblock \bibinfo{journal}{\emph{arXiv preprint arXiv:2405.13934}} (\bibinfo{year}{2024}).
\newblock


\bibitem[Yuan et~al\mbox{.}(2025)]%
        {yuangraver}
\bibfield{author}{\bibinfo{person}{Haonan Yuan}, \bibinfo{person}{Qingyun Sun}, \bibinfo{person}{Junhua Shi}, \bibinfo{person}{Xingcheng Fu}, \bibinfo{person}{Bryan Hooi}, \bibinfo{person}{Jianxin Li}, {and} \bibinfo{person}{Philip~S Yu}.} \bibinfo{year}{2025}\natexlab{}.
\newblock \showarticletitle{GRAVER: Generative Graph Vocabularies for Robust Graph Foundation Models Fine-tuning}. In \bibinfo{booktitle}{\emph{NeurIPS}}.
\newblock


\bibitem[Yuan et~al\mbox{.}(2021)]%
        {yuan2021florence}
\bibfield{author}{\bibinfo{person}{Lu Yuan}, \bibinfo{person}{Dongdong Chen}, \bibinfo{person}{Yi-Ling Chen}, \bibinfo{person}{Noel Codella}, \bibinfo{person}{Xiyang Dai}, \bibinfo{person}{Jianfeng Gao}, \bibinfo{person}{Houdong Hu}, \bibinfo{person}{Xuedong Huang}, \bibinfo{person}{Boxin Li}, \bibinfo{person}{Chunyuan Li}, {et~al\mbox{.}}} \bibinfo{year}{2021}\natexlab{}.
\newblock \showarticletitle{Florence: A new foundation model for computer vision}.
\newblock \bibinfo{journal}{\emph{arXiv preprint arXiv:2111.11432}} (\bibinfo{year}{2021}).
\newblock


\bibitem[Zhang et~al\mbox{.}(2019)]%
        {zhang2019heterogeneous}
\bibfield{author}{\bibinfo{person}{Chuxu Zhang}, \bibinfo{person}{Dongjin Song}, \bibinfo{person}{Chao Huang}, \bibinfo{person}{Ananthram Swami}, {and} \bibinfo{person}{Nitesh~V Chawla}.} \bibinfo{year}{2019}\natexlab{}.
\newblock \showarticletitle{Heterogeneous graph neural network}. In \bibinfo{booktitle}{\emph{SIGKDD}}. \bibinfo{pages}{793--803}.
\newblock


\bibitem[Zhao et~al\mbox{.}(2024a)]%
        {zhao2024all}
\bibfield{author}{\bibinfo{person}{Haihong Zhao}, \bibinfo{person}{Aochuan Chen}, \bibinfo{person}{Xiangguo Sun}, \bibinfo{person}{Hong Cheng}, {and} \bibinfo{person}{Jia Li}.} \bibinfo{year}{2024}\natexlab{a}.
\newblock \showarticletitle{All in one and one for all: A simple yet effective method towards cross-domain graph pretraining}. In \bibinfo{booktitle}{\emph{KDD}}. \bibinfo{pages}{4443--4454}.
\newblock


\bibitem[Zhao et~al\mbox{.}(2024b)]%
        {zhao2024graphany}
\bibfield{author}{\bibinfo{person}{Jianan Zhao}, \bibinfo{person}{Hesham Mostafa}, \bibinfo{person}{Michael Galkin}, \bibinfo{person}{Michael Bronstein}, \bibinfo{person}{Zhaocheng Zhu}, {and} \bibinfo{person}{Jian Tang}.} \bibinfo{year}{2024}\natexlab{b}.
\newblock \showarticletitle{Graphany: A foundation model for node classification on any graph}.
\newblock \bibinfo{journal}{\emph{arXiv preprint arXiv:2405.20445}}  \bibinfo{volume}{29} (\bibinfo{year}{2024}).
\newblock


\bibitem[Zhao et~al\mbox{.}(2023)]%
        {zhaolearning}
\bibfield{author}{\bibinfo{person}{Jianan Zhao}, \bibinfo{person}{Meng Qu}, \bibinfo{person}{Chaozhuo Li}, \bibinfo{person}{Hao Yan}, \bibinfo{person}{Qian Liu}, \bibinfo{person}{Rui Li}, \bibinfo{person}{Xing Xie}, {and} \bibinfo{person}{Jian Tang}.} \bibinfo{year}{2023}\natexlab{}.
\newblock \showarticletitle{Learning on Large-scale Text-attributed Graphs via Variational Inference}. In \bibinfo{booktitle}{\emph{ICLR}}.
\newblock


\bibitem[Zhao et~al\mbox{.}(2020)]%
        {zhao2020persistence}
\bibfield{author}{\bibinfo{person}{Qi Zhao}, \bibinfo{person}{Ze Ye}, \bibinfo{person}{Chao Chen}, {and} \bibinfo{person}{Yusu Wang}.} \bibinfo{year}{2020}\natexlab{}.
\newblock \showarticletitle{Persistence enhanced graph neural network}. In \bibinfo{booktitle}{\emph{ICAIS}}. \bibinfo{pages}{2896--2906}.
\newblock


\bibitem[Zhu et~al\mbox{.}(2025)]%
        {zhu2025graphclip}
\bibfield{author}{\bibinfo{person}{Yun Zhu}, \bibinfo{person}{Haizhou Shi}, \bibinfo{person}{Xiaotang Wang}, \bibinfo{person}{Yongchao Liu}, \bibinfo{person}{Yaoke Wang}, \bibinfo{person}{Boci Peng}, \bibinfo{person}{Chuntao Hong}, {and} \bibinfo{person}{Siliang Tang}.} \bibinfo{year}{2025}\natexlab{}.
\newblock \showarticletitle{Graphclip: Enhancing transferability in graph foundation models for text-attributed graphs}. In \bibinfo{booktitle}{\emph{WWW}}. \bibinfo{pages}{2183--2197}.
\newblock


\end{thebibliography}

\clearpage
\newpage
\appendix
\section*{Appendices}
\renewcommand\thesubsection{\Alph{subsection}}
\renewcommand\thesubsubsection{\thesubsection.\arabic{subsection}}
\begin{figure*}[t]
\centering
\includegraphics[width=1\linewidth]{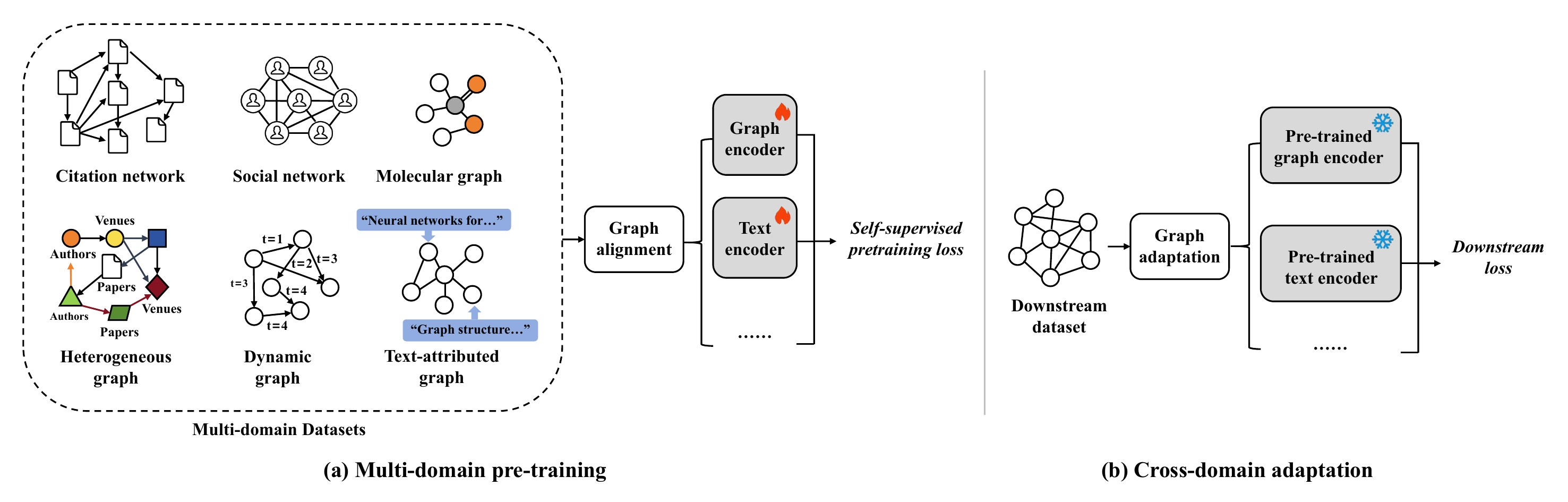}
\caption{Graph foundation models pipeline.}
\label{fig.gfm-pipeline}
\end{figure*}

\subsection{GFMs Pipeline}\label{app.pipeline}
We illustrate the GFMs pipeline in Fig.~\ref{fig.gfm-pipeline}.

\subsection{Benchmark Pipeline}\label{app.ben-pipeline}
\begin{figure*}[t]
\centering
\includegraphics[width=1\linewidth]{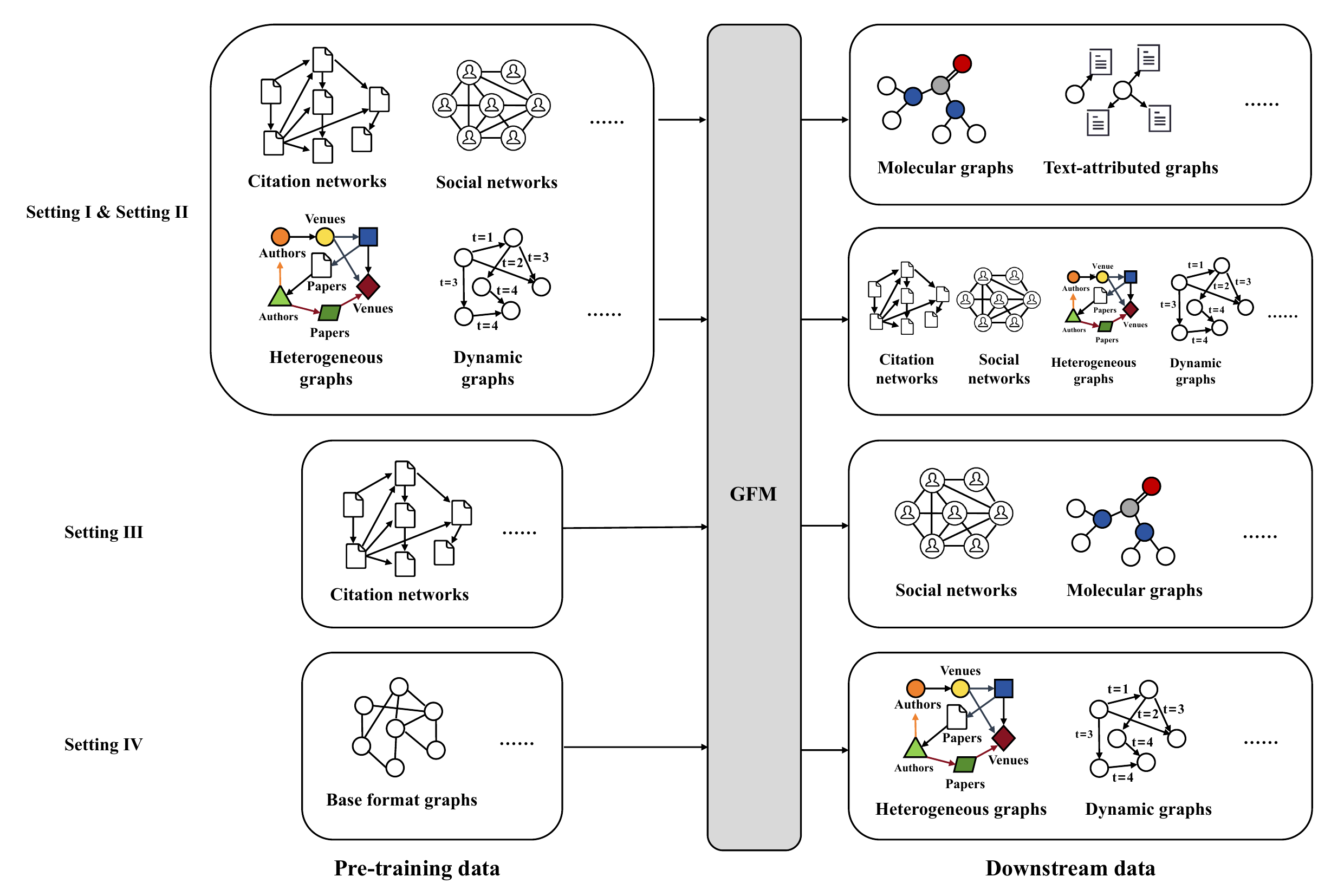}
\caption{Four evaluation settings in our benchmark.}
\label{fig.ben-pipeline}
\end{figure*}

We illustrate the four evaluation settings in Fig.~\ref{fig.ben-pipeline}.

\subsection{Evaluated and GFM-style Methods}\label{app.baselines}

We summarize the representative GFMs in Table~\ref{table.gfms}. The column \textit{Labeled source} indicates whether a method relies on supervision, such as labeled objectives or instruction-tuning signals, during pre-training. The \textit{Alignment} columns categorize the primary mechanisms used to reduce domain gaps: \textit{Structure} and \textit{Feature} denote whether the method explicitly aligns cross-domain structural patterns or feature spaces, respectively; \textit{Codebook} indicates whether a learnable codebook is introduced for discretization or alignment; \textit{Graph mod.} and \textit{Feature mod.} indicate whether the method modifies the input graph topology or transforms features to facilitate alignment; and \textit{Textual} indicates whether textual information is used to bridge domain gaps. The \textit{PLM/LLM} column indicates whether a pretrained language model is a core component (e.g., for feature extraction, cross-modal alignment, prediction, or graph/text generation). \textit{Adaptation} specifies the primary downstream adaptation strategy, such as fine-tuning or prompting.

\stitle{Evaluated GFMs.}
We evaluate a set of representative GFMs that embody different mechanism designs for mitigating domain gaps. Concretely, we include:
\begin{itemize}[leftmargin=1.2em]
    \item \method{GCOPE}~\cite{zhao2024all} builds a joint training graph by adding a learnable coordinator (virtual) node for each domain and connecting it to all nodes in that domain and to other coordinators to align various domains.
    \item \method{MDGPT}~\cite{yu2024text} introduces domain tokens to harmonize heterogeneous feature spaces, and adopts a dual-prompt design to condition the encoder on domain-specific characteristics during adaptation.
    \item \method{SAMGPT}~\cite{yu2025samgpt} extends MDGPT by explicitly addressing structural discrepancies across domains. It injects structure-aware tokens and prompts to modulate the message-passing process in the graph encoder, thereby unifying domain-dependent structural inductive biases.
    \item \method{MDGFM}~\cite{wang2025multi} propose a topology-alignment framework that explicitly matches cross-domain graph structural patterns to learn domain-invariant graph representations, enabling robust knowledge structural transfer.
    \item \method{G2P2}~\cite{wen2023augmenting} jointly pre-trains a text encoder and a graph encoder via graph-grounded contrastive learning.  It leverages a document-node bijection and neighborhood-derived summary text to construct positive pairs, and optimizes multiple interaction losses. %(text--node, text--summary, and node--summary alignment).
    \item \method{GraphCLIP} \cite{zhu2025graphclip} employs a LLM to generate natural language summaries for sampled subgraphs, then pre-trains a graph encoder with a CLIP-style contrastive objective that aligns graph embeddings with summary text embeddings.
    \item \method{GFT} \cite{wang2024gft} learns a transferable tree vocabulary by treating the message-passing computation trees induced by a GNN as discrete structural tokens, and pretrains a single backbone with a computation-tree reconstruction objective so it can transfer robustly across domains and tasks while mitigating negative transfer.
    \item \method{UniGraph2}~\cite{he2025unigraph2} learns a unified embedding space for multi-modal graphs by binding heterogeneous modalities into a shared representation, enabling consistent cross-modal semantics and structure-aware fusion.
\end{itemize}

\stitle{Other representative GFM-style methods.} We further discuss other GFM-style methods. Note that these methods are excluded from our primary evaluation as they typically require labels in the pre-training datasets, which diverges from our experimental setting. Concretely, we discuss:
\begin{itemize}[leftmargin=1.2em]
    \item \method{OFA}~\cite{liu2023one} unifies diverse graph learning tasks into a single framework by representing all data as TAGs and introducing nodes-of-interest prompts to standardize task specifications across domains.
    \item \method{Prodigy}~\cite{huang2023prodigy} introduces an in-context learning framework that unifies diverse tasks using a prompt graph representation, and optimizes a self-supervised neighbor matching objective to facilitate out-of-the-box generalization.
    \item \method{GraphGPT}~\cite{tang2023graphgpt} aligns graph structural knowledge with LLMs via a dual-stage instruction tuning paradigm, utilizing a self-supervised graph matching task for structural grounding and Chain-of-Thought distillation to enhance step-by-step reasoning and generalization capabilities. 
    \item \method{HiGPT}~\cite{tang2024higpt} extends GraphGPT by explicitly addressing relation heterogeneity and distribution shifts across domains. It introduces an in-context heterogeneous graph tokenizer and Mixture-of-Thought instructions to align diverse node and edge semantics, thereby unifying domain-dependent heterogeneity patterns for zero-shot adaptation.
    \item \method{OpenGraph}~\cite{xia2024opengraph} develops a scalable zero-shot foundation model that employs a unified topological tokenizer and graph transformer to resolve node shifts and capture dependencies, while utilizing LLM-based augmentation to mitigate data scarcity.
    \item \method{GRAVER}~\cite{yuangraver} stabilizes few-shot fine-tuning by augmenting support samples with adaptive graph vocabularies generated via graphon-based experts and a mixture-of-experts routing network.
    \item \method{Llaga}~\cite{chen2024llaga} adapts frozen LLMs by organizing node neighborhoods into structure-preserving sequences via parameter-free templates and mapping them to the token space using a versatile projector trained on unified QA tasks.
    \item \method{ZeroG}~\cite{li2024zerog} facilitates cross-dataset zero-shot transfer by mapping node attributes and class descriptions into a unified semantic space via a pre-trained LM, utilizing prompt-based subgraph sampling to mitigate negative transfer and reformulating classification as a text similarity task.
    \item \method{GraphAny}~\cite{zhao2024graphany} establishes a fully inductive foundation model for node classification on arbitrary graphs by deriving analytical solutions via LinearGNNs and fusing them using an attention module based on entropy-normalized distance features.
\end{itemize}

\stitle{Supervised learning methods.} We evaluate representative supervised GNNs as task-specific baselines, tailed for homogeneous, heterogeneous, and heterophilic graphs:
\begin{itemize}[leftmargin=1.2em]
    \item \method{GCN}~\cite{kipf2016semi} employs a normalized neighborhood aggregation that effectively acts as mean-pooling over adjacent nodes’ features, enabling integration of local structural information under the homophily assumption.
    \item \method{GAT}~\cite{velivckovic2017graph} enhances neighborhood aggregation by incorporating a learnable attention mechanism that assigns distinct importance weights to each neighbor, enabling the model to emphasize more relevant connections during message passing adaptively. 
    \item \method{SIMPLE-HGN}~\cite{lv2021AreWeReally} extends attention-based aggregation to heterogeneous graphs by incorporating edge type embeddings into the attention computation, allowing relation-aware inter-type message passing.
    \item \method{FAGCN}~\cite{bo2021beyond} introduces a frequency adaptation mechanism with a self-gating filter that adaptively integrates both low-frequency (smooth) and high-frequency (discriminative) signals in propagation, thereby enhancing robustness in heterophilic settings.
\end{itemize}

\stitle{Single-domain pre-training and adaptation methods.} We further compare against state-of-the-art self-supervised pre-training methods. They are pre-trained on a single domain and adapt to the same domain. These methods are categorized by their adaptation paradigms, covering both pre-training \& fine-tuning frameworks and prompt tuning strategies designed for downstream adaptation:
\begin{itemize}[leftmargin=1.2em]
    \item \method{DGI}~\cite{velivckovic2018deep} a self-supervised pre-training approach that learns node representations by maximizing the mutual information between local node (patch) embeddings and a high-level global summary of the graph, thereby capturing both local and global structural context in an unsupervised manner.
    \item \method{GraphPrompt}~\cite{liu2023graphprompt} unifies pre‑training and downstream tasks into a common subgraph similarity learning template. It uses a learnable prompt during pre‑training to guide extraction of task‑relevant knowledge and tunes the prompt during adaptation to adjust aggregation behavior so that the model better leverages pre‑trained representations for specific tasks such as node and graph classification.
    \item \method{HeCo}~\cite{wang2021self} proposes a co‑contrastive learning mechanism for heterogeneous graphs by jointly contrasting node representations from network schema and meta‑path views so that the views collaboratively supervise each other to capture both local typed structure and high‑order semantics. 
    \item \method{TGN}~\cite{tgn_icml_grl2020} introduces a general framework for continuous‑time dynamic graphs. It maintains a memory vector for each node that is updated upon interactions to capture evolving history, and combines temporal encoding with event‑driven message propagation to produce time‑aware node embeddings reflecting both structural and temporal dependencies. 
    \item \method{DDGCL}~\cite{tian2021self} proposes a debiased contrastive learning objective for dynamic graphs that alleviates sampling bias in instance discrimination by adjusting how positive and negative temporal views are contrasted.
    \item \method{DSSL}~\cite{sadeghi2021DSSL} proposes an autoencoder-based deep clustering approach that jointly optimizes reconstruction and clustering objectives, weighting data points according to their similarity to cluster centers.
\end{itemize}

\subsection{Datasets Description}\label{app.datasets}
In this section, we provide comprehensive descriptions of the datasets used in this benchmark, summarized in Table~\ref{table.datasets}.
\begin{itemize}[leftmargin=1.2em]
    \item \textit{Cora}\footnote{\url{https://github.com/LechengKong/OneForAll/blob/main/data/single_graph/Cora/cora.pt}} 
    is a citation network representing papers in the computer science domain, first introduced by~\cite{chen2023exploring}. In this graph, nodes utilize titles and abstracts as raw features, edges indicate citation relationships, and labels correspond to paper categories. We employ the version provided by~\cite{liu2023one}, which reformats these text features as node features and generates detailed descriptions using GPT-3.5-Turbo (ChatGPT).
    \item \textit{Pubmed}\footnote{\url{https://github.com/LechengKong/OneForAll/blob/main/data/single_graph/Pubmed/pubmed.pt}} 
    is a citation network representing papers in the biomedical domain, first introduced by~\cite{chen2023exploring}. Similar to \textit{Cora}, nodes utilize titles and abstracts as raw features, and edges indicate citation relationships. The labels correspond to three specific categories: diabetes mellitus experimental, diabetes mellitus type 1, and diabetes mellitus Type 2. We employ the version provided by~\cite{liu2023one}, which uses the same processing procedure as \textit{Cora}.
    \item \textit{ogbn-arxiv}\footnote{\url{https://snap.stanford.edu/ogb/data/misc/ogbn_arxiv}} 
    is a citation network representing Computer Science arXiv papers, provided by the Open Graph Benchmark (OGB)~\cite{hu2020open}. In this graph, nodes correspond to papers and edges denote citation relationships. The task is to predict the primary subject area of each paper across 40 distinct categories (e.g., cs.AI).
    \item \textit{ogbn-mag}\footnote{\url{https://pytorch-geometric.readthedocs.io/en/latest/generated/torch_geometric.datasets.OGB_MAG.html}} 
    is a heterogeneous network composed of a subset of the Microsoft Academic Graph (MAG), provided by the OGB benchmark. It contains four types of entities: papers, authors, institutions, and fields of study. The edges represent relationships, including ``affiliated with'', ``writes'',  ``cites'', and ``has topics''. In this graph, paper nodes are associated with 128-dimensional word2vec feature vectors. For other entity types that lack raw features, we employ MetaPath2Vec~\cite{dong2017metapath2vec}  to generate their input embeddings. The task is to predict the venue (conference or journal) of each paper node across 349 distinct categories. 
    \item \textit{ACM}\footnote{\url{https://pytorch-geometric.readthedocs.io/en/latest/generated/torch_geometric.datasets.HGBDataset.html}} 
    is a heterogeneous citation network provided by PyTorch Geometric (PyG), sourced from~\cite{lv2021AreWeReally}. It contains four types of entities: papers, authors, subjects, and terms. The edges represent relationships, including ``cites'', ``writes'', ``belongs to'', and ``mentions''. The task is to predict the primary category of each paper across three distinct classes: Database, Wireless Communication, and Data Mining.
    \item \textit{DBLP}\footnote{\url{https://pytorch-geometric.readthedocs.io/en/latest/generated/torch_geometric.datasets.HGBDataset.html}} 
    is a heterogeneous bibliography network provided by PyG, sourced from~\cite{lv2021AreWeReally}. It contains four types of entities: authors, papers, terms, and venues. The edges represent relationships, including ``writes'', ``mentions'', and ``published in''. The task is to predict the research area of each author node across four distinct categories: Database, Data Mining, Artificial Intelligence, and Information Retrieval.
    \item \textit{Reddit}\footnote{\url{https://pytorch-geometric.readthedocs.io/en/latest/generated/torch_geometric.datasets.JODIEDataset.html}} 
    is a temporal interaction network provided by PyG, sourced from~\cite{kumar2019predicting}. It contains two types of entities: users and posts. The edges represent relationships, including ``interacts''. The task is to predict the state of each user node, specifically whether the user is banned.
    \item \textit{Wikipedia}\footnote{\url{https://pytorch-geometric.readthedocs.io/en/latest/generated/torch_geometric.datasets.JODIEDataset.html}} 
    is a temporal interaction network provided by PyG, sourced from~\cite{kumar2019predicting}. It contains two types of entities: users and pages. The edges represent relationships, including ``edits''. The task is to predict the state of each user node, specifically whether the user is banned.
    \item \textit{Actor}\footnote{\url{https://github.com/honey0219/LLM4HeG/blob/main/dataset/Actor.npz}} 
    is a heterophilic actor network provided by ~\cite{li2025hetgb}. It contains nodes representing actors extracted from Wikipedia. The edges indicate co-occurrence relationships on the same Wikipedia page. The task is to predict the category of each actor across five distinct classes, such as ``American film actors'' and ``English actors''.
    \item \textit{Texas}\footnote{\url{https://github.com/honey0219/LLM4HeG/tree/main/dataset/Texas.npz}}, \textit{Wisconsin}\footnote{\url{https://github.com/honey0219/LLM4HeG/blob/main/dataset/Wisconsin.npz}},  \textit{Cornell}\footnote{\url{https://github.com/honey0219/LLM4HeG/tree/main/dataset/Cornell.npz}} 
    are heterophilic web page networks provided by ~\cite{li2025hetgb}. It contains nodes representing web pages from the computer science department of the three universities. The edges represent hyperlinks between pages. The task is to predict the category of each web page across five distinct classes: student, project, course, staff, and faculty.
    \item \textit{Chameleon}\footnote{\url{https://pytorch-geometric.readthedocs.io/en/latest/generated/torch_geometric.datasets.WikipediaNetwork.html}}
    is a Wikipedia page network provided by PyG, sourced from~\cite{rozemberczki2021multi}. It contains nodes representing web pages related to the topic of chameleons, and edges representing hyperlinks between them. The task is to predict the average daily traffic of each web page across five distinct categories.
    \item \textit{IMDB}\footnote{\url{https://pytorch-geometric.readthedocs.io/en/latest/generated/torch_geometric.datasets.HGBDataset.html}} 
    is a heterogeneous movie network provided by PyG, sourced from~\cite{lv2021AreWeReally}. It comprises four entity types: movies, actors, directors, and keywords. The edges indicate relationships including ``directed by'', ``acted in'', and ``described by''. The task is to predict the genre of each movie node across five distinct categories: Action, Comedy, Drama, Romance, and Thriller. 
    \item \textit{Computers}\footnote{\url{https://pytorch-geometric.readthedocs.io/en/latest/generated/torch_geometric.datasets.Amazon.html}}  and \textit{Photo}\footnote{\url{https://pytorch-geometric.readthedocs.io/en/latest/generated/torch_geometric.datasets.Amazon.html}}
    are segments of the Amazon co-purchase graph provided by PyG, sourced from~\cite{shchur2018pitfalls}. In these graphs, nodes represent goods, and edges indicate that two goods are frequently bought together. The task is to predict the respective product category for each good.
    \item \textit{Amazon}\footnote{\url{https://github.com/SY1706203/GATNE/tree/master/dataset/amazon}}
    is an heterogeneous network sourced from~\cite{cen2019representation}, specifically focusing on the Electronics category. In this graph, nodes represent products, and edges denote distinct types of heterogeneous relationships, including co-viewing and co-purchasing links. The dataset incorporates product metadata such as price, sales-rank, brand, and category.
    \item \textit{Amazon-HeTGB}\footnote{\url{https://github.com/honey0219/LLM4HeG/tree/main/dataset/Amazon.npz}} 
    is a heterophilic product co-purchasing network provided by ~\cite{li2025hetgb}. It contains nodes representing products such as books, music CDs, DVDs, and VHS tapes. The edges indicate relationships where products are frequently bought together. The task is to predict the average rating of each product across five distinct classes.
    \item \textit{Products}\footnote{\url{https://utexas.app.box.com/s/i7y03rzm40xt9bjbaj0dfdgxeyjx77gb/file/1441818321356}} 
    is a co-purchasing network employed by~\cite{chen2024llaga}, originally derived from the OGB benchmark. In this graph, nodes represent products sold on Amazon, and edges indicate co-purchasing relationships between them. The task is to predict the category of each product across 47 distinct classes.
    \item \textit{Wiki}\footnote{\url{https://pytorch-geometric.readthedocs.io/en/latest/generated/torch_geometric.datasets.Wikidata5M.html}} 
    is a large-scale knowledge graph provided by PyG, sourced from~\cite{wang2021kepler}. It contains nodes representing real-world entities derived from Wikipedia and Wikidata, while edges denote relational facts connecting them.  The task is typically knowledge graph completion, aiming to predict missing links between entities.
    \item \textit{FB15K-237}\footnote{\url{https://pykeen.readthedocs.io/en/stable/api/pykeen.datasets.FB15k237.html}}
    is a knowledge graph dataset provided by the PyKEEN library, sourced from~\cite{toutanova2015observed}. It is a subset of the Freebase knowledge base, constructed to remove inverse relations found in the original FB15k dataset to prevent data leakage. In this graph, nodes represent general knowledge entities, and edges denote semantic relationships. The task is link prediction, aiming to infer missing relationships between entities.
    \item \textit{WN18RR}\footnote{\url{https://pykeen.readthedocs.io/en/stable/api/pykeen.datasets.WN18RR.html}}
    is a knowledge graph dataset provided by the PyKEEN library, sourced from~\cite{dettmers2018convolutional}. It is a subset of WordNet, derived from the WN18 dataset by removing inverse relations. In this graph, nodes correspond to English words or synsets, and edges represent lexical relationships such as hypernyms and hyponyms. Similar to \textit{FB15K-237}, the primary task is link prediction.
    \item \textit{NELL}\footnote{\url{https://pytorch-geometric.readthedocs.io/en/latest/generated/torch_geometric.datasets.NELL.html}}
    is a knowledge graph dataset provided by PyG, sourced from~\cite{carlson2010toward} and processed by~\cite{yang2016revisiting}. It is a subset of data extracted from the Never-Ending Language Learning project. In this graph, nodes represent entities, and edges indicate semantic relationships between them. The task is to predict the category of each entity node across distinct classes.
    \item \textit{T-Finance}\footnote{\url{https://github.com/squareRoot3/Rethinking-Anomaly-Detection/tree/master}}
    is a financial transaction network sourced from~\cite{tang2022rethinking}. In this graph, nodes represent accounts, and edges denote transaction relationships between them. The node features consist of various transaction attributes and statistical information. The task is anomaly detection, aiming to classify each account node as either normal or anomalous (fraudulent).
    \item \textit{Elliptic}\footnote{\url{https://pytorch-geometric.readthedocs.io/en/latest/generated/torch_geometric.datasets.EllipticBitcoinDataset.html}}
    is a Bitcoin transaction network provided by PyG, sourced from~\cite{weber2019anti}. In this graph, nodes correspond to confirmed Bitcoin transactions, and directed edges represent the flow of Bitcoin currency between them. Each node is associated with 166 features, including time steps, local transaction information, and aggregated neighbor features. The task is to classify transactions into two categories: licit or illicit.
    \item \textit{DGraph}\footnote{\url{https://pytorch-geometric.readthedocs.io/en/latest/generated/torch_geometric.datasets.DGraphFin.html}}
    is a large-scale financial social network provided by PyG, sourced from~\cite{huang2022dgraph}. It represents a real-world graph from Finvolution, where nodes correspond to users, and edges denote 11 distinct interaction relationships such as emergency contacts. The nodes are associated with anonymized features reflecting user profiles and behavioral information. The task is anomaly detection, aiming to distinguish between normal users and fraudsters.
    \item \textit{HIV}\footnote{\url{https://pytorch-geometric.readthedocs.io/en/latest/generated/torch_geometric.datasets.MoleculeNet.html}}  and \textit{PCBA}\footnote{\url{https://pytorch-geometric.readthedocs.io/en/latest/generated/torch_geometric.datasets.MoleculeNet.html}} 
    are large-scale molecular graph datasets provided by PyG, sourced from~\cite{wu2018moleculenet}. In these graphs, nodes represent atoms, and edges denote chemical bonds. The task is graph classification, aiming to predict specific biological properties, such as the ability to inhibit HIV replication or biological activity against various targets.
    \item \textit{COX2}\footnote{\url{https://pytorch-geometric.readthedocs.io/en/latest/generated/torch_geometric.datasets.TUDataset.html}}  and \textit{BZR}\footnote{\url{https://pytorch-geometric.readthedocs.io/en/latest/generated/torch_geometric.datasets.TUDataset.html}}  
    are molecular graph datasets provided by PyG, sourced from~\cite{morris2020tudataset}. Each graph represents a chemical compound, where nodes correspond to atoms and edges represent chemical bonds. The task is binary graph classification, predicting whether the molecule is active or inactive against specific receptors.
    \item \textit{PROTEINS}\footnote{\url{https://pytorch-geometric.readthedocs.io/en/latest/generated/torch_geometric.datasets.TUDataset.html}}  and \textit{ENZYMES}\footnote{\url{https://pytorch-geometric.readthedocs.io/en/latest/generated/torch_geometric.datasets.TUDataset.html}} 
    are bioinformatics datasets provided by PyG, sourced from~\cite{morris2020tudataset}. These graphs represent protein tertiary structures, where nodes correspond to secondary structure elements (helices, sheets, or turns) and edges indicate spatial proximity or chemical connections. The task is graph classification, predicting the functional category of the proteins.
    \item \textit{ogbn-proteins}\footnote{\url{https://ogb.stanford.edu/docs/nodeprop}} 
    is an undirected, weighted protein-protein association network sourced from the OGB benchmark. In this graph, nodes represent proteins across eight different species, and edges denote biologically meaningful associations, such as physical interactions or co-expression. Each edge is associated with 8-dimensional features representing the confidence scores of different association types. The task is multi-label node classification, aiming to predict the presence of 112 distinct protein functions.
\end{itemize}

\subsection{Data Preparation}\label{app.pre-process}

In this section, we elaborate on the data pre-processing protocols introduced in Section~\ref{sec.setting}. To bridge the gap between diverse raw data formats (e.g., temporal streams, heterogeneous schemas) and the unified input requirements of GFMs, we implement a standardized data preparation pipeline. This pipeline converts all datasets into a unified graph object while preserving structural and semantic integrity.

\stitle{Unified data object interface.} Regardless of the original format, every pre-processed graph is encapsulated in a unified data object. Let $N$ be the number of nodes and $M$ be the number of edges. The object consists of the following standardized attributes:
\begin{itemize}[leftmargin=1.2em]
    \item \texttt{x}: Node feature matrix $\mathbf{X} \in \mathbb{R}^{N \times F}$. Defaulted to one-hot vector if unavailable.
    \item \texttt{edge\_index}: Sparse adjacency matrix $\mathbf{E} \in \mathbb{Z}^{2 \times M}$.
    \item \texttt{edge\_attr}: Edge feature matrix $\mathbf{E}_{attr} \in \mathbb{R}^{M \times D}$. Defaulted to zero vectors if unavailable.
    \item \texttt{node\_type}: Integer Tensor $\mathbf{T}_{node} \in \mathbb{Z}^{N}$ mapping each node to its original heterogeneous type ID. Defaults to a zero-filled tensor for homogeneous graphs.
    \item \texttt{edge\_type}: Integer Tensor $\mathbf{T}_{edge} \in \mathbb{Z}^{M}$ mapping each edge to its relation type ID. Defaults to a zero-filled tensor if relation types are unspecified.
    \item \texttt{batch}: Integer Tensor $\mathbf{B} \in \mathbb{Z}^{N}$ mapping nodes to graph indices within a mini-batch (only present in multi-graph datasets).
    \item \texttt{raw\_texts}: A list of length $N$ containing raw text for each node (e.g., titles, abstracts, or type texts). Defaulted to `N/A' if unavailable.
    \item \texttt{relation\_texts}: A list of length $K$ (number of edge types) describing the semantics of relations (e.g., ``cites'', ``co-authors''). Defaults to `to' if unavailable.
    \item \texttt{label\_names}: A list of natural language names of target classes.
    \item \texttt{label\_descs}: A list of detailed textual descriptions of target classes to support zero-shot inference.
    \item \texttt{name}: A unique string denoting the name of the dataset.
\end{itemize}
Notably, we only extract the necessary attributes from this unified object to serve as inputs, tailored to the specific requirements of each model. A distinct exception is \method{GraphCLIP}, which ignores the original feature matrix and instead generates \texttt{x} by encoding the \texttt{raw\_texts} of each node using the \method{all-MiniLM-L6-v2} Sentence Transformer, deriving embeddings via mean pooling of the last hidden states. Table~\ref{tab:raw_text_examples} provides specific examples of these \texttt{raw\_texts} and \texttt{relation\_texts} attributes across varying domains.

\begin{table*}[tbp]
    \centering
    \footnotesize
    \caption{Examples of standardized \texttt{raw\_texts} and \texttt{relation\_texts} encapsulated in our unified data objects.}
    \addtolength{\tabcolsep}{1.0mm}
    \label{tab:raw_text_examples}
    \resizebox{1\linewidth}{!}{
    \begin{tabular}{l|p{9cm}|p{5cm}}
    \toprule
    \textbf{Dataset} & \textbf{\texttt{raw\_texts}} & \textbf{\texttt{relation\_texts}} \\
    \midrule\midrule
    Cora, ogbn-arxiv & Title: "Further facts..." Abstract: "Previous results..." / Title: "Spring embedders..." Abstract: "Force-directed..." & cites \\
    \midrule
    ACM & author, term, subject, paper & cites, mentions, writes, belongs\_to, ... \\
    \midrule
    DBLP & venue, author, term, paper & published\_in, mentions, writes, ... \\
    \midrule
    Reddit & user, post & interacts, rev\_interacts \\
    \midrule
    Texas, Cornell, Wisconsin & Content: Automated Theorem Proving Group... / Content: Steve Seitz's View Interpolations... & links\_to \\
    \midrule
    IMDB & movie, director, actor, keyword & directed\_by, stars, described\_by, ... \\
    \midrule
    Amazon-HeTGB & Name/title: I Ching Workbook. Salesrank: 160648. Reviews: rating: 5 votes: 1... & co-purchasing \\
    \midrule
    Amazon, Photo, Computers & product & co-purchasing, also\_bought, also\_viewed \\
    \midrule
    HIV, COX2, PROTEINS, ENZYMES & N/A & to \\
    \midrule
    FB15K-237 & Entity: /m/06n90 & /film/film/film\_art\_direction\_by, ... \\
    \midrule
    NELL, Elliptic & entity & relation; transaction \\
    \bottomrule
    \end{tabular}}
\end{table*}

\stitle{Heterogeneous graph homogenization.} We flatten heterogeneous graphs into a homogeneous format to ensure compatibility with standard backbones. This process involves specific strategies for feature alignment and structural merging.
Since different node types often possess varying feature dimensions or lack features entirely, we apply a rigorous alignment strategy:
\begin{itemize}[leftmargin=1.2em]
    \item \textbf{General initialization:} For specific node types (normally one) lacking intrinsic attributes, we initialize their features using unique one-hot vectors based on node identities.
    \item \textbf{Specifics for ogbn-mag:} In \textit{ogbn-mag}, we incorporate pre-trained 128-dimensional metapath2vec embeddings to create feature vectors. Specifically, ``paper'' node features are concatenated with these embeddings (yielding 256 dimensions), while auxiliary nodes (``author'', ``institution'', ``field of study'') exclusively use the metapath2vec embeddings instead of high-dimensional sparse one-hot vectors.
    \item \textbf{Zero-padding alignment:} We calculate the maximum feature dimension $D_{max}$ across all node types. Feature vectors with dimension $d < D_{max}$ are zero-padded to $\mathbb{R}^{D_{max}}$. Finally, all features are concatenated into a global matrix $\mathbf{X} \in \mathbb{R}^{N_{total} \times D_{max}}$.
\end{itemize}

Then we compute cumulative node offsets for each node type. Local edge indices are converted to global indices by adding the respective offsets. Furthermore, we consolidate multiple relations between the same pair of nodes into a single composite edge. For instance, in the \textit{Amazon}, if Product A is linked to Product B via both ``also\_bought'' and ``also\_viewed'' relations, these are merged into a single edge with the composite relation text ``also\_bought, also\_viewed'' and assigned a unique \texttt{edge\_type} ID.

\stitle{Temporal graph snapshot construction.} For dynamic graphs provided as temporal event streams, we construct a static snapshot encompassing historical information:
\begin{itemize}[leftmargin=1.2em]
    \item \textbf{Structural aggregation:} We aggregate all interactions over the timeline. To support standard message passing, we treat the graph as undirected by generating reverse edges for every interaction, explicitly distinguishing semantics in \texttt{relation\_texts} (e.g., distinguishing ``interacts'' from ``rev\_interacts'' in the \textit{Reddit}).
    \item \textbf{Latest-state labeling:} Since node labels may evolve, we adopt a ``latest-priority'' policy. For any node $u$, its label $y_u$ is determined by the label associated with its last interaction timestamp $t_{max}$.
\end{itemize}

\stitle{Graph construction policy and label management.}
We enforce a task-dependent and dataset-specific graph construction policy:
\begin{itemize}[leftmargin=1.2em]
    \item \textbf{Undirected:} For most datasets, graphs are symmetrized to facilitate bi-directional information flow during message passing.
    \item \textbf{Directed preservation:} For datasets with inherent asymmetric semantics, including \textit{Wiki}, \textit{FB15K-237}, \textit{WN18RR}, and \textit{DGraph}, we strictly preserve edge directionality throughout the pipeline, as the meaning of $u \to v$ is fundamentally different from $v \to u$. Specifically, an adaptation is applied to \textit{DGraph} during the node classification task: while the global graph structure is kept directed, the sampled subgraphs are symmetrized (converted to undirected) to enhance local neighborhood aggregation.
\end{itemize}

To enhance flexibility and prevent data leakage, target labels $\mathbf{y}$ are excluded from the unified data object interface. Instead, ground-truth labels and their indices are persisted in local storage. The alignment of $\mathbf{y}$ adapts to the specific task level:
\begin{itemize}[leftmargin=1.2em]
    \item \textbf{Node-level:} Labels align with global node indices. For heterogeneous graphs where valid labels exist only on specific target types (e.g., ``Paper'' in \textit{ogbn-mag}), indices corresponding to non-target nodes are filled with \texttt{-1}. This value acts as a mask token, ensuring the loss function is computed strictly on valid target nodes.
    \item \textbf{Edge-level:} Labels correspond to the indices of edges within the global edge list.
    \item \textbf{Graph-level:} Labels correspond to the graph-level \texttt{batch} ID.
\end{itemize}
This decoupling strategy allows the same unified graph structure to serve multiple diverse downstream tasks without modification.

\stitle{Dataset-Specific adaptations.}  We further introduce multi-granularity tasks to bridge the discrepancy between the rich raw data labels and the simplified standard protocols:
\begin{itemize}[leftmargin=1.2em]
    \item \textit{Elliptic:} Although the standard protocol is binary (licit vs. illicit), the raw data includes a third 'others' category. We thus construct both 2-class and 3-class tasks to leverage the full label set.
    \item \textit{DGraph:} While the original benchmark focuses on binary fraud detection, the raw data contains 4 distinct label indices, conceptually mapped to 3 semantic roles (Normal, Fraudster, Background Node). We extend this to include 2-class, 3-class, and 4-class tasks to evaluate multi-granularity generalization.
\end{itemize}

\subsection{Dataset Sampling}\label{app.data-split}
In this section, we elaborate on the data sampling protocols
introduced in Section~\ref{sec.setting}.

\stitle{Pre-training.} 
Since different GFMs employ distinct input processing mechanisms, we implement model-specific sampling strategies:

\begin{itemize}[leftmargin=1.2em]
    \item \textbf{Multi-domain subgraph batching:} 
    Models (\method{MDGPT}, \method{MDGPF}, \method{SAMGPT}, \method{G2P2}, \method{GFT}, \method{UniGraph2}) accept whole-graph inputs. To handle the scale discrepancy between large single-graph datasets (e.g., \textit{ogbn-arxiv}) and multi-graph datasets (e.g., \textit{HIV}), and to enable cross-domain attention within a single training step, we adopt a novel strategy: 
    1) \textit{Decomposition:} Large single graphs are decomposed into subgraphs by \texttt{NeighborLoader} from PyG, while individual graphs in multi-graph datasets are treated directly as atomic units without further splitting.
    2) \textit{Shuffling and merging:} We maintain a mixed pool of subgraphs from all domains. During each iteration, subgraphs from diverse sources are randomly sampled, shuffled, and concatenated into a single large ``super-graph'' (batch).
    3) \textit{Size control:} To manage memory constraints given the variance in subgraph sizes, we control the batch size by enforcing a maximum node budget $N_{max}$ per batch, rather than a fixed number of graphs.
    
    \item \textbf{Ego-Graph Construction:} 
    For models that require node-centric structural contexts, we construct specific subgraphs for each target node:
    \begin{itemize}[leftmargin=1.2em]
        \item \method{GCOPE} utilizes $k$-hop sampling to extract the immediate local neighborhood structure centered around each node. Notably, we exclude the original label-dependent padding for small subgraphs ($<10$ nodes) to prevent data leakage, relying strictly on actual structural neighbors.
        \item \method{GraphCLIP} employs Random Walk sampling to generate trace-based subgraphs, capturing both local structure and higher-order proximity.
    \end{itemize}
\end{itemize}

\stitle{Downstream.}
To handle the diverse scale of downstream datasets and varying model architectures, we implement an adaptive sampling strategy:

\begin{itemize}[leftmargin=1.5em]
    \item \textbf{Full-Batch Processing (Small-Scale):} 
    For datasets that fit entirely within GPU memory, we feed the full graph directly into the model without sampling. This ensures exact computation of representations using the complete adjacency matrix.

    \item \textbf{Mini-Batch Sampling (Large-Scale):} 
    To mitigate Out-Of-Memory issues on large-scale datasets, we employ subgraph sampling techniques. Specifically, we utilize the \texttt{NeighborLoader} (for node tasks) and \texttt{LinkNeighborLoader} (for edge tasks) from PyTorch Geometric. This strategy is applied to: \textit{ogbn-mag}, \textit{Products}, \textit{ogbn-proteins}, \textit{T-Finance}, \textit{DGraph}, and \textit{Wiki}. By sampling local neighborhoods dynamically, we enable scalable training on these massive graphs.

\item \textbf{Model-Specific Ego-Graph Batching:}
    Aligning with their pre-training paradigms, \method{GCOPE} and \method{GraphCLIP} operate on independent ego-graphs rather than a global graph view. For these models, we construct batches comprised of fixed numbers of pre-extracted ego-subgraphs.

    \item \textbf{Graph-Level Batching:} 
    For graph classification tasks, we employ standard graph mini-batching. In each iteration, a fixed number of disjoint graphs are sampled and collated into a diagonal block matrix for parallel processing.
\end{itemize}

\stitle{Discussion: Sampling Trade-offs.}
We compare node-centric ego-graphs against our dynamic neighborhood sampling. While node-centric methods offer deterministic memory control, they suffer from preprocessing bottlenecks and structural isolation. Our dynamic approach overcomes these by enabling efficient on-the-fly processing while preserving better global structural connectivity.

\subsection{Implementation Details}\label{app.hyperparams}
In this section, we elaborate on the full implementation details and hyperparameter
configurations introduced in Section~\ref{sec.setting}.

\stitle{General settings}\label{app.general-setting}
\noindent\textbf{Optimizer.} For all experiments, we use the Adam optimizer~\cite{kingma2015adam}, except for \method{GraphCLIP}, which uses AdamW~\cite{loshchilovdecoupled}.

\noindent\textbf{Environment.} The environment in which we run experiments is:
\begin{itemize}[label=-]
    \item Linux version: 5.15.0-122-generic
    \item Operating system: Ubuntu 22.04.5 LTS (Jammy Jellyfish)
    \item CPU information: AMD EPYC 7763 (64-cores)
    \item Memory: 512 GB RAM
    \item GPU information: 4$\times$ NVIDIA L40 (48 GB). \method{GraphCLIP} pre-training is distributed across all 4 GPUs, while other experiments are conducted on a single GPU.
    \item CUDA Version: 12.4
    \item Python Version: 3.10
\end{itemize}

\stitle{General Hyperparameters.}
Table~\ref{tab:general_hyperparams} details the specific backbone architectures and hyperparameter configurations for each GFM across both pre-training and downstream phases.

\begin{table*}[tbp]
    \centering
    \footnotesize
    \caption{General hyperparameter configurations.}
    \addtolength{\tabcolsep}{1.0mm}
    \label{tab:general_hyperparams}
    \resizebox{1\linewidth}{!}{
    \begin{tabular}{l|ccccc|ccc|ccc}
    \toprule
    \multirow{2}{*}{\textbf{Method}} & \multicolumn{5}{c|}{\textbf{GNN Architecture}} & \multicolumn{3}{c|}{\textbf{Pretraining}} & \multicolumn{3}{c}{\textbf{Downstream Adaptation}} \\
    \cmidrule(lr){2-6} \cmidrule(lr){7-9} \cmidrule(lr){10-12}
    & \textbf{Backbone} & \textbf{Unified Input} & \textbf{Hidden Dim} & \textbf{Output Dim} & \textbf{Layers} & \textbf{LR} & \textbf{Epochs} & \textbf{Patience} & \textbf{LR} & \textbf{Epochs} & \textbf{Patience} \\
    \midrule\midrule
    \method{GCOPE}     & \method{FAGCN}    & 100 & 128  & 128  & 2  & $1\text{e-}4$ & 100 & 100 & $1\text{e-}4$ & 100 & 100 \\
    \method{MDGPT}     & \method{GCN}      & 50  & 256  & 256  & 3  & $1\text{e-}3$ & 200 & 20  & $1\text{e-}2$ & 400 & 10  \\
    \method{MDGFM}     & \method{GCN}      & 50  & 256  & 256  & 3  & $1\text{e-}3$ & 200 & 20  & $1\text{e-}2$ & 400 & 20  \\
    \method{SAMGPT}    & \method{GCN}      & 50  & 256  & 256  & 3  & $1\text{e-}3$ & 200 & 20  & $1\text{e-}2$ & 400 & 10  \\
    \method{G2P2}      & \method{GCN}      & 128 & 128  & 128  & 2  & $2\text{e-}5$ & 10  & 3   & $1\text{e-}2$ & 50  & 50  \\
    \method{GraphCLIP} & \method{GraphGPS}~\cite{rampavsek2022recipe} & 384 & 1024 & 384 & 12 & $1\text{e-}5$ & 30  & 30  & $1\text{e-}4$ & 100 & 100 \\
    \method{GFT}       & \method{SAGE}~\cite{hamilton2017inductive} & 50 & 128  & 128  & 3  & $1\text{e-}4$ & 50 & 50  & $1\text{e-}3$ & 200 & 10  \\
    \method{UniGraph2} & GAT & 50 & 768  & 768  & 3  & $1\text{e-}4$ & 50 & 50  & - & - & -  \\
    \bottomrule
    \end{tabular}}
    \vspace{1mm}
    \footnotesize{ Unified Input denotes the dimension to which raw node features are projected before entering the backbone.
    `-' indicates that the specific parameter is not applicable.}
\end{table*}

\stitle{Model-Specific Hyperparameters.}
Below, we detail the specific architectures and hyperparameter settings for each Graph Foundation Model (GFM). Unless otherwise stated, the output dimension matches the hidden dimension.

\begin{itemize}[leftmargin=1.5em]
    \item \method{GCOPE} introduces one coordinator for each source dataset and assigns 0.2 as the default reconstruction weight.

    \item \method{MDGPT} adopts graph contrastive learning as the pretraining objective and employs a random edge augmentation strategy with a perturbation rate of 0.1.

    \item \method{MDGFM} adopts link prediction as the pretraining objective and reconstructs the graph structure via k-nearest neighbor (kNN) sparsification with $k=15$ (consistent across both pretraining and downstream phases), applying a 0.5 dropout rate to the refined edge weights.

    \item \method{SAMGPT} follows the same configurations as \method{MDGPT}.

    \item \method{G2P2} performs joint training using a 63M-parameter transformer (12 layers, 512 hidden dim, 8 heads) as text encoder based on lower-cased BPE with a vocabulary of 49,152. It sets the number of sampled neighbors $\eta$ to 3 and the batch size $\beta$ to 128, utilizing $M=4$ learnable prompt tokens and a unified regularization weight $\lambda=10$.

    \item \method{GraphCLIP} utilizes a frozen fine-tuned MiniLM text encoder and projects the 1024-dimensional graph embeddings into a 384-dimensional unified subspace, optimizing only the graph backbone and projector. Regarding the textual input generation, while the original implementation deploys the open-source \method{Qwen2-72B-Instruct} model locally to summarize ego-subgraphs, we utilize the \method{ChatGPT-5-mini} API (OpenAI) with minimal reasoning enabled for efficient text generation.

    \item \method{GFT} configures the vector-quantized codebook with a commitment weight of 0.25 and applies orthonormal regularization with a weight of 1.0 over a maximum of 32 codes.
    
    \item \method{UniGraph2} employs a multimodal masked autoencoder with a 4-head GAT backbone and an 8-expert MoE module (top-2 selected). It minimizes a composite objective of feature reconstruction ($\gamma=2.0$) and structural SPD prediction ($\lambda=0.5$) under a 0.1 masking rate, performing inference without fine-tuning.
\end{itemize}

\subsection{Adaptation to Unseen Datasets}\label{app.exp1}
We report the Macro-F1 of 1-shot node classification on unseen datasets in Table~\ref{table.exp1-1shotnc-macrof}. The 5-shot node classification results are shown in Tables~\ref{table.exp1-5shotnc-acc} and~\ref{table.exp1-5shotnc-macrof}, and the 5-shot edge and graph classification results are summarized in Table~\ref{table.exp1-5shotecgc}. These results exhibit the same qualitative patterns as those discussed in Sec.~\ref{sec.exp1}.

We additionally summarize downstream adaptation time (total and per-epoch) for 1-shot node classification in Tables~\ref{table.exp1-1shotnc-t} and~\ref{table.exp1-1shotnc-avgt}, for 5-shot node classification in Tables~\ref{table.exp1-5shotnc-t} and~\ref{table.exp1-5shotnc-avgt}, for 1-shot edge and graph classification in Table~\ref{table.exp1-1shotecgc-t}, and for 5-shot edge and graph classification in Table~\ref{table.exp1-5shotecgc-t}.

\begin{table*}[tbp] % [!t]
    \centering
    \footnotesize
    \caption{Macro-F1 of 1-shot node classification on unseen domains.}
    \addtolength{\tabcolsep}{1.0mm}
    \label{table.exp1-1shotnc-macrof}
    \resizebox{1\linewidth}{!}{
    \begin{tabular}{l|ccccccccc}
    \toprule
    {Methods} & {Pubmed} & {ogbn-mag} & {Wikipedia} & {Actor}  & {Chameleon} & {Products} & {T-Finance} & {DGraph} & {ogbn-proteins}\\
    \midrule\midrule
    \method{GCN} &43.39{\tiny$\pm$11.32} &\textbf{4.49}{\tiny$\pm$0.27} &28.72{\tiny$\pm$19.65} &20.24{\tiny$\pm$7.38} &21.59{\tiny$\pm$4.32} &\underline{12.17}{\tiny$\pm$1.70} &26.94{\tiny$\pm$21.92} &23.42{\tiny$\pm$7.91} &37.96{\tiny$\pm$9.95}\\
    \method{GAT} &39.05{\tiny$\pm$10.94} &3.39{\tiny$\pm$0.25} &31.13{\tiny$\pm$12.89} &18.38{\tiny$\pm$7.29} &21.96{\tiny$\pm$4.18} &12.13{\tiny$\pm$1.85} &33.67{\tiny$\pm$20.34} &23.95{\tiny$\pm$8.20}&31.99{\tiny$\pm$6.60}\\\midrule
    \method{GCOPE}&39.94{\tiny$\pm$8.76} &0.06{\tiny$\pm$0.01} &29.82{\tiny$\pm$13.47} &19.02{\tiny$\pm$3.97} &19.51{\tiny$\pm$3.61} &OOT &32.62{\tiny$\pm$11.35} &OOT &32.43{\tiny$\pm$7.98}\\
    \method{MDGPT}&38.80{\tiny$\pm$7.39} &\underline{4.41}{\tiny$\pm$0.32} &33.70{\tiny$\pm$12.85} &\textbf{24.58}{\tiny$\pm$5.67} &\textbf{25.98}{\tiny$\pm$5.12} &9.99{\tiny$\pm$1.54} &42.96{\tiny$\pm$14.41} &\underline{27.06}{\tiny$\pm$6.17} &\textbf{47.88}{\tiny$\pm$9.99}\\
    \method{MDGFM} &35.37{\tiny$\pm$6.65} &2.06{\tiny$\pm$0.75} &29.67{\tiny$\pm$14.59} &14.80{\tiny$\pm$3.49} &23.04{\tiny$\pm$3.75} &5.09{\tiny$\pm$1.07} &41.65{\tiny$\pm$11.32} &21.77{\tiny$\pm$12.54} &44.52{\tiny$\pm$6.79}\\
    \method{SAMGPT} &\underline{47.82}{\tiny$\pm$8.75} &3.77{\tiny$\pm$0.23} &\underline{35.88}{\tiny$\pm$12.86} &21.59{\tiny$\pm$5.80} &\underline{25.05}{\tiny$\pm$5.33} &\textbf{12.53}{\tiny$\pm$1.58} &43.41{\tiny$\pm$10.06} &\textbf{27.45}{\tiny$\pm$6.83} &\underline{47.83}{\tiny$\pm$10.66}\\
    \method{G2P2} &46.76{\tiny$\pm$6.38} &- &19.68{\tiny$\pm$3.72} &\underline{24.56}{\tiny$\pm$5.28} &- &- &\textbf{46.46}{\tiny$\pm$12.16} &23.25{\tiny$\pm$4.13} &-\\
    \method{GraphCLIP} &34.76{\tiny$\pm$4.41} &- &17.24{\tiny$\pm$8.33} &18.47{\tiny$\pm$3.03} &- &- &32.55{\tiny$\pm$11.58} &21.74{\tiny$\pm$11.80} &-\\
    \method{GFT} &\textbf{53.34}{\tiny$\pm$10.53} &0.95{\tiny$\pm$0.09} &\textbf{37.42}{\tiny$\pm$8.63} &21.08{\tiny$\pm$3.88} &24.61{\tiny$\pm$4.10} &6.05{\tiny$\pm$1.80} &37.05{\tiny$\pm$15.75} &26.56{\tiny$\pm$6.94} &46.95{\tiny$\pm$6.75}\\
    \method{UniGraph2} &43.43{\tiny$\pm$10.13} &0.45{\tiny$\pm$0.11} &31.69{\tiny$\pm$17.85} &20.67{\tiny$\pm$5.26} &22.88{\tiny$\pm$5.33} &11.79{\tiny$\pm$1.57} &\underline{43.55}{\tiny$\pm$15.34} &26.37{\tiny$\pm$5.94} &45.06{\tiny$\pm$4.65}\\\bottomrule
    \end{tabular}}
\end{table*}

\begin{table*}[tbp] % [!t]
    \centering
    \footnotesize
    \caption{Accuracy of 5-shot node classification on unseen domains.}
    \addtolength{\tabcolsep}{1.0mm}
    \label{table.exp1-5shotnc-acc}
    \resizebox{1\linewidth}{!}{
    \begin{tabular}{l|ccccccccc}
    \toprule
    {Methods} & {Pubmed} & {ogbn-mag} & {Wikipedia} & {Actor}  & {Chameleon} & {Products} & {T-Finance} & {DGraph} & {ogbn-proteins}\\
    \midrule\midrule
    \method{GCN} &64.30{\tiny$\pm$6.69} &11.32{\tiny$\pm$0.55} &46.14{\tiny$\pm$28.17} &\underline{40.51}{\tiny$\pm$8.57} &35.71{\tiny$\pm$3.48} &\underline{34.95}{\tiny$\pm$8.44} &48.03{\tiny$\pm$39.90} &40.84{\tiny$\pm$9.77} &51.75{\tiny$\pm$12.54}\\
    \method{GAT} &\underline{65.71}{\tiny$\pm$6.16} &10.05{\tiny$\pm$0.66} &48.14{\tiny$\pm$19.91} &39.72{\tiny$\pm$9.00} &35.33{\tiny$\pm$3.28} &\textbf{43.66}{\tiny$\pm$5.16} &59.05{\tiny$\pm$33.99} &41.20{\tiny$\pm$9.92} &49.96{\tiny$\pm$13.96}\\\midrule
    \method{GCOPE} &59.13{\tiny$\pm$5.40} &0.25{\tiny$\pm$0.06} &51.77{\tiny$\pm$18.38} &29.71{\tiny$\pm$8.06} &25.41{\tiny$\pm$2.19} &OOT &35.86{\tiny$\pm$13.81} &OOT &42.44{\tiny$\pm$11.57}\\
    \method{MDGPT} &53.58{\tiny$\pm$6.27} &\textbf{11.93}{\tiny$\pm$0.64} &54.48{\tiny$\pm$18.49} &\textbf{41.33}{\tiny$\pm$5.26} &\underline{36.33}{\tiny$\pm$3.26} &25.61{\tiny$\pm$5.47} &58.90{\tiny$\pm$17.82} &38.17{\tiny$\pm$7.42} &\underline{54.00}{\tiny$\pm$9.61}\\
    \method{MDGFM} &51.71{\tiny$\pm$7.34} &OOM &49.81{\tiny$\pm$19.18} &23.56{\tiny$\pm$5.66} &31.70{\tiny$\pm$3.50} &8.97{\tiny$\pm$3.85} &54.14{\tiny$\pm$12.99} &39.44{\tiny$\pm$13.73} &50.98{\tiny$\pm$5.72}\\
    \method{SAMGPT} &62.21{\tiny$\pm$4.90} &\underline{11.41}{\tiny$\pm$0.81} &\textbf{58.08}{\tiny$\pm$14.24} &38.94{\tiny$\pm$5.48} &\textbf{37.86}{\tiny$\pm$3.59} &31.24{\tiny$\pm$6.93} &60.00{\tiny$\pm$10.20} &40.57{\tiny$\pm$10.19} &\textbf{54.51}{\tiny$\pm$11.51}\\
    \method{G2P2} &56.55{\tiny$\pm$5.41} &- &43.72{\tiny$\pm$8.80} &37.64{\tiny$\pm$5.23} &- &- &67.44{\tiny$\pm$11.81} &\textbf{43.52}{\tiny$\pm$8.90} &-\\
    \method{GraphCLIP} &45.58{\tiny$\pm$4.54} &- &43.65{\tiny$\pm$12.48} &27.49{\tiny$\pm$4.44} &- &- &45.42{\tiny$\pm$25.80} &36.43{\tiny$\pm$14.03} &-\\
    \method{GFT} &\textbf{65.73}{\tiny$\pm$4.76} &3.76{\tiny$\pm$0.64} &\underline{56.74}{\tiny$\pm$10.69} &35.97{\tiny$\pm$5.31} &33.21{\tiny$\pm$2.72} &20.84{\tiny$\pm$3.54} &\underline{67.83}{\tiny$\pm$13.97} &\underline{42.22}{\tiny$\pm$7.73} &52.96{\tiny$\pm$5.27}\\
    \method{UniGraph2} &57.99{\tiny$\pm$7.02} &3.61{\tiny$\pm$1.94} &51.36{\tiny$\pm$19.89} &32.91{\tiny$\pm$6.61} &32.93{\tiny$\pm$3.33} &26.63{\tiny$\pm$13.32} &\textbf{69.07}{\tiny$\pm$13.72} &37.61{\tiny$\pm$7.95} &50.45{\tiny$\pm$3.19}\\\bottomrule
    \end{tabular}}
\end{table*}

\begin{table*}[tbp] % [!t]
    \centering
    \footnotesize
    \caption{Macro-F1 of 5-shot node classification on unseen domains.}
    \addtolength{\tabcolsep}{1.0mm}
    \label{table.exp1-5shotnc-macrof}
    \resizebox{1\linewidth}{!}{
    \begin{tabular}{l|ccccccccc}
    \toprule
    {Methods} & {Pubmed} & {ogbn-mag} & {Wikipedia} & {Actor}  & {Chameleon} & {Products} & {T-Finance} & {DGraph} & {ogbn-proteins}\\
    \midrule\midrule
    \method{GCN} &63.59{\tiny$\pm$7.28} &\textbf{10.51}{\tiny$\pm$0.29} &31.07{\tiny$\pm$13.99} &\underline{32.90}{\tiny$\pm$5.64} &34.91{\tiny$\pm$3.53} &\underline{17.84}{\tiny$\pm$3.39} &31.14{\tiny$\pm$23.50} &29.72{\tiny$\pm$4.80} &41.13{\tiny$\pm$11.52}\\
    \method{GAT} &\underline{65.14}{\tiny$\pm$6.32} &9.08{\tiny$\pm$0.29} &33.42{\tiny$\pm$9.54} &30.96{\tiny$\pm$6.43} &34.68{\tiny$\pm$3.40} &\textbf{24.08}{\tiny$\pm$1.39} &40.97{\tiny$\pm$22.11} &\underline{30.11}{\tiny$\pm$4.85} &33.54{\tiny$\pm$6.99}\\\midrule
    \method{GCOPE} &59.15{\tiny$\pm$5.35} &0.06{\tiny$\pm$0.01} &35.54{\tiny$\pm$8.85} &24.29{\tiny$\pm$4.88} &23.83{\tiny$\pm$1.96} &OOT &28.58{\tiny$\pm$8.11} &OOT &30.93{\tiny$\pm$7.43}\\
    \method{MDGPT} &52.84{\tiny$\pm$6.44} &\underline{9.39}{\tiny$\pm$0.29} &36.50{\tiny$\pm$8.66} &\textbf{33.60}{\tiny$\pm$3.56} &\underline{35.19}{\tiny$\pm$3.19} &13.15{\tiny$\pm$1.92} &43.33{\tiny$\pm$10.49} &28.41{\tiny$\pm$3.62} &\textbf{52.04}{\tiny$\pm$9.65}\\
    \method{MDGFM} &50.99{\tiny$\pm$7.71} &OOM &34.26{\tiny$\pm$9.04} &18.46{\tiny$\pm$3.87} &30.51{\tiny$\pm$4.16} &4.77{\tiny$\pm$1.51} &41.47{\tiny$\pm$7.71} &28.49{\tiny$\pm$6.78} &47.66{\tiny$\pm$4.13}\\
    \method{SAMGPT} &62.16{\tiny$\pm$4.87} &8.91{\tiny$\pm$0.64} &\underline{38.56}{\tiny$\pm$5.89} &31.67{\tiny$\pm$3.65} &\textbf{37.22}{\tiny$\pm$3.77} &15.60{\tiny$\pm$1.94} &44.89{\tiny$\pm$5.95} &29.47{\tiny$\pm$4.99} &\underline{51.52}{\tiny$\pm$10.31}\\
    \method{G2P2} &55.01{\tiny$\pm$5.34} &- &21.65{\tiny$\pm$2.99} &31.80{\tiny$\pm$3.64} &- &- &\underline{49.52}{\tiny$\pm$7.43} &25.35{\tiny$\pm$3.06} &-\\
    \method{GraphCLIP} &45.09{\tiny$\pm$4.92} &- &21.51{\tiny$\pm$4.32} &21.32{\tiny$\pm$2.50} &- &- &31.97{\tiny$\pm$14.00} &21.76{\tiny$\pm$5.42} &-\\
    \method{GFT} &\textbf{65.80}{\tiny$\pm$4.65} &2.25{\tiny$\pm$0.23} &\textbf{38.85}{\tiny$\pm$5.23} &29.69{\tiny$\pm$3.53} &31.70{\tiny$\pm$2.94} &9.93{\tiny$\pm$1.28} &\textbf{50.68}{\tiny$\pm$9.18} &\textbf{31.29}{\tiny$\pm$4.12} &51.15{\tiny$\pm$4.45}\\
    \method{UniGraph2} &56.17{\tiny$\pm$6.87} &0.57{\tiny$\pm$1.60} &33.08{\tiny$\pm$10.86} &27.12{\tiny$\pm$4.17} &29.96{\tiny$\pm$3.68} &12.77{\tiny$\pm$1.51} &40.41{\tiny$\pm$5.57} &26.69{\tiny$\pm$4.16} &47.98{\tiny$\pm$2.17}\\\bottomrule
    \end{tabular}}
\end{table*}

\begin{table*}[tbp] % [!t]
    \centering
    \footnotesize
    \caption{Evaluation of 5-shot edge classification and graph classification on unseen domains.}
    \addtolength{\tabcolsep}{1.0mm}
    \label{table.exp1-5shotecgc}
    \resizebox{1\linewidth}{!}{
    \begin{tabular}{l|cccccc|cccc}
    \toprule
    \multirow{3}{*}{Methods} & \multicolumn{6}{c|}{Edge classification} & \multicolumn{4}{c}{Graph classification}\\
     &\multicolumn{2}{c}{DGraph} &\multicolumn{2}{c}{Wiki} &\multicolumn{2}{c|}{WN18RR} &\multicolumn{2}{c}{PCBA} &\multicolumn{2}{c}{BZR} \\
    &Acc &MacroF &Acc &MacroF &Acc &MacroF &Acc &MacroF &Acc &MacroF\\
    \midrule\midrule
    \method{GCN} &10.31{\tiny$\pm$1.77} &7.68{\tiny$\pm$0.78} &\textbf{21.49}{\tiny$\pm$3.38} &\textbf{7.40}{\tiny$\pm$0.19} &\underline{20.18}{\tiny$\pm$13.60} &\textbf{20.14}{\tiny$\pm$4.81} &55.24{\tiny$\pm$30.77} &32.74{\tiny$\pm$15.69} &60.22{\tiny$\pm$13.99} &49.94{\tiny$\pm$9.70}\\
    \method{GAT} &\underline{10.47}{\tiny$\pm$1.75} &\underline{7.77}{\tiny$\pm$0.80} &10.96{\tiny$\pm$1.28} &4.93{\tiny$\pm$0.18} &12.54{\tiny$\pm$10.60} &\underline{16.66}{\tiny$\pm$3.19} &\underline{57.48}{\tiny$\pm$24.48} &34.82{\tiny$\pm$12.64} &56.33{\tiny$\pm$13.48} &49.00{\tiny$\pm$9.42}\\\midrule
    \method{GCOPE} &OOT &OOT &OOT &OOT &18.25{\tiny$\pm$4.67} &14.35{\tiny$\pm$2.53} &47.70{\tiny$\pm$32.64} &28.97{\tiny$\pm$16.13} &41.41{\tiny$\pm$16.23} &38.46{\tiny$\pm$13.22}\\
    \method{MDGPT} &\textbf{10.85}{\tiny$\pm$1.90} &\textbf{8.02}{\tiny$\pm$0.97} &14.00{\tiny$\pm$2.58} &5.34{\tiny$\pm$0.23} &18.43{\tiny$\pm$3.62} &16.03{\tiny$\pm$2.66} &\textbf{58.31}{\tiny$\pm$16.94} &\textbf{36.25}{\tiny$\pm$7.37} &53.97{\tiny$\pm$11.76} &50.35{\tiny$\pm$9.68}\\
    \method{MDGFM} &9.92{\tiny$\pm$2.17} &6.97{\tiny$\pm$0.84} &11.32{\tiny$\pm$2.96} &3.47{\tiny$\pm$0.40} &17.32{\tiny$\pm$2.86} &13.79{\tiny$\pm$1.60} &55.27{\tiny$\pm$27.36} &33.52{\tiny$\pm$13.25} &37.26{\tiny$\pm$19.41} &30.48{\tiny$\pm$11.90}\\
    \method{SAMGPT} &10.43{\tiny$\pm$2.07} &7.62{\tiny$\pm$0.84} &\underline{15.24}{\tiny$\pm$2.48} &\underline{5.92}{\tiny$\pm$0.20} &18.14{\tiny$\pm$3.47} &14.85{\tiny$\pm$1.75} &56.40{\tiny$\pm$14.87} &\underline{35.67}{\tiny$\pm$6.41} &\textbf{65.39}{\tiny$\pm$10.28} &\textbf{57.64}{\tiny$\pm$7.05}\\
    \method{G2P2} &- &- &- &- &16.90{\tiny$\pm$2.89} &13.50{\tiny$\pm$1.50} &- &- &60.89{\tiny$\pm$9.65} &\underline{54.87}{\tiny$\pm$6.82}\\
    \method{GraphCLIP} &- &- &- &- &\textbf{20.20}{\tiny$\pm$4.30} &15.19{\tiny$\pm$1.78} &- &- &53.72{\tiny$\pm$17.44} &44.88{\tiny$\pm$10.52}\\
    \method{GFT} &10.21{\tiny$\pm$2.00} &7.17{\tiny$\pm$0.88} &OOT &OOT &16.29{\tiny$\pm$5.48} &12.78{\tiny$\pm$1.70} &52.13{\tiny$\pm$15.91} &33.70{\tiny$\pm$9.22} &\underline{62.70}{\tiny$\pm$14.25} &52.18{\tiny$\pm$10.83}\\
    \method{UniGraph2} &10.22{\tiny$\pm$6.11} &7.64{\tiny$\pm$0.89} &11.13{\tiny$\pm$0.24} &5.02{\tiny$\pm$0.20} &15.35{\tiny$\pm$8.76} &11.93{\tiny$\pm$1.26} &51.66{\tiny$\pm$13.08} &33.38{\tiny$\pm$6.00} &62.46{\tiny$\pm$10.80} &49.78{\tiny$\pm$6.20}\\\bottomrule
    \end{tabular}}
\end{table*}

%%%%%%%%%%%%%%%%%%%%%%%%%%%%%%Time
\begin{table*}[tbp] % [!t]
    \centering
    \footnotesize
    \caption{Time of downstream tuning for 1-shot node classification on unseen domains.}
    \addtolength{\tabcolsep}{1.0mm}
    \label{table.exp1-1shotnc-t}
    \resizebox{1\linewidth}{!}{
    \begin{tabular}{l|ccccccccc}
    \toprule
    {Methods} & {Pubmed} & {ogbn-mag} & {Wikipedia} & {Actor}  & {Chameleon} & {Products} & {T-Finance} & {DGraph} & {ogbn-proteins}\\
    \midrule\midrule
    \method{GCOPE} &4982.84 &145726.60 &3609.70 &2399.45 &2040.97 &OOT &10597.74 &OOT &30164.69\\
    \method{MDGPT} &996.96 &21295.45 &578.18 &559.66 &621.26 &41059.80 &1677.46 &6052.41 &2246.63\\
    \method{MDGFM} &4276.28 &68365.23 &520.57 &571.15 &977.57 &206484.25 &5429.48 &4385.13 &9203.33\\
    \method{SAMGPT} &3100.42 &37458.71 &1302.22 &1395.42 &1413.17 &101031.49 &4254.17 &7890.00 &2561.89\\
    \method{G2P2} &268.5 &- &284.01 &284.77 &- &- &3150.19 &1928.43 &-\\
    \method{GraphCLIP} &1143.31 &- &549.77 &603.42 &- &- &1578.20 &10320.75 &-\\
    \method{GFT} &1264.49 &126155.83 &5586.78 &394.60 &363.13 &157924.23 &41145.66 &23168.21 &63876.06\\
    \method{UniGraph2} &1778.73 &107988.18 &683.08 &214.92 &593.02 &138796.78 &22396.36 &19710.11 &104354.87\\\bottomrule
    \end{tabular}}
\end{table*}

\begin{table*}[tbp] % [!t]
    \centering
    \footnotesize
    \caption{Time/epoch of downstream tuning for 1-shot node classification on unseen domains.}
    \addtolength{\tabcolsep}{1.0mm}
    \label{table.exp1-1shotnc-avgt}
    \resizebox{1\linewidth}{!}{
    \begin{tabular}{l|ccccccccc}
    \toprule
    {Methods} & {Pubmed} & {ogbn-mag} & {Wikipedia} & {Actor}  & {Chameleon} & {Products} & {T-Finance} & {DGraph} & {ogbn-proteins}\\
    \midrule\midrule
    \method{GCOPE} &0.9966 &29.1453 &0.7219 &0.4799 &0.4082 &OOT &2.1195 &OOT &6.0329\\
    \method{MDGPT} &0.0499 &2.3057 &0.0289 &0.0280 &0.0311 &5.4761 &1.1796 &0.3600 &2.2443\\
    \method{MDGFM} &0.2952 &4.5010 &0.1541 &0.0990 &0.0744 &12.7894 &1.3986 &0.6626 &2.5256\\
    \method{SAMGPT} &0.1554 &2.7046 &0.0721 &0.0707 &0.0711 &6.5106 &1.2225 &0.4005 &2.4340\\
    \method{G2P2} &0.1074 &- &0.1136 &0.1139 &- &- &1.2601 &0.7714 &-\\
    \method{GraphCLIP} &0.2287 &- &0.1100 &0.1207 &- &- &0.3156 &2.0642 &-\\
    \method{GFT} &0.0903 &12.6156 &0.4422 &0.0244 &0.0235 &16.6922 &8.5135 &2.8209 &14.2089\\
    \method{UniGraph2} &- &- &- &- &- &- &- &- &-\\\bottomrule
    \end{tabular}}
\end{table*}

\begin{table*}[tbp] % [!t]
    \centering
    \footnotesize
    \caption{Time of downstream tuning for 5-shot node classification on unseen domains.}
    \addtolength{\tabcolsep}{1.0mm}
    \label{table.exp1-5shotnc-t}
    \resizebox{1\linewidth}{!}{
    \begin{tabular}{l|ccccccccc}
    \toprule
    {Methods} & {Pubmed} & {ogbn-mag} & {Wikipedia} & {Actor}  & {Chameleon} & {Products} & {T-Finance} & {DGraph} & {ogbn-proteins}\\
    \midrule\midrule
    \method{GCOPE} &4967.40 &145716.29 &3583.13 &2419.86 &2036.05 &OOT &10599.70 &OOT &30412.11\\
    \method{MDGPT} &91.60 &30592.08 &573.91 &559.37 &620.14 &29485.19 &1983.93 &6030.87 &2309.75\\
    \method{MDGFM} &4103.18 &OOM &549.41 &582.99 &998.64 &195360.13 &4520.15 &5207.33 &11524.04\\
    \method{SAMGPT} &3107.38 &97086.74 &1462.31 &1375.35 &1276.83 &105312.52 &4113.18 &7786.37 &2616.61\\
    \method{G2P2} &279.28 &- &284.19 &278.05 &- &- &3365.97 &1920.82 &-\\
    \method{GraphCLIP} &1234.62 &- &597.31 &675.31 &- &- &1624.89 &10517.09 &-\\
    \method{GFT} &68.65 &153450.58 &8856.30 &1090.68 &501.34 &158267.54 &33193.50 &18052.72 &72111.36\\
    \method{UniGraph2} &1673.26 &108452.68 &746.61 &282.63 &642.26 &141652.54 &21014.23 &19560.55 &108606.81\\\bottomrule
    \end{tabular}}
\end{table*}

\begin{table*}[tbp] % [!t]
    \centering
    \footnotesize
    \caption{Time/epoch of downstream tuning for 5-shot node classification on unseen domains.}
    \addtolength{\tabcolsep}{1.0mm}
    \label{table.exp1-5shotnc-avgt}
    \resizebox{1\linewidth}{!}{
    \begin{tabular}{l|ccccccccc}
    \toprule
    {Methods} & {Pubmed} & {ogbn-mag} & {Wikipedia} & {Actor}  & {Chameleon} & {Products} & {T-Finance} & {DGraph} & {ogbn-proteins}\\
    \midrule\midrule
    \method{GCOPE} &0.9935 &29.1433 &0.7166 &0.4840 &0.4072 &OOT &2.1199 &OOT &6.0824\\
    \method{MDGPT} &0.0503 &4.0784 &0.0287 &0.0280 &0.0311 &6.6558 &1.2596 &0.3580 &2.3765\\
    \method{MDGFM} &0.2977 &OOM &0.1592 &0.0988 &0.0819 &16.6476 &1.4073 &0.5761 &3.0493\\
    \method{SAMGPT} &0.1554 &5.5339 &0.0731 &0.0688 &0.0638 &9.0304 &1.3162 &0.3964 &2.6281\\
    \method{G2P2} &0.1117 &- &0.1137 &0.1112 &- &- &1.3464 &0.7683 &-\\
    \method{GraphCLIP} &0.2469 &- &0.1195 &0.1351 &- &- &0.3250 &2.1034 &-\\
    \method{GFT} &0.0900 &15.3451 &0.5737 &0.0557 &0.0269 &16.7534 &6.8271 &3.6105 &14.3866\\
    \method{UniGraph2} &- &- &- &- &- &- &- &- &-\\\bottomrule
    \end{tabular}}
\end{table*}

\begin{table*}[tbp] % [!t]
    \centering
    \footnotesize
    \caption{Time and time/epoch of downstream tuning for 1-shot edge classification and graph classification on unseen domains.}
    \addtolength{\tabcolsep}{1.0mm}
    \label{table.exp1-1shotecgc-t}
    \resizebox{1\linewidth}{!}{
    \begin{tabular}{l|cccccc|cccc}
    \toprule
    \multirow{3}{*}{Methods} & \multicolumn{6}{c|}{Edge classification} & \multicolumn{4}{c}{Graph classification}\\
     &\multicolumn{2}{c}{DGraph} &\multicolumn{2}{c}{Wiki} &\multicolumn{2}{c|}{WN18RR} &\multicolumn{2}{c}{PCBA} &\multicolumn{2}{c}{BZR} \\
    &Time &Avg. T. &Time &Avg. T. &Time &Avg. T. &Time &Avg. T. &Time &Avg. T. \\
    \midrule\midrule
    \method{GCOPE} &OOT &OOT &OOT &OOT &5366.42 &1.0733 &7849.48 &1.5689 &1999.33 &0.3999\\
    \method{MDGPT} &6691.48 &0.6109 &27480.67 &2.1883 &1482.20 &0.0741 &1171.83 &0.0777 &373.24 &0.0254\\
    \method{MDGFM} &16287.70 &1.1188 &79650.24 &4.0031 &12818.37 &0.7522 &9000.25 &1.8664 &194.41 &0.0515\\
    \method{SAMGPT} &14194.53 &0.8092 &52318.30 &2.6658 &5529.34 &0.2765 &2046.01 &0.1056 &413.24 &0.0242\\
    \method{G2P2} &- &- &- &- &311.16 &0.1245 &- &- &260.62 &0.1042\\
    \method{GraphCLIP} &- &- &- &- &2435.14 &0.4870 &- &- &1500.62 &0.3001\\
    \method{GFT} &14844.22 &0.9222 &OOT &OOT &342.12 &0.0779 &57977.69 &12.4772 &237.40 &0.1020\\
    \method{UniGraph2} &5701.38 &- &73366.01 &- &426.74 &- &13861.10 &- &227.95 &-\\\bottomrule
    \end{tabular}}
\end{table*}

\begin{table*}[tbp] % [!t]
    \centering
    \footnotesize
    \caption{Time and time/epoch of downstream tuning for 5-shot edge classification and graph classification on unseen domains.}
    \addtolength{\tabcolsep}{1.0mm}
    \label{table.exp1-5shotecgc-t}
    \resizebox{1\linewidth}{!}{
    \begin{tabular}{l|cccccc|cccc}
    \toprule
    \multirow{3}{*}{Methods} & \multicolumn{6}{c|}{Edge classification} & \multicolumn{4}{c}{Graph classification}\\
     &\multicolumn{2}{c}{DGraph} &\multicolumn{2}{c}{Wiki} &\multicolumn{2}{c|}{WN18RR} &\multicolumn{2}{c}{PCBA} &\multicolumn{2}{c}{BZR} \\
    &Time &Avg. T. &Time &Avg. T. &Time &Avg. T. &Time &Avg. T. &Time &Avg. T.\\
    \midrule\midrule
    \method{GCOPE} &OOT &OOT &OOT &OOT &5377.28 &1.0755 &7869.82 &1.5740 &1990.91 &0.3982\\
    \method{MDGPT} &3239.32 &0.5356 &33128.83 &3.1503 &1467.97 &0.0744 &1159.48 &0.0925 &421.82 &0.0232\\
    \method{MDGFM} &14083.09 &1.1941 &78096.08 &3.9079 &12869.09 &0.7497 &9296.78 &1.0705 &233.65 &0.0366\\
    \method{SAMGPT} &17314.93 &0.9869 &53692.22 &2.8210 &5531.02 &0.2766 &1943.27 &0.1037 &297.63 &0.0152\\
    \method{G2P2} &- &- &- &- &310.80 &0.1243 &- &- &261.65 &0.1047\\
    \method{GraphCLIP} &- &- &- &- &2421.57 &0.4843 &- &- &1522.43 &0.3045\\
    \method{GFT} &14859.82 &0.9757 &OOT &OOT &604.05 &0.1468 &60763.81 &12.9928 &185.39 &0.0813\\
    \method{UniGraph2} &5288.73 &- &74826.76 &- &423.24 &- &14087.09 &- &165.73 &-\\\bottomrule
    \end{tabular}}
\end{table*}

\subsection{Adaptation to seen Datasets}\label{app.exp2}
We report the 5-shot node classification results in Table~\ref{table.exp2-5shotnc}, and summarize the 5-shot edge and graph classification results in Table~\ref{table.exp2-5shotecgc}. These results follow the same qualitative trends as those discussed in Sec.~\ref{sec.exp2}.

We further report downstream adaptation time for 1-shot and 5-shot node classification in Tables~\ref{table.exp2-1shotnc-t} and~\ref{table.exp2-5shotnc-t}, and for 1-shot and 5-shot edge and graph classification in Tables~\ref{table.exp2-1shotecgc-t} and~\ref{table.exp2-5shotecgc-t}, respectively.

\begin{table*}[tbp] % [!t]
    \centering
    \footnotesize
    \caption{Evaluation of 5-shot node classification on seen domains.}
    %\addtolength{\tabcolsep}{1.0mm}
    \label{table.exp2-5shotnc}
    \resizebox{1\linewidth}{!}{
    \begin{tabular}{l|cccccccccccc}
    \toprule
    \multirow{2}{*}{Methods} & \multicolumn{2}{c}{Cora} & \multicolumn{2}{c}{ACM} & \multicolumn{2}{c}{Reddit} & \multicolumn{2}{c}{Wisconsin}  & \multicolumn{2}{c}{Photo} & \multicolumn{2}{c}{Elliptic} \\
    &Acc &MacroF &Acc &MacroF &Acc &MacroF &Acc &MacroF &Acc &MacroF &Acc &MacroF\\
    \midrule\midrule
    \method{GCN} &\underline{69.86}{\tiny$\pm$4.93} &\underline{68.39}{\tiny$\pm$4.21} &52.49{\tiny$\pm$12.80} &44.46{\tiny$\pm$16.02} &51.02{\tiny$\pm$20.66} &35.75{\tiny$\pm$10.64} &42.85{\tiny$\pm$4.74} &31.75{\tiny$\pm$3.62} &76.09{\tiny$\pm$7.03} &74.45{\tiny$\pm$6.73} &51.90{\tiny$\pm$12.45} &42.21{\tiny$\pm$7.92}\\
    \method{GAT} &68.57{\tiny$\pm$4.38} &67.55{\tiny$\pm$4.03} &39.28{\tiny$\pm$7.90} &26.88{\tiny$\pm$11.74} &49.74{\tiny$\pm$15.65} &35.81{\tiny$\pm$7.48} &40.35{\tiny$\pm$4.80} &30.58{\tiny$\pm$4.14} &76.11{\tiny$\pm$4.54} &75.25{\tiny$\pm$3.57} &50.43{\tiny$\pm$14.12} &40.17{\tiny$\pm$8.47}\\
    \method{Simple-HGN} &- &- &66.31{\tiny$\pm$20.71} &62.29{\tiny$\pm$26.23} &- &- &- &- &- &- &- &-\\
    \method{FAGCN} &- &- &- &- &- &- &\textbf{59.42}{\tiny$\pm$9.09} &\underline{47.42}{\tiny$\pm$6.96} &- &- &- &-\\\midrule
    \method{DGI} &58.70{\tiny$\pm$6.05} &55.59{\tiny$\pm$5.33} &42.48{\tiny$\pm$10.45} &38.31{\tiny$\pm$11.38} &50.85{\tiny$\pm$9.54} &36.41{\tiny$\pm$4.31} &31.93{\tiny$\pm$7.41} &25.35{\tiny$\pm$5.89} &56.69{\tiny$\pm$7.18} &55.48{\tiny$\pm$6.59} &44.95{\tiny$\pm$14.57} &35.26{\tiny$\pm$10.62}\\
    \method{GraphPrompt} &\textbf{72.78}{\tiny$\pm$2.65} &\textbf{69.58}{\tiny$\pm$2.71} &67.97{\tiny$\pm$12.91} &66.78{\tiny$\pm$14.02} &49.66{\tiny$\pm$18.04} &35.10{\tiny$\pm$8.83} &38.38{\tiny$\pm$4.22} &31.08{\tiny$\pm$4.00} &66.44{\tiny$\pm$6.83} &66.02{\tiny$\pm$5.80} &\underline{53.45}{\tiny$\pm$18.15} &\underline{42.73}{\tiny$\pm$13.81}\\
    \method{HeCo} &- &- &\textbf{86.81}{\tiny$\pm$1.83} &\textbf{86.90}{\tiny$\pm$1.84} &- &- &- &- &- &- &- &-\\
    \method{TGN} &- &- &- &- &48.57{\tiny$\pm$9.11} &35.65{\tiny$\pm$4.47} &- &- &- &- &- &-\\
    \method{DDGCL} &- &- &- &- &53.30{\tiny$\pm$11.44} &\underline{37.31}{\tiny$\pm$5.03} &- &- &- &- &- &-\\
    \method{DSSL} &- &- &- &- &- &- &49.24{\tiny$\pm$5.44} &41.63{\tiny$\pm$4.45} &- &- &- &-\\
    \midrule
    \method{GCOPE} &62.88{\tiny$\pm$3.86} &60.78{\tiny$\pm$3.60} &34.34{\tiny$\pm$1.42} &31.35{\tiny$\pm$2.47} &22.66{\tiny$\pm$23.82} &17.31{\tiny$\pm$13.55} &24.95{\tiny$\pm$6.06} &22.92{\tiny$\pm$5.09} &46.91{\tiny$\pm$3.89} &42.02{\tiny$\pm$4.16} &42.43{\tiny$\pm$8.01} &34.78{\tiny$\pm$4.34}\\
    \method{MDGPT} &61.79{\tiny$\pm$4.36} &60.96{\tiny$\pm$4.07} &77.11{\tiny$\pm$7.95} &76.54{\tiny$\pm$8.59} &43.53{\tiny$\pm$15.68} &32.45{\tiny$\pm$8.55} &36.15{\tiny$\pm$5.54} &31.39{\tiny$\pm$4.22} &\underline{80.14}{\tiny$\pm$4.47} &\textbf{79.30}{\tiny$\pm$3.83} &51.60{\tiny$\pm$10.34} &40.87{\tiny$\pm$6.87}\\
    \method{MDGFM} &51.03{\tiny$\pm$7.39} &50.29{\tiny$\pm$7.08} &67.83{\tiny$\pm$11.65} &67.41{\tiny$\pm$11.93} &\underline{53.42}{\tiny$\pm$19.37} &36.68{\tiny$\pm$9.44} &30.53{\tiny$\pm$7.64} &24.98{\tiny$\pm$5.94} &73.86{\tiny$\pm$5.18} &72.37{\tiny$\pm$4.47} &\textbf{55.92}{\tiny$\pm$9.88} &\textbf{44.37}{\tiny$\pm$6.93}\\
    \method{SAMGPT} &62.19{\tiny$\pm$5.32} &61.49{\tiny$\pm$4.91} &72.19{\tiny$\pm$6.33} &71.61{\tiny$\pm$6.71} &49.56{\tiny$\pm$14.88} &35.64{\tiny$\pm$7.26} &45.35{\tiny$\pm$7.50} &38.25{\tiny$\pm$5.24} &\textbf{80.36}{\tiny$\pm$4.63} &\underline{78.58}{\tiny$\pm$3.99} &53.12{\tiny$\pm$11.07} &42.66{\tiny$\pm$7.21}\\
    \method{G2P2} &57.25{\tiny$\pm$4.24} &56.82{\tiny$\pm$3.96} &70.77{\tiny$\pm$5.48} &69.34{\tiny$\pm$7.58} &47.31{\tiny$\pm$9.66} &23.55{\tiny$\pm$3.43} &42.63{\tiny$\pm$6.20} &36.10{\tiny$\pm$4.07} &- &- &44.75{\tiny$\pm$5.92} &30.25{\tiny$\pm$3.04}\\
    \method{GraphCLIP} &49.41{\tiny$\pm$4.44} &47.36{\tiny$\pm$3.80} &31.54{\tiny$\pm$2.29} &22.82{\tiny$\pm$1.54} &39.44{\tiny$\pm$10.60} &20.78{\tiny$\pm$3.72} &37.07{\tiny$\pm$5.63} &32.69{\tiny$\pm$4.39} &- &- &40.13{\tiny$\pm$11.00} &26.00{\tiny$\pm$4.29}\\
    \method{GFT} &64.15{\tiny$\pm$3.23} &63.56{\tiny$\pm$3.15} &56.66{\tiny$\pm$3.99} &55.17{\tiny$\pm$4.13} &\textbf{54.38}{\tiny$\pm$7.99} &\textbf{38.63}{\tiny$\pm$3.85} &\underline{57.19}{\tiny$\pm$6.03} &\textbf{50.36}{\tiny$\pm$4.74} &75.53{\tiny$\pm$5.10} &74.25{\tiny$\pm$4.55} &47.94{\tiny$\pm$8.13} &39.62{\tiny$\pm$4.89}\\
    \method{UniGraph2} &64.44{\tiny$\pm$4.09} &61.57{\tiny$\pm$3.78} &\underline{80.15}{\tiny$\pm$11.06} &\underline{78.50}{\tiny$\pm$12.37} &47.57{\tiny$\pm$14.68} &34.44{\tiny$\pm$7.29} &26.35{\tiny$\pm$6.77} &21.13{\tiny$\pm$5.19} &79.26{\tiny$\pm$5.61} &78.34{\tiny$\pm$4.99} &48.84{\tiny$\pm$8.35} &38.44{\tiny$\pm$5.10}\\\bottomrule
    \end{tabular}}
\end{table*}

\begin{table*}[tbp] % [!t]
    \centering
    \footnotesize
    \caption{Evaluation of 5-shot edge classification and graph classification on seen datasets.}
    \addtolength{\tabcolsep}{1.0mm}
    \label{table.exp2-5shotecgc}
    \resizebox{1\linewidth}{!}{
    \begin{tabular}{l|cc|cccccc}
    \toprule
    \multirow{3}{*}{Methods} & \multicolumn{2}{c|}{Edge classification} & \multicolumn{6}{c}{Graph classification}\\
     &\multicolumn{2}{c|}{FB15K-237} &\multicolumn{2}{c}{HIV} &\multicolumn{2}{c}{COX2} &\multicolumn{2}{c}{PROTEINS} \\
    &Acc &MacroF &Acc &MacroF &Acc &MacroF &Acc &MacroF \\
    \midrule\midrule
    \method{GCN} &\textbf{26.85}{\tiny$\pm$1.55} &\textbf{25.90}{\tiny$\pm$0.97} &46.39{\tiny$\pm$28.97} &30.91{\tiny$\pm$15.60} &58.73{\tiny$\pm$11.32} &50.45{\tiny$\pm$7.36} &57.79{\tiny$\pm$8.91} &55.05{\tiny$\pm$8.51}\\
    \method{GAT} &24.62{\tiny$\pm$1.33} &\underline{24.57}{\tiny$\pm$0.81} &48.22{\tiny$\pm$20.26} &34.07{\tiny$\pm$10.62} &57.52{\tiny$\pm$10.63} &50.90{\tiny$\pm$7.38} &58.40{\tiny$\pm$8.52} &56.20{\tiny$\pm$8.28}\\
    %\method{Simple-HGN} \\
    %\method{TGN} \\
    %\method{FAGFN} \\
    \midrule
    \method{DGI} &8.54{\tiny$\pm$1.23} &9.40{\tiny$\pm$0.58} &\underline{57.22}{\tiny$\pm$17.97} &\underline{38.35}{\tiny$\pm$7.93} &51.93{\tiny$\pm$10.63} &45.77{\tiny$\pm$6.65} &56.77{\tiny$\pm$8.38} &52.55{\tiny$\pm$7.95}\\
    %\method{HeCo} \\
    %\method{DDGCL} \\
    %\method{DSSL} \\
    \method{GraphPrompt} &12.47{\tiny$\pm$1.16} &10.42{\tiny$\pm$0.55} &51.07{\tiny$\pm$9.62} &36.41{\tiny$\pm$4.35} &53.11{\tiny$\pm$7.82} &47.14{\tiny$\pm$4.53} &54.35{\tiny$\pm$8.31} &52.96{\tiny$\pm$7.86}\\\midrule
    \method{GCOPE} &1.92{\tiny$\pm$0.33} &1.20{\tiny$\pm$0.14} &\textbf{66.47}{\tiny$\pm$26.43} &\textbf{41.41}{\tiny$\pm$12.13} &38.11{\tiny$\pm$17.34} &32.81{\tiny$\pm$11.76} &\textbf{64.15}{\tiny$\pm$8.17} &\textbf{63.44}{\tiny$\pm$8.09}\\
    \method{MDGPT} &22.48{\tiny$\pm$1.38} &18.55{\tiny$\pm$0.62} &48.35{\tiny$\pm$10.12} &35.34{\tiny$\pm$5.11} &51.76{\tiny$\pm$7.61} &47.43{\tiny$\pm$5.15} &53.50{\tiny$\pm$9.94} &48.71{\tiny$\pm$11.18}\\
    \method{MDGFM} &17.62{\tiny$\pm$1.26} &15.58{\tiny$\pm$1.06} &49.36{\tiny$\pm$20.94} &34.45{\tiny$\pm$10.43} &58.97{\tiny$\pm$22.17} &41.10{\tiny$\pm$11.54} &53.86{\tiny$\pm$8.70} &51.89{\tiny$\pm$8.81}\\
    \method{SAMGPT} &22.30{\tiny$\pm$1.22} &18.14{\tiny$\pm$0.81} &51.96{\tiny$\pm$13.65} &36.85{\tiny$\pm$6.47} &\textbf{60.12}{\tiny$\pm$7.84} &\textbf{53.42}{\tiny$\pm$6.01} &57.43{\tiny$\pm$8.06} &55.59{\tiny$\pm$7.89}\\
    \method{G2P2} &\underline{24.87}{\tiny$\pm$1.19} &22.30{\tiny$\pm$0.65} &50.76{\tiny$\pm$11.37} &36.52{\tiny$\pm$5.41} &58.48{\tiny$\pm$7.52} &\underline{51.79}{\tiny$\pm$5.50} &56.83{\tiny$\pm$7.22} &54.41{\tiny$\pm$6.73}\\
    \method{GraphCLIP} &3.09{\tiny$\pm$0.44} &1.30{\tiny$\pm$0.17} &52.30{\tiny$\pm$17.30} &36.66{\tiny$\pm$9.08} &48.34{\tiny$\pm$19.32} &39.03{\tiny$\pm$11.49} &\underline{60.49}{\tiny$\pm$7.01} &\underline{59.65}{\tiny$\pm$6.80}\\
    \method{GFT} &7.71{\tiny$\pm$0.61} &7.05{\tiny$\pm$0.50} &52.20{\tiny$\pm$12.65} &36.84{\tiny$\pm$5.90} &\underline{59.34}{\tiny$\pm$16.56} &47.67{\tiny$\pm$11.90} &56.71{\tiny$\pm$8.81} &52.38{\tiny$\pm$10.55}\\
    \method{UniGraph2} &2.51{\tiny$\pm$1.97} &1.52{\tiny$\pm$0.13} &50.57{\tiny$\pm$7.64} &33.40{\tiny$\pm$3.69} &50.01{\tiny$\pm$14.81} &42.79{\tiny$\pm$8.64} &53.10{\tiny$\pm$12.27} &51.18{\tiny$\pm$11.26}\\\bottomrule
    \end{tabular}}
\end{table*}

%%%%%%%%%%%%%%%%%%%Time

\begin{table*}[tbp] % [!t]
    \centering
    \footnotesize
    \caption{Time and time/epoch of downstream tuning for 1-shot node classification on seen datasets.}
    \addtolength{\tabcolsep}{1.0mm}
    \label{table.exp2-1shotnc-t}
    \resizebox{1\linewidth}{!}{
    \begin{tabular}{l|cccccccccccc}
    \toprule
    \multirow{2}{*}{Methods} & \multicolumn{2}{c}{Cora} & \multicolumn{2}{c}{ACM} & \multicolumn{2}{c}{Reddit} & \multicolumn{2}{c}{Wisconsin}  & \multicolumn{2}{c}{Photo} & \multicolumn{2}{c}{Elliptic} \\
    &Time &Avg. T. &Time &Avg. T. &Time &Avg. T. &Time &Avg. T. &Time &Avg. T. &Time &Avg. T.\\
    \midrule\midrule
    \method{GCN} &31.31 &0.0140 &122.84 &0.0636 &127.02 &0.1119 &129.34 &0.0255 &828.39 &0.0698 &131.64 &0.1707\\
    \method{GAT} &50.53 &0.0298 &184.55 &0.0955 &159.57 &0.1886 &104.03 &0.0406 &410.57 &0.0788 &174.17 &0.2399\\
    \method{Simple-HGN} &- &- &3764.31 &0.2510 &- &- &- &- &- &- &- &-\\
    \method{TGN} &- &- &- &- &1660.39 &0.1660 &- &- &- &- &- &-\\
    \method{FAGFN} &- &- &- &- &- &- &217.31 &0.0232 &- &- &- &-\\\midrule
    \method{DGI} &33.32 &0.0577 &565.19 &0.9661 &2269.40 &4.0743 &97.84 &0.0142 &625.27 &0.7018 &306.42 &0.5211\\
    \method{HeCo} &- &- &811.97 &0.3248 &- &- &- &- &- &- &- &-\\
    \method{DDGCL} &- &- &- &- &5242.64 &0.5243 &- &- &- &- &- &-\\
    \method{DSSL} &- &- &- &- &- &- &25.98 &0.0026 &- &- &- &-\\
    \method{GraphPrompt} &26.02 &0.0109 &108.08 &0.1965 &2016.28 &3.6660 &126.91 &0.0489 &763.13 &1.1875 &287.58 &0.2498\\\midrule
    \method{GCOPE} &1889.88 &0.3780 &2011.55 &0.4023 &4133.96 &0.8268 &1385.02 &0.2770 &2791.60 &0.5583 &37958.68 &7.2870\\
    \method{MDGPT} &572.78 &0.0286 &1598.63 &0.0886 &3652.43 &0.1826 &560.69 &0.0280 &1057.15 &0.0553 &6157.62 &0.3573\\
    \method{MDGFM} &1244.31 &0.0858 &2104.96 &0.2524 &580.64 &0.1847 &445.44 &0.0751 &799.05 &0.0625 &55421.48 &3.7830\\
    \method{SAMGPT} &1244.34 &0.0624 &3563.67 &0.2179 &7999.01 &0.4203 &1428.11 &0.0729 &2796.97 &0.1424 &25579.69 &1.2908\\
    \method{G2P2} &315.28 &0.1261 &415.70 &0.1663 &421.09 &0.1684 &408.96 &0.1636 &- &- &436.85 &0.1747\\
    \method{GraphCLIP} &818.55 &0.1637 &1008.32 &0.2017 &781.38 &0.1563 &797.49 &0.1595 &- &- &1423.74 &0.2847\\
    \method{GFT} &257.49 &0.0138 &1506.18 &0.3203 &4037.43 &0.8261 &338.40 &0.1055 &1228.08 &0.0756 &1793.17 &0.3915\\
    \method{UniGraph2} &73.43 &- &1877.38 &- &867.44 &- &107.03 &- &1233.20 &- &4653.17 &-\\\bottomrule
    \end{tabular}}
\end{table*}

\begin{table*}[tbp] % [!t]
    \centering
    \footnotesize
    \caption{Time and time/epoch of downstream tuning for 5-shot node classification on seen datasets.}
    \addtolength{\tabcolsep}{1.0mm}
    \label{table.exp2-5shotnc-t}
    \resizebox{1\linewidth}{!}{
    \begin{tabular}{l|cccccccccccc}
    \toprule
    \multirow{2}{*}{Methods} & \multicolumn{2}{c}{Cora} & \multicolumn{2}{c}{ACM} & \multicolumn{2}{c}{Reddit} & \multicolumn{2}{c}{Wisconsin}  & \multicolumn{2}{c}{Photo} & \multicolumn{2}{c}{Elliptic} \\
    &Time &Avg. T. &Time &Avg. T. &Time &Avg. T. &Time &Avg. T. &Time &Avg. T. &Time &Avg. T.\\
    \midrule\midrule
    \method{GCN} &257.61 &0.0297 &246.73 &0.0681 &318.01 &0.1295 &560.74 &0.0310 &196.12 &0.0269 &171.76 &0.1286\\
    \method{GAT} &296.70 &0.0464 &228.42 &0.1224 &220.35 &0.1805 &195.15 &0.0213 &500.67 &0.0357 &245.24 &0.1822\\
    \method{Simple-HGN} &- &- &3765.52 &0.2510 &- &- &- &- &- &- &- &-\\
    \method{TGN} &- &- &- &- &1664.02 &0.1664 &- &- &- &- &- &-\\
    \method{FAGFN} &- &- &- &- &- &- &282.46 &0.0303 &- &- &- &-\\\midrule
    \method{DGI} &46.37 &0.0594 &676.29 &1.0926 &2811.11 &4.9842 &130.31 &0.0253 &719.83 &1.0883 &331.31 &0.5895\\
    \method{HeCo} &- &- &812.28 &0.3249 &- &- &- &- &- &- &- &-\\
    \method{DDGCL} &- &- &- &- &5264.01 &0.5264 &- &- &- &- &- &-\\
    \method{DSSL} &- &- &- &- &- &- &33.75 &0.0034 &- &- &- &-\\
    \method{GraphPrompt} &28.18 &0.0127 &183.35 &0.1991 &2599.13 &4.2819 &130.49 &0.0511 &790.62 &1.2425 &327.36 &0.2775\\\midrule
    \method{GCOPE} &1884.36 &0.3769 &1979.95 &0.3960 &4175.79 &0.8352 &1379.19 &0.2758 &2788.78 &0.5578 &37996.33 &7.3405\\
    \method{MDGPT} &567.61 &0.0288 &1604.10 &0.0884 &3654.25 &0.1827 &559.22 &0.0280 &917.99 &0.0536 &5731.58 &0.3593\\
    \method{MDGFM} &1286.06 &0.0867 &1999.56 &0.2566 &575.39 &0.1860 &336.84 &0.0581 &764.06 &0.0667 &38802.31 &3.8225\\
    \method{SAMGPT} &1261.57 &0.0631 &4639.72 &0.2320 &7674.79 &0.4112 &1206.69 &0.0603 &3308.01 &0.1654 &25805.74 &1.2676\\
    \method{G2P2} &328.73 &0.1315 &415.54 &0.1662 &426.76 &0.1707 &403.93 &0.1616 &- &- &430.23 &0.1721\\
    \method{GraphCLIP} &892.39 &0.1785 &1047.32 &0.2095 &857.76 &0.1716 &952.70 &0.1905 &- &- &1389.87 &0.2780\\
    \method{GFT} &241.75 &0.0201 &1591.40 &0.3318 &3819.50 &0.8196 &385.16 &0.1241 &1267.71 &0.0771 &1441.96 &0.3497\\
    \method{UniGraph2} &73.02 &- &2168.43 &- &861.27 &- &125.94 &- &1907.82 &- &4226.65 &-\\\bottomrule
    \end{tabular}}
\end{table*}

\begin{table*}[tbp] % [!t]
    \centering
    \footnotesize
    \caption{Time and time/epoch of downstream tuning for 1-shot edge classification and graph classification on seen datasets.}
    \addtolength{\tabcolsep}{1.0mm}
    \label{table.exp2-1shotecgc-t}
    \resizebox{0.8\linewidth}{!}{
    \begin{tabular}{l|cc|cccccc}
    \toprule
    \multirow{3}{*}{Methods} & \multicolumn{2}{c|}{Edge classification} & \multicolumn{6}{c}{Graph classification}\\
     &\multicolumn{2}{c|}{FB15K-237} &\multicolumn{2}{c}{HIV} &\multicolumn{2}{c}{COX2} &\multicolumn{2}{c}{PROTEINS} \\
    &Time &Avg. T. &Time &Avg. T. &Time &Avg. T. &Time &Avg. T. \\
    \midrule\midrule
    \method{GCN} &1220.51 &0.0639 &166.80 &0.0414 &47.53 &0.0250 &44.65 &0.0346\\
    \method{GAT} &923.32 &0.0782 &145.08 &0.0715 &42.43 &0.0280 &39.23 &0.0346\\
    %\method{Simple-HGN} \\
    %\method{TGN} \\
    %\method{FAGFN} \\
    \midrule
    \method{DGI} &4468.84 &3.4212 &3769.68 &2.8540 &43.64 &0.0248 &119.66 &0.2176\\
    %\method{HeCo} \\
    %\method{DDGCL} \\
    %\method{DSSL} \\
    \method{GraphPrompt} &2239.06 &0.2528 &2495.85 &2.5379 &35.17 &0.0640 &181.10 &0.3293\\\midrule
    \method{GCOPE} &2361.41 &0.4723 &2029.85 &0.4060 &1465.00 &0.2930 &1518.61 &0.3037\\
    \method{MDGPT} &3073.88 &0.1537 &369.77 &0.0267 &376.72 &0.0236 &53.92 &0.0109\\
    \method{MDGFM} &4454.41 &0.2445 &1443.36 &0.1282 &321.10 &0.0518 &483.72 &0.0525\\
    \method{SAMGPT} &5435.56 &0.2718 &1229.38 &0.0630 &801.66 &0.0448 &1085.71 &0.0557\\
    \method{G2P2} &2013.12 &0.8052 &331.19 &0.1325 &307.83 &0.1231 &309.76 &0.1239\\
    \method{GraphCLIP} &5032.46 &1.0065 &2913.34 &0.5827 &718.89 &0.1438 &738.50 &0.1477\\
    \method{GFT} &6007.92 &1.2729 &3389.08 &0.9007 &410.75 &0.0785 &700.49 &0.1712\\
    \method{UniGraph2} &5643.13 &- &2835.38 &- &686.40 &- &1159.89 &-\\\bottomrule
    \end{tabular}}
\end{table*}

\begin{table*}[tbp] % [!t]
    \centering
    \footnotesize
    \caption{Time and time/epoch of downstream tuning for 5-shot edge classification and graph classification on seen datasets.}
    \addtolength{\tabcolsep}{1.0mm}
    \label{table.exp2-5shotecgc-t}
    \resizebox{0.8\linewidth}{!}{
    \begin{tabular}{l|cc|cccccc}
    \toprule
    \multirow{3}{*}{Methods} & \multicolumn{2}{c|}{Edge classification} & \multicolumn{6}{c}{Graph classification}\\
     &\multicolumn{2}{c|}{FB15K-237} &\multicolumn{2}{c}{HIV} &\multicolumn{2}{c}{COX2} &\multicolumn{2}{c}{PROTEINS} \\
    &Time &Avg. T. &Time &Avg. T. &Time &Avg. T. &Time &Avg. T. \\
    \midrule\midrule
    \method{GCN} &1002.11 &0.0796 &332.07 &0.0303 &375.00 &0.0277 &127.64 &0.0163\\
    \method{GAT} &779.50 &0.0876 &290.26 &0.0269 &95.51 &0.0110 &119.32 &0.0121\\
    %\method{Simple-HGN} \\
    %\method{TGN} \\
    %\method{FAGFN} \\
    \midrule
    \method{DGI} &4640.11 &3.6991 &3807.46 &2.9227 &56.63 &0.0302 &131.69 &0.2394\\
    %\method{HeCo} \\
    %\method{DDGCL} \\
    %\method{DSSL} \\
    \method{GraphPrompt} &2599.97 &0.2667 &2651.73 &2.8213 &44.45 &0.0808 &181.88 &0.3307\\\midrule
    \method{GCOPE} &2371.46 &0.4743 &2014.62 &0.4029 &1450.19 &0.2900 &1508.49 &0.3017\\
    \method{MDGPT} &3394.25 &0.1697 &314.36 &0.0335 &332.59 &0.0267 &91.82 &0.0329\\
    \method{MDGFM} &4498.39 &0.2447 &1397.38 &0.1435 &460.08 &0.0597 &411.02 &0.0518\\
    \method{SAMGPT} &5450.74 &0.2725 &1408.20 &0.0704 &1156.79 &0.0641 &695.99 &0.0393\\
    \method{G2P2} &2013.91 &0.8056 &340.58 &0.1362 &316.66 &0.1267 &295.84 &0.1183\\
    \method{GraphCLIP} &11790.04 &2.3580 &3306.35 &0.6613 &748.74 &0.1497 &999.50 &0.1999\\
    \method{GFT} &10723.83 &2.6556 &3459.66 &1.0651 &469.25 &0.1450 &407.72 &0.1432\\
    \method{UniGraph2} &5686.28 &- &3548.80 &- &620.54 &- &1241.42 &-\\\bottomrule
    \end{tabular}}
\end{table*}

\subsection{Topic Domain Adaptation}\label{app.exp3}
We further report the 1shot Macro-F1 of node classification results for topic domain adaptation in Table.~\ref{table.exp3-1shotnc-macrof},
the 5-shot accuracy and Macro-F1 of node classification results in Table~\ref{table.exp3-5shotnc} and Table~\ref{table.exp3-5shotnc-macrof}, and the 5-shot edge and graph classification results in Table~\ref{table.exp3-5shotecgc}.

\begin{table*}[tbp] % [!t]
    \centering
    \footnotesize
    \caption{Marco-F1 of 1-shot node classification on topic domain adaptation.}
    \vspace{-2mm}
    \addtolength{\tabcolsep}{0.5mm}
    \label{table.exp3-1shotnc-macrof}
    %\resizebox{1\linewidth}{!}{
    \begin{tabular}{l|cc|ccc|c|cc|c}
    \toprule
    \multirow{2}{*}{Methods} &\multicolumn{2}{c|}{Citation} &\multicolumn{3}{c|}{Social \& Web} &{E-commerce} &\multicolumn{2}{c|}{Finance} &{Proteins} \\
    &Pubmed &ogbn-mag & {Wikipedia} & {Actor}  & {Chameleon} & {Products} & {T-Finance} & {DGraph} & {ogbn-proteins}\\
    \midrule\midrule
    \method{GCOPE} &36.09{\tiny$\pm$6.14} &0.14{\tiny$\pm$0.01} &34.55{\tiny$\pm$14.27} &17.67{\tiny$\pm$4.50} &19.46{\tiny$\pm$3.48} &OOT &26.43{\tiny$\pm$12.11} &OOT &34.16{\tiny$\pm$7.92} \\
    \method{MDGPT} &45.38{\tiny$\pm$11.09} &\underline{3.41}{\tiny$\pm$0.39} &33.40{\tiny$\pm$14.27} &21.76{\tiny$\pm$5.68} &\underline{24.47}{\tiny$\pm$4.99} &\underline{9.81}{\tiny$\pm$1.54} &\underline{43.76}{\tiny$\pm$16.18} &\underline{26.27}{\tiny$\pm$6.42} &\underline{44.41}{\tiny$\pm$9.33} \\
    \method{MDGFM} &29.01{\tiny$\pm$5.05} &0.83{\tiny$\pm$0.85} &32.82{\tiny$\pm$14.06} &13.97{\tiny$\pm$3.09} &21.24{\tiny$\pm$4.12} &1.63{\tiny$\pm$0.34} &36.51{\tiny$\pm$12.34} &21.65{\tiny$\pm$12.89} &43.64{\tiny$\pm$6.17} \\
    \method{SAMGPT} &\underline{46.20}{\tiny$\pm$7.43} &\textbf{4.98}{\tiny$\pm$0.27} &\underline{36.27}{\tiny$\pm$14.67} &19.74{\tiny$\pm$4.39} &\textbf{25.21}{\tiny$\pm$4.37} &\textbf{13.33}{\tiny$\pm$1.35} &37.19{\tiny$\pm$9.70} &\textbf{27.00}{\tiny$\pm$7.55} &\textbf{49.89}{\tiny$\pm$10.20} \\
    \method{G2P2} &42.16{\tiny$\pm$8.28} &- &21.80{\tiny$\pm$2.66} &\textbf{22.98}{\tiny$\pm$6.09} &- &- &\textbf{44.30}{\tiny$\pm$10.97} &22.99{\tiny$\pm$3.99} &- \\
    \method{GraphCLIP} &32.77{\tiny$\pm$4.16} &- &20.08{\tiny$\pm$8.40} &16.77{\tiny$\pm$2.55} &- &- &32.49{\tiny$\pm$11.49} &21.52{\tiny$\pm$12.25} &-\\
    \method{GFT} &42.78{\tiny$\pm$7.76} &0.93{\tiny$\pm$0.10} &\textbf{38.13}{\tiny$\pm$13.54} &22.24{\tiny$\pm$5.00} &22.98{\tiny$\pm$4.76} &7.24{\tiny$\pm$1.15} &42.08{\tiny$\pm$15.08} &25.54{\tiny$\pm$7.40} &43.24{\tiny$\pm$11.16} \\
    \method{UniGraph2} &\textbf{48.84}{\tiny$\pm$10.27} &1.02{\tiny$\pm$0.50} &31.33{\tiny$\pm$10.57} &\underline{22.32}{\tiny$\pm$5.20} &22.85{\tiny$\pm$4.09} &7.47{\tiny$\pm$1.28} &36.63{\tiny$\pm$9.27} &19.34{\tiny$\pm$4.96} &40.80{\tiny$\pm$7.94}\\\bottomrule
    \end{tabular}%}
\end{table*}

\begin{table*}[tbp] % [!t]
    \centering
    \footnotesize
    \caption{Accuracy of 5-shot node classification on topic domain adaptation.}
    \vspace{-2mm}
    \addtolength{\tabcolsep}{0.5mm}
    \label{table.exp3-5shotnc}
    %\resizebox{1\linewidth}{!}{
    \begin{tabular}{l|cc|ccc|c|cc|c}
    \toprule
    \multirow{2}{*}{Methods} &\multicolumn{2}{c|}{Citation} &\multicolumn{3}{c|}{Social \& Web} &{E-commerce} &\multicolumn{2}{c|}{Finance} &{Proteins} \\
    &Pubmed &ogbn-mag & {Wikipedia} & {Actor}  & {Chameleon} & {Products} & {T-Finance} & {DGraph} & {ogbn-proteins}\\
    \midrule\midrule
    \method{GCOPE} &55.76{\tiny$\pm$5.66} &0.26{\tiny$\pm$0.02} &53.74{\tiny$\pm$20.62} &30.36{\tiny$\pm$8.68} &25.42{\tiny$\pm$2.32} &OOT &29.14{\tiny$\pm$16.34} &OOT &41.77{\tiny$\pm$9.42}\\
    \method{MDGPT} &\underline{62.51}{\tiny$\pm$6.95} &\underline{9.55}{\tiny$\pm$0.90} &\underline{58.41}{\tiny$\pm$17.83} &37.83{\tiny$\pm$7.13} &33.79{\tiny$\pm$2.88} &\underline{21.28}{\tiny$\pm$9.81} &61.10{\tiny$\pm$12.17} &39.44{\tiny$\pm$9.47} &53.75{\tiny$\pm$10.80}\\
    \method{MDGFM} &42.34{\tiny$\pm$7.76} &OOM &49.89{\tiny$\pm$19.61} &19.70{\tiny$\pm$4.81} &27.50{\tiny$\pm$4.17} &6.47{\tiny$\pm$2.25} &49.76{\tiny$\pm$17.06} &\underline{43.37}{\tiny$\pm$17.01} &50.50{\tiny$\pm$6.30}\\
    \method{SAMGPT} &61.36{\tiny$\pm$5.91} &\textbf{11.23}{\tiny$\pm$0.66} &54.74{\tiny$\pm$14.32} &\textbf{38.92}{\tiny$\pm$5.53} &\textbf{37.75}{\tiny$\pm$3.44} &\textbf{31.11}{\tiny$\pm$6.32} &49.73{\tiny$\pm$9.53} &40.75{\tiny$\pm$8.97} &\underline{55.17}{\tiny$\pm$9.88}\\
    \method{G2P2} &61.21{\tiny$\pm$6.44} &- &47.93{\tiny$\pm$8.18} &\underline{38.05}{\tiny$\pm$5.40} &- &- &64.46{\tiny$\pm$11.21} &\textbf{44.04}{\tiny$\pm$8.22} &- \\
    \method{GraphCLIP} &42.82{\tiny$\pm$3.73} &- &40.54{\tiny$\pm$13.30} &23.73{\tiny$\pm$4.65} &- &- &48.45{\tiny$\pm$22.19} &39.19{\tiny$\pm$16.43} &-\\
    \method{GFT} &62.16{\tiny$\pm$5.24} &5.45{\tiny$\pm$3.45} &\textbf{60.25}{\tiny$\pm$11.82} &37.68{\tiny$\pm$5.20} &\underline{34.83}{\tiny$\pm$3.16} &21.18{\tiny$\pm$4.86} &\textbf{69.27}{\tiny$\pm$14.32} &42.89{\tiny$\pm$10.94} &\textbf{55.60}{\tiny$\pm$11.10} \\
    \method{UniGraph2} &\textbf{63.87}{\tiny$\pm$4.48} &3.91{\tiny$\pm$0.88} &53.45{\tiny$\pm$21.50} &36.53{\tiny$\pm$6.23} &30.57{\tiny$\pm$2.47} &17.81{\tiny$\pm$8.43} &\underline{65.87}{\tiny$\pm$9.00} &40.85{\tiny$\pm$10.51} &51.33{\tiny$\pm$8.50}\\\bottomrule
    \end{tabular}%}
\end{table*}

\begin{table*}[tbp] % [!t]
    \centering
    \footnotesize
    \caption{Marco-F1 of 5-shot node classification on topic domain adaptation.}
    \vspace{-2mm}
    \addtolength{\tabcolsep}{0.5mm}
    \label{table.exp3-5shotnc-macrof}
    %\resizebox{1\linewidth}{!}{
    \begin{tabular}{l|cc|ccc|c|cc|c}
    \toprule
    \multirow{2}{*}{Methods} &\multicolumn{2}{c|}{Citation} &\multicolumn{3}{c|}{Social \& Web} &{E-commerce} &\multicolumn{2}{c|}{Finance} &{Proteins} \\
    &Pubmed &ogbn-mag & {Wikipedia} & {Actor}  & {Chameleon} & {Products} & {T-Finance} & {DGraph} & {ogbn-proteins}\\
    \midrule\midrule
    \method{GCOPE} &55.38{\tiny$\pm$5.62} &0.14{\tiny$\pm$0.01} &36.23{\tiny$\pm$10.58} &24.36{\tiny$\pm$5.49} &24.39{\tiny$\pm$2.11} &OOT &24.05{\tiny$\pm$9.55} &OOT &31.69{\tiny$\pm$7.16}\\
    \method{MDGPT} &\textbf{62.55}{\tiny$\pm$6.89} &\underline{7.20}{\tiny$\pm$0.51} &\underline{38.30}{\tiny$\pm$8.14} &30.31{\tiny$\pm$4.24} &31.96{\tiny$\pm$3.40} &8.18{\tiny$\pm$2.32} &45.30{\tiny$\pm$7.33} &\underline{29.07}{\tiny$\pm$4.80} &\underline{47.79}{\tiny$\pm$9.92}\\
    \method{MDGFM} &39.55{\tiny$\pm$7.68} &OOM &34.33{\tiny$\pm$9.29} &15.39{\tiny$\pm$2.36} &26.50{\tiny$\pm$4.72} &3.45{\tiny$\pm$0.80} &38.26{\tiny$\pm$10.09} &28.54{\tiny$\pm$6.77} &47.36{\tiny$\pm$4.75} \\
    \method{SAMGPT} &61.06{\tiny$\pm$5.94} &\textbf{9.24}{\tiny$\pm$0.32} &37.08{\tiny$\pm$6.25} &30.88{\tiny$\pm$3.47} &\textbf{37.01}{\tiny$\pm$3.46} &\underline{15.53}{\tiny$\pm$1.90} &39.05{\tiny$\pm$5.66} &\textbf{29.91}{\tiny$\pm$4.46} &\textbf{53.46}{\tiny$\pm$9.39}\\
    \method{G2P2} &60.53{\tiny$\pm$6.56} &- &22.97{\tiny$\pm$2.58} &\textbf{32.22}{\tiny$\pm$3.73} &- &- &\underline{47.66}{\tiny$\pm$6.83} &25.55{\tiny$\pm$2.96} &- \\
    \method{GraphCLIP} &41.90{\tiny$\pm$4.23} &- &20.61{\tiny$\pm$5.50} &17.85{\tiny$\pm$2.32} &- &- &34.15{\tiny$\pm$11.96} &23.13{\tiny$\pm$7.10} &-\\
    \method{GFT} &61.54{\tiny$\pm$5.32} &4.18{\tiny$\pm$0.39} &\textbf{40.41}{\tiny$\pm$5.70} &\underline{31.73}{\tiny$\pm$3.20} &\underline{33.70}{\tiny$\pm$3.44} &10.18{\tiny$\pm$1.12} &\textbf{51.73}{\tiny$\pm$10.16} &21.98{\tiny$\pm$4.06} &46.05{\tiny$\pm$6.11} \\
    \method{UniGraph2} &\underline{62.28}{\tiny$\pm$4.78} &3.08{\tiny$\pm$0.86} &33.50{\tiny$\pm$9.74} &29.65{\tiny$\pm$4.29} &25.86{\tiny$\pm$2.93} &\textbf{20.81}{\tiny$\pm$1.14} &39.54{\tiny$\pm$9.32} &19.97{\tiny$\pm$5.64} &43.71{\tiny$\pm$3.66}\\\bottomrule
    \end{tabular}%}
\end{table*}

\begin{table*}[tbp] % [!t]
    \centering
    \footnotesize
    \caption{Evaluation of 5-shot edge classification and graph classification on topic domain adaptation.}
    \addtolength{\tabcolsep}{1.0mm}
    \label{table.exp3-5shotecgc}
    \resizebox{1\linewidth}{!}{
    \begin{tabular}{l|cc|cccc|cccc}
    \toprule
    \multirow{3}{*}{Methods} &\multicolumn{2}{c|}{Finance} &\multicolumn{4}{c|}{Common sense} &\multicolumn{4}{c}{Molecule}\\
    &\multicolumn{2}{c}{DGraph} &\multicolumn{2}{c}{Wiki} &\multicolumn{2}{c|}{WN18RR} &\multicolumn{2}{c}{PCBA} &\multicolumn{2}{c}{BZR} \\
    &Acc &MacroF &Acc &MacroF &Acc &MacroF &Acc &MacroF &Acc &MacroF\\
    \midrule\midrule
    \method{GCOPE} &OOT &OOT &OOT &OOT &18.15{\tiny$\pm$4.25} &13.59{\tiny$\pm$2.37} &51.25{\tiny$\pm$15.99} &33.35{\tiny$\pm$7.09} &55.69{\tiny$\pm$12.07} &50.56{\tiny$\pm$8.51}\\
    \method{MDGPT} &10.13{\tiny$\pm$2.59} &\underline{10.13}{\tiny$\pm$2.59} &\underline{8.79}{\tiny$\pm$1.63} &\underline{2.24}{\tiny$\pm$0.26} &\underline{19.27}{\tiny$\pm$3.32} &\underline{15.54}{\tiny$\pm$2.12} &\textbf{61.39}{\tiny$\pm$15.70} &\textbf{37.73}{\tiny$\pm$6.11} &57.21{\tiny$\pm$9.53} &\underline{53.11}{\tiny$\pm$7.66}\\
    \method{MDGFM} &9.42{\tiny$\pm$2.35} &6.47{\tiny$\pm$0.81} &4.78{\tiny$\pm$2.52} &1.07{\tiny$\pm$0.15} &15.15{\tiny$\pm$3.33} &10.90{\tiny$\pm$1.64} &54.32{\tiny$\pm$30.48} &32.64{\tiny$\pm$14.65} &39.62{\tiny$\pm$19.77} &31.45{\tiny$\pm$8.42}\\
    \method{SAMGPT} &\underline{10.41}{\tiny$\pm$1.48} &7.71{\tiny$\pm$0.57} &\textbf{15.11}{\tiny$\pm$2.72} &\textbf{5.90}{\tiny$\pm$0.24} &18.06{\tiny$\pm$3.49} &14.70{\tiny$\pm$1.84} &\underline{57.40}{\tiny$\pm$11.87} &\underline{36.38}{\tiny$\pm$4.76} &\textbf{66.34}{\tiny$\pm$9.90} &\textbf{58.03}{\tiny$\pm$6.84}\\
    \method{G2P2} &- &- &- &- &18.82{\tiny$\pm$3.85} &15.08{\tiny$\pm$1.98} &- &- &\underline{59.48}{\tiny$\pm$10.30} &52.40{\tiny$\pm$6.52}\\
    \method{GraphCLIP} &- &- &- &- &\textbf{20.76}{\tiny$\pm$5.17} &\textbf{15.61}{\tiny$\pm$2.01} &- &- &50.74{\tiny$\pm$13.78} &44.57{\tiny$\pm$7.66}\\
    \method{GFT} &\textbf{11.37}{\tiny$\pm$5.57} &\textbf{10.42}{\tiny$\pm$0.82} &OOT &OOT &17.93{\tiny$\pm$4.85} &13.21{\tiny$\pm$1.31} &46.36{\tiny$\pm$22.55} &29.94{\tiny$\pm$11.77} &59.43{\tiny$\pm$20.60} &45.18{\tiny$\pm$14.45}\\
    \method{UniGraph2} &10.00{\tiny$\pm$3.12} &5.52{\tiny$\pm$1.23} &5.30{\tiny$\pm$1.90} &1.33{\tiny$\pm$0.51} &10.37{\tiny$\pm$5.97} &8.48{\tiny$\pm$1.51} &50.55{\tiny$\pm$11.13} &32.76{\tiny$\pm$5.07} &51.96{\tiny$\pm$4.04} &46.77{\tiny$\pm$3.10}\\\bottomrule
    \end{tabular}}
\end{table*}

\subsection{Format Domain Adaptation}\label{app.exp4}
We show the 5-shot node classification results for format domain adaptation in Table~\ref{table.exp4-5shotnc}, and the 5-shot edge and graph classification results in Table~\ref{table.exp4-5shotecgc}.

\begin{table*}[tbp] % [!t]
    \centering
    \footnotesize
    \caption{Evaluation of 5-shot node classification on format domain adaptation.}
    \addtolength{\tabcolsep}{1.0mm}
    \label{table.exp4-5shotnc}
    \resizebox{1\linewidth}{!}{
    \begin{tabular}{l|cc|cccc|cc|cccc}
    \toprule
    \multirow{3}{*}{Methods} &\multicolumn{2}{c|}{Heterogeneous} &\multicolumn{4}{c|}{Heterophilic} &\multicolumn{2}{c|}{Dynamic} &\multicolumn{4}{c}{Textual} \\
     & \multicolumn{2}{c|}{ogbn-mag} & \multicolumn{2}{c}{Chameleon} & \multicolumn{2}{c|}{Actor} & \multicolumn{2}{c|}{Wikipedia} &\multicolumn{2}{c}{Pubmed} & \multicolumn{2}{c}{Products}  \\
     &Acc &MacroF &Acc &MacroF &Acc &MacroF &Acc &MacroF &Acc &MacroF &Acc &MacroF\\
    \midrule\midrule
    \method{GCOPE} &0.34{\tiny$\pm$0.05} &0.11{\tiny$\pm$0.01} &26.77{\tiny$\pm$1.82} &25.24{\tiny$\pm$1.90} &27.24{\tiny$\pm$7.28} &23.32{\tiny$\pm$4.79} &\underline{57.89}{\tiny$\pm$16.82} &\underline{38.58}{\tiny$\pm$7.89} &56.56{\tiny$\pm$5.41} &55.48{\tiny$\pm$5.32} &OOT &OOT\\
    \method{MDGPT} &\underline{11.33}{\tiny$\pm$0.64} &\underline{8.63}{\tiny$\pm$0.37} &34.58{\tiny$\pm$3.58} &32.84{\tiny$\pm$4.24} &\underline{38.14}{\tiny$\pm$7.42} &\underline{30.52}{\tiny$\pm$4.47} &49.72{\tiny$\pm$15.67} &34.65{\tiny$\pm$7.57} &59.52{\tiny$\pm$5.76} &59.83{\tiny$\pm$6.03} &\underline{26.69}{\tiny$\pm$7.85} &\underline{12.29}{\tiny$\pm$2.04}\\
    \method{MDGFM} &OOM &OOM &25.50{\tiny$\pm$3.38} &23.78{\tiny$\pm$3.82} &19.75{\tiny$\pm$5.24} &14.87{\tiny$\pm$2.27} &49.71{\tiny$\pm$23.54} &33.45{\tiny$\pm$11.56} &37.95{\tiny$\pm$5.56} &34.91{\tiny$\pm$5.16} &5.39{\tiny$\pm$3.62} &2.42{\tiny$\pm$0.71}\\
    \method{SAMGPT} &\textbf{12.33}{\tiny$\pm$0.71} &\textbf{9.86}{\tiny$\pm$0.33} &\textbf{37.27}{\tiny$\pm$3.48} &\textbf{36.35}{\tiny$\pm$3.50} &\textbf{38.52}{\tiny$\pm$5.37} &\textbf{30.69}{\tiny$\pm$3.36} &52.67{\tiny$\pm$12.71} &36.30{\tiny$\pm$5.89} &\underline{63.88}{\tiny$\pm$5.30} &\underline{63.78}{\tiny$\pm$5.29} &\textbf{32.07}{\tiny$\pm$8.02} &\textbf{15.53}{\tiny$\pm$1.98}\\
    \method{G2P2} &- &- &- &- &35.73{\tiny$\pm$9.31} &27.23{\tiny$\pm$5.76} &49.17{\tiny$\pm$19.45} &32.17{\tiny$\pm$10.53} &54.59{\tiny$\pm$5.87} &52.58{\tiny$\pm$5.88} &- &-\\
    \method{GraphCLIP} &- &- &- &- &22.17{\tiny$\pm$6.58} &16.39{\tiny$\pm$2.94} &38.17{\tiny$\pm$13.34} &19.50{\tiny$\pm$4.85} &35.64{\tiny$\pm$3.81} &33.78{\tiny$\pm$3.17} &- &-\\
    \method{GFT} &3.21{\tiny$\pm$1.05} &0.62{\tiny$\pm$0.04} &\underline{34.93}{\tiny$\pm$3.49} &\underline{33.80}{\tiny$\pm$3.74} &35.98{\tiny$\pm$5.39} &29.90{\tiny$\pm$3.47} &\textbf{63.23}{\tiny$\pm$16.19} &\textbf{41.26}{\tiny$\pm$7.47} &63.77{\tiny$\pm$5.56} &63.31{\tiny$\pm$5.31} &13.28{\tiny$\pm$5.31} &7.55{\tiny$\pm$1.19}\\
    \method{UniGraph2} &3.98{\tiny$\pm$0.86} &0.92{\tiny$\pm$0.59} &26.53{\tiny$\pm$4.44} &23.94{\tiny$\pm$4.04} &23.12{\tiny$\pm$9.68} &15.47{\tiny$\pm$4.14} &48.83{\tiny$\pm$16.44} &31.59{\tiny$\pm$12.50} &\textbf{74.36}{\tiny$\pm$10.53} &\textbf{68.29}{\tiny$\pm$2.38} &16.13{\tiny$\pm$6.69} &9.19{\tiny$\pm$2.11}\\\bottomrule
    \end{tabular}}
\end{table*}

\begin{table*}[tbp] % [!t]
    \centering
    \footnotesize
    \caption{Evaluation of 5-shot edge classification and graph classification on format domain adaptation.}
    \addtolength{\tabcolsep}{1.0mm}
    \label{table.exp4-5shotecgc}
    \resizebox{0.7\linewidth}{!}{
    \begin{tabular}{l|cccc|cc}
    \toprule
    \multirow{3}{*}{Methods} &\multicolumn{4}{c|}{Relational} &\multicolumn{2}{c}{Textual} \\
    &\multicolumn{2}{c}{Wiki} &\multicolumn{2}{c|}{WN18RR} &\multicolumn{2}{c}{PCBA} \\
    &Acc &MacroF &Acc &MacroF &Acc &MacroF \\
    \midrule\midrule
    \method{GCOPE} &OOT &OOT &18.91{\tiny$\pm$4.13} &14.92{\tiny$\pm$2.17} &56.03{\tiny$\pm$26.67} &33.97{\tiny$\pm$12.64}\\
    \method{MDGPT} &\underline{12.24}{\tiny$\pm$2.74} &\underline{4.14}{\tiny$\pm$0.65} &19.02{\tiny$\pm$3.24} &\textbf{15.73}{\tiny$\pm$2.10} &\textbf{56.56}{\tiny$\pm$18.54} &35.36{\tiny$\pm$8.28}\\
    \method{MDGFM} &8.42{\tiny$\pm$3.77} &1.49{\tiny$\pm$0.30} &\underline{19.09}{\tiny$\pm$5.56} &13.34{\tiny$\pm$2.32} &53.99{\tiny$\pm$44.41} &28.92{\tiny$\pm$22.20}\\
    \method{SAMGPT} &\textbf{15.15}{\tiny$\pm$2.81} &\textbf{6.10}{\tiny$\pm$0.22} &17.06{\tiny$\pm$3.39} &14.08{\tiny$\pm$1.74} &\underline{56.29}{\tiny$\pm$14.33} &\underline{35.66}{\tiny$\pm$6.18}\\
    \method{G2P2} &- &- &16.99{\tiny$\pm$8.41} &10.82{\tiny$\pm$2.67} &- &-\\
    \method{GraphCLIP} &- &- &\textbf{20.68}{\tiny$\pm$5.30} &\underline{15.06}{\tiny$\pm$2.01} &- &-\\
    \method{GFT} &OOT &OOT &16.49{\tiny$\pm$5.60} &10.06{\tiny$\pm$1.14} &55.21{\tiny$\pm$17.11} &\textbf{40.30}{\tiny$\pm$9.57}\\
    \method{UniGraph2} &6.60{\tiny$\pm$3.01} &3.61{\tiny$\pm$0.29} &11.93{\tiny$\pm$7.96} &8.49{\tiny$\pm$1.54} &52.79{\tiny$\pm$15.63} &34.30{\tiny$\pm$8.02}\\\bottomrule
    \end{tabular}}
\end{table*}

\end{document}